\documentclass{ecai}
\usepackage{graphicx}
\usepackage{latexsym}
\usepackage{times}
\usepackage{soul}
\usepackage{url}
\usepackage[hidelinks]{hyperref}
\usepackage[utf8]{inputenc}
\usepackage{graphicx}
\usepackage{amsmath}
\usepackage{amsthm}
\usepackage{float}
\usepackage[small]{caption2}
\usepackage{listings}
\usepackage{amssymb}
\usepackage{amsfonts}
\usepackage{bm}
\usepackage{multicol}
\usepackage{subfigure}
\usepackage{url}
\usepackage{booktabs} 
\usepackage[ruled,linesnumbered]{algorithm2e}
\usepackage{color}
\usepackage{arydshln} 
\usepackage{enumitem}

\begin{document}

\begin{frontmatter}

\title{BiERL: A Meta Evolutionary Reinforcement Learning Framework via Bilevel Optimization}

\author[1]{\fnms{Junyi}~\snm{Wang}$^\ast$\thanks{}}
\author[1]{\fnms{Yuanyang}~\snm{Zhu}$^\ast$\thanks{$\ast$ Equal contributions.}}
\author[1]{\fnms{Zhi}~\snm{Wang}$^\dag$\thanks{$\dag$ Correspondence to: Zhi Wang (zhiwang@nju.edu.cn).}}
\author[2]{\fnms{Yan}~\snm{Zheng}}
\author[2]{\fnms{Jianye}~\snm{Hao}}
\author[1]{\fnms{Chunlin}~\snm{Chen}$~$\thanks{Published as a conference paper at ECAI 2023.}} 


\address[1]{Department of Control Science and Intelligent Engineering, Nanjing University}
\address[2]{College of Intelligence and Computing, Tianjin University}

\begin{abstract}
Evolutionary reinforcement learning (ERL) algorithms recently raise attention in tackling complex reinforcement learning (RL) problems due to high parallelism, while they are prone to insufficient exploration or model collapse without carefully tuning hyperparameters (aka meta-parameters). In the paper, we propose a general meta ERL framework via bilevel optimization (BiERL) to jointly update hyperparameters in parallel to training the ERL model within a single agent, which relieves the need for prior domain knowledge or costly optimization procedure before model deployment. We design an elegant meta-level architecture that embeds the inner-level's evolving experience into an informative population representation and introduce a simple and feasible evaluation of the meta-level fitness function to facilitate learning efficiency. We perform extensive experiments in MuJoCo and Box2D tasks to verify that as a general framework, BiERL outperforms various baselines and consistently improves the learning performance for a diversity of ERL algorithms.
\end{abstract}

\end{frontmatter}

\section{Introduction}
\label{sec1}
Reinforcement learning (RL) is achieving tremendous breakthroughs in various fields, such as games~\cite{vinyals2019grandmaster}, robotics~\cite{zhu2021rule}, and even matrix multiplication in fundamental mathematics~\cite{fawzi2022discovering}.
An alternative approach to solving high-dimensional RL problems is using black-box optimization, yielding an explosive growth of evolutionary reinforcement learning (ERL) algorithms such as evolution-guided policy gradients~\cite{khadka2018evolution}, CERL~\cite{khadka2019collaborative}, MERL~\cite{majumdar2020evolutionary}, ERL-Re$^2$~\cite{li2022erl}, etc.
In this paper, we focus on a particular set of optimization algorithms in this class, evolution strategies (ES)~\cite{salimans2017evolution}, which has been reported to be competitive to popular backpropagation-based algorithms such as policy gradient and deep Q-learning due to better parallelization and strong exploration ability~\cite{wang2022instance}.

Typically, ES computes the zero-order search gradient by aggregating a population of individuals that come from perturbing the network parameters with Gaussian noise.
Unlike policy gradient methods that directly differentiate the learning objective, perturbing the parameter space has no explicit correlations to the policy outputs.
Hence, current ES algorithms are extremely sensitive to hyperparameters (aka meta-parameters), e.g., the noise covariance and learning rate, especially the noise level that controls the magnitude of exploration.
It could easily lead to insufficient exploration or model collapse without carefully tuning these hyperparameters.
Well-known procedures, such as random search and sequential optimization~\cite{jaderberg2017population}, require fully training enormous models to identify appropriate hyperparameters. 
While effective for simple tasks, they lack adaptation property in weight updates, which is validated to be effective from commonly used adaptive optimizers, such as Adam~\cite{kingma2014adam}.
Research also shows that the performance could still be brittle using static hyperparameters in complex and dynamic domains~\cite{choromanski2020provably}.
To leverage adaptation of hyperparameters, a straightforward idea is to train multiple agents with a large array of hyperparameters in parallel as in population-based training  (PBT)~\cite{jaderberg2017population}.
However, it could be too costly in practice due to maintaining enormous agents simultaneously within a single learning process~\cite{tang2020online}. 
Generally, existing works either require prior domain knowledge or expensive computation for reliable hyperparameter optimization.

To address this issue, we propose a meta ERL framework via bilevel optimization (BiERL) to adaptively update meta-parameters parallelly to training the ERL model within a single agent, relieving the need for explicit domain knowledge in advance or costly optimization procedure before model deployment.
The meta-level adaptively adjusts hyperparameters of the ES model according to the inner-level's population representation over the agent's experience, while the inner-level trains the ES model via the hyperparameters given by the meta-level.
We design an informative population representation encoder that considers the history of how the inner level evolves over a certain time horizon.
The ES algorithm also trains the meta-level network to maximize the final returns achieved after inner-level learning. 
To facilitate learning efficiency, we introduce a simple and feasible evaluation of the meta-level fitness function using a truncated estimate of complete inner-level optimization.
Also, we develop a nonparametric implementation based on Bayesian optimization (BO) to learn adaptive hyperparameters, promoting BiERL as a more general framework. 
Usually, the nonparametric meta-level design with BO is easier to be implemented with a simpler procedure, while the parametric one with a neural network can facilitate more effective joint adaptation on multiple hyperparameters~\cite{gu2021optimizing}.

In summary, our main contributions can be enumerated as follows: 
\begin{enumerate}
    \item We propose a novel and general meta ERL framework that produces efficient meta-parameter optimization without the requirement for prior domain knowledge or costly optimization procedure before model deployment.
    \item We introduce an elegant meta-level architecture that embeds the inner-level's evolving experience into an informative population representation, and we design a simple and feasible evaluation of the meta-level fitness function to facilitate learning efficiency.
    \item We perform extensive experiments to verify that as a general framework, BiERL outperforms various baselines and consistently improves learning performance across a diverse range of ERL algorithms.
\end{enumerate}

\section{Related Work}
\label{sec2}

\paragraph{Evolution strategies.}
ES algorithms Evolution Strategies (ES) algorithms have emerged as a scalable alternative for RL tasks, owing to their high parallelism and strong exploration capabilities~\cite{salimans2017evolution,khadka2018evolution}. 
However, previous works have observed that ES algorithms are exclusively on-policy, extracting only a limited amount of information from samples.
Consequently, in comparison to backpropagation-based algorithms, ES algorithms often necessitate more rollouts and are susceptible to local optima~\cite{choromanski2020provably}.
Previous efforts to enhance the performance of ES algorithms have included adding regularization terms to encourage exploration~\cite{conti2018improving} or employing various noise perturbations~\cite{ajani2022adaptive}. 
While ES algorithms are generally less sensitive to hyperparameter settings than traditional DRL~\cite{salimans2017evolution}, prior research has shown that hyperparameter fine-tuning can still be effective for them. 
Although some feasible ideas on hyperparameters have been proposed in these works~\cite{parker2020effective,suri2022off},  they are predominantly empirical and primarily focus on a single hyperparameter. 
In contrast to these previous efforts, our method can embed the agent's evolving experience to dynamically change multiple hyperparameters.

\paragraph{Hyperparameter optimization.}
Hyperparameter optimization methods for RL are mainly sequential optimization~\cite{jaderberg2017population}, such as Bayesian optimization. 
However, these methods need multiple training runs to find optimal hyperparameters. 
Some other hyperparameter optimization methods, e.g., evolutionary algorithms, are categorized as parallel search ~\cite{liashchynskyi2019grid}. These methods have high parallelism but need more computational resources. 
Both of these methods find static hyperparameters before training.
Nevertheless, in complex tasks, static hyperparameters may lead to a decline in performance.

To overcome this limitation, PBT~\cite{jaderberg2017population}, which combines sequential optimization with evolutionary algorithms, has shown promising potential in online hyperparameter optimization due to its generality and robustness. 
Inspired by this concept, several successful works have been proposed, such as ALFA~\cite{baik2020meta}, OHT-ES~\cite{tang2020online}, and HyperDistill~\cite{lee2021online}. 
In these methods, multiple agents with a large array of hyperparameters are updated in parallel to the model training. 
However, it could be too costly in practice due to maintaining enormous agents simultaneously.
Additionally, while the online hyperparameter optimization of ES algorithms has already shown its effectiveness in previous works~\cite{plappert2018parameter}, most works on ES algorithms still rely on fine-tuning to find static hyperparameters before model training, which is exactly the main novel contribution of our methods. 




\paragraph{Meta-ES.}
To improve performance, some works attempt to combine ES algorithms with meta-learning. These hybrid methods are named meta-ES. Existing meta-ES methods can be divided into two categories, multi-task meta-ES, and single-task meta-ES.

Most meta-ES methods belong to the multi-task meta-ES, which aims to train agents that can quickly adapt to similar but unseen tasks with ES algorithms.  Representative works include ES-MAML~\cite{song2019maml}, Evolvability ES~\cite{gajewski2019evolvability}, and Baldwinian Meta-Learning~\cite{fernando2018meta}, which are all based on the MAML framework~\cite{finn2017model}. Other works, such as EPG~\cite{houthooft2018evolved} and EvoGrad~\cite{bohdal2021evograd}, take advantage of the gradient-free features of ES algorithms to avoid calculating second-order derivatives in meta-learning. 
Nevertheless, these methods merely treat ES algorithms as an auxiliary technique and ignore the performance of the ES algorithm itself. 

On the contrary, single-task meta-ES methods focus on the meta-parameters  (aka hyperparameters) of ES algorithms. These works point out that the meta-parameters need to vary as learning progresses for a better performance~\cite{plappert2018parameter}. To realize an adaptive meta-parameters, existing meta-ES methods include  $[1, 2 (1, \lambda)\gamma ]$-meta-ES~\cite{hellwig2019analysis}, OMPAC~\cite{elfwing2018online}, and Meta-evolution~\cite{bossens2020learning}. However, these methods only apply simple techniques such as random search or heuristic function, which are only for solving low-dimensional tasks. Different from these works, we make it feasible for our method to dynamically change meta-parameters in more complex RL tasks.

\paragraph{Bilevel optimization.}
Bilevel optimization includes two sub-objectives called the meta/upper-level and inner/lower-level objectives, respectively. The meta-level objective must be solved subject to the optimality of the inner-level one and vice versa. 

The mixture of RL algorithms and bilevel optimization framework has achieved success in many domains~\cite{sinha2017review}, such as neural architecture search~\cite{pang2021rl}, imitation learning ~\cite{arora2020provable}, intrinsic rewards~\cite{stadie2020learning}, and hyperparameter optimization~\cite{mackay2018self}. Our method also utilizes a bilevel optimization framework for better generalizability in different ERL methods. Additionally, to compensate for additional computation costs from the conventional bilevel optimization framework, we propose a simple and feasible approximation of the meta-level evaluation and introduce a warm starting mechanism to replace random initialization in our method.

\section{Method}
\label{sec4}

\subsection{Bilevel Framework}
\label{sec4.1}
Our proposed BiERL is a meta ERL framework via bilevel optimization that jointly updates hyperparameters $\mathcal{H}_t$ at a certain timestep $t$ in parallel to training the parameters of the ES model within a single agent.  
We aim to solve a bilevel optimization problem where our goal is to find hyperparameters $\mathcal{H}_t$ that maximize the meta-level fitness function $F$ of the model parameterized by ${\theta_t}$ and trained with the inner-level fitness function $f$ and $\mathcal{H}_t$:
\begin{equation}
\begin{aligned}
& \mathcal{H}_{t+1}^* = \arg\max_{\mathcal{H}_{t}}F (\mathcal{H}_{t}, \theta_t^* (\mathcal{H}_{t})), \\
\text{where}~& \theta_t^* (\mathcal{H}_t)=\arg\max_{\theta_t}f (\mathcal{H}_t, \theta_t).
\end{aligned}
\end{equation}
  
\begin{figure}[tb]
    \centering
    \includegraphics[width=\columnwidth]{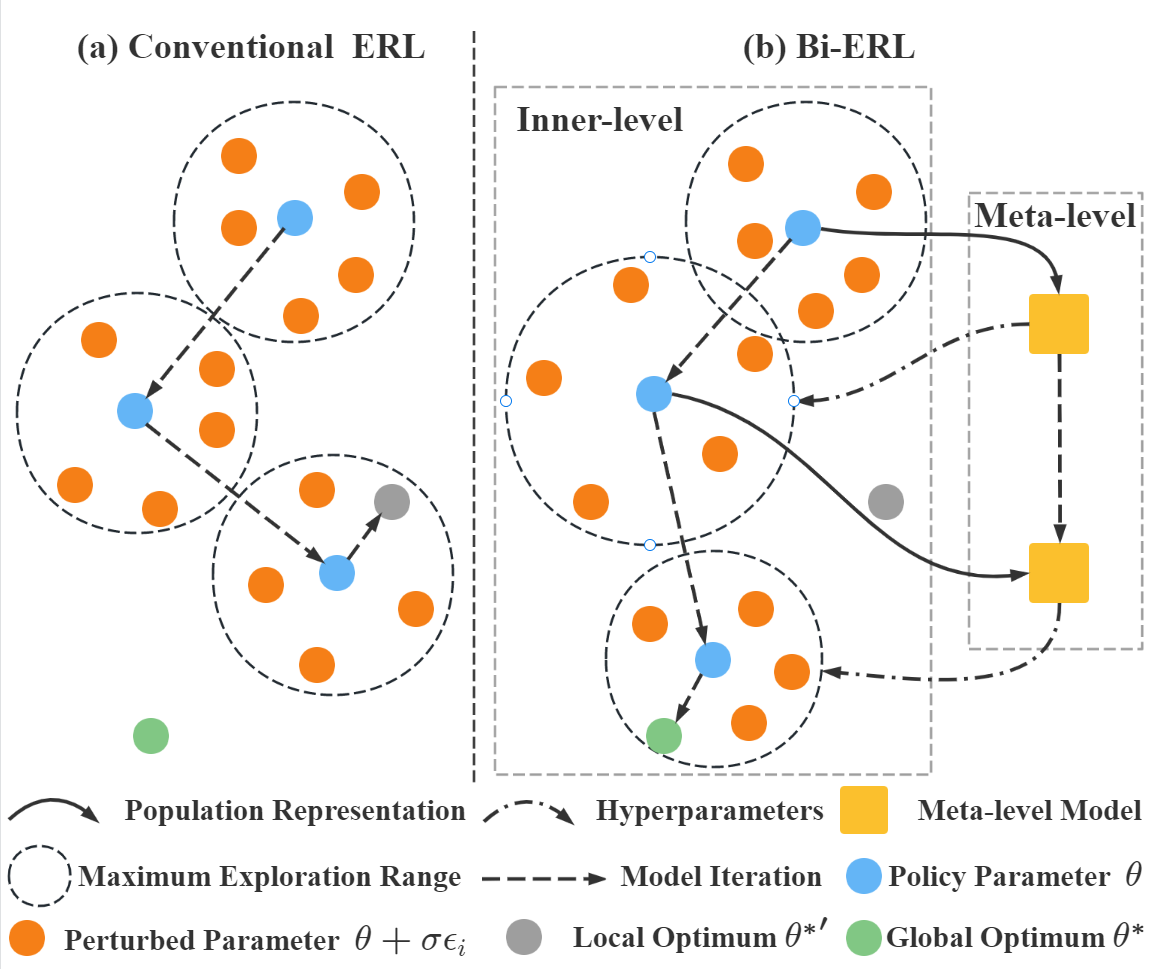}
    \caption{The overall structure of BiERL.} 
\label{fig.1}
\end{figure}

Figure~\ref{fig.1} presents the overall structure of BiERL and the comparison to conventional ERL methods.
Generally, conventional ERL methods train the model with static hyperparameters, e.g., the noise covariance and learning rate of ES. 
However, static hyperparameters could easily lead to inappropriate exploration or even model collapse without explicit domain knowledge in advance when dealing with complex tasks.
On the contrary, our BiERL framework attempts to tackle such deficiencies by introducing adaptive hyperparameter optimization.
The meta-level will adaptively adjust hyperparameters of ERL according to how the inner-level evolves over a certain time horizon.
Then, the inner level trains the ERL model for several timesteps using the hyperparameters given by the meta-level.
Taking the noise level of ES as an example, BiERL is more likely to bypass bad local optima by enlarging the noise covariance to encourage more exploration in the parameter space and also avoid model collapse by annealing the parameter noise.

\subsection{Inner-level}
\label{sec4.3}
In the inner level of BiERL, the agent trains the parameters $\theta_t$ via ES to maximize the fitness function $f (\theta_t (\mathcal{H}_t))$ based on the  hyperparameters $\mathcal{H}_t$ from the meta-level. 

\paragraph{Evolution strategies.} Inspired by natural evolution, ES algorithms are representative algorithms of ``black-box optimization". ES algorithms seek to optimize the fitness function  $f (\theta)=\mathbb{E}_{\tau\sim\pi_\theta (\tau)}\left[\sum\nolimits_{i=0}^{\infty}\gamma^ir_i\right]$, which evaluates the parameters $\theta$. Instead of directly calculating the gradient of the return like backpropagation-based algorithms, ES algorithms use an estimator of the ``search gradient" based on a population of perturbed parameters.

In each iteration, perturbed parameters $\theta+\sigma\epsilon_i,\ i=1,2,...,n$ are sampled and evaluated, where $\sigma$ is the noise covariance, $n$ is the size of the population, and  $\epsilon_i\sim \mathcal{N} (0,  \bm{I})$ is the Gaussian noise. 
Thus, the Monte Carlo estimate of the search gradient $\nabla_\theta {f} (\theta)$ becomes
\begin{equation}
\begin{aligned}
    \nabla_\theta  {f} (\theta)&=\nabla_\theta \mathbb{E}_{\epsilon\sim \mathcal{N} (0,   \bm{I})}[f (\theta+\sigma\epsilon)]\\&=\frac{1}{\sigma}\mathbb{E}_{\epsilon\sim \mathcal{N} (0,   \bm{I})}[f (\theta+\sigma\epsilon)\epsilon]\\&\approx \frac{1}{n\sigma}\sum_{i=1}^{n}f (\theta +\sigma \epsilon_i)\epsilon_i.
\end{aligned}
\end{equation}
Then, the model parameters $\theta_t$ are updated as 
\begin{equation}
\begin{aligned}
    \theta_{t+1}\leftarrow \theta_t+ \frac{\alpha}{n\sigma}\sum_{i=1}^{n}f (\theta_t + \sigma\epsilon_i)\epsilon_i,
\end{aligned}
\end{equation}
where $\alpha$ is the learning rate. 
    
As a general framework, BiERL is able to update hyperparameters for various ERL methods, e.g., Vanilla ES~\cite{salimans2017evolution}, NSR-ES~\cite{conti2018improving}, and ESAC~\cite{suri2022off}.
Experimental results have verified that our BiERL framework can improve the learning performance for multiple ES algorithms.
During the training of the inner-level network, the population representation at each timestep $\boldsymbol{f}_t$ will be stored in a small-scale replay buffer for the meta-level.
The agent will update the meta-level model when the inner-level is trained with $k$ iterations.

\subsection{Meta-level}
\label{sec4.2}

First, we introduce the network architecture of the meta-level design with the informative population representation encoder. 
Then, we present the training process of the meta-level, where a simple and feasible evaluation of the fitness function is crucial for efficient learning.
In addition, a warm starting mechanism is introduced to facilitate the training process.
Finally, we give the nonparametric implementation of the meta-level to learn adaptive hyperparameters.

\paragraph{Network architecture.}
The meta-level aims to adjust the meta-parameters according to the inner-level's evolving situation.
An elegant design of the meta-level architecture should at least satisfy three conditions: (1) maintaining a small scale to alleviate the costly optimization procedure; (2) capturing the history information of the evolution to gain better representation; (3) exhibiting good generalization across different tasks and diverse architectures of the inner-level.

\begin{figure}[tb]
\centering
\includegraphics[width=0.98\columnwidth]{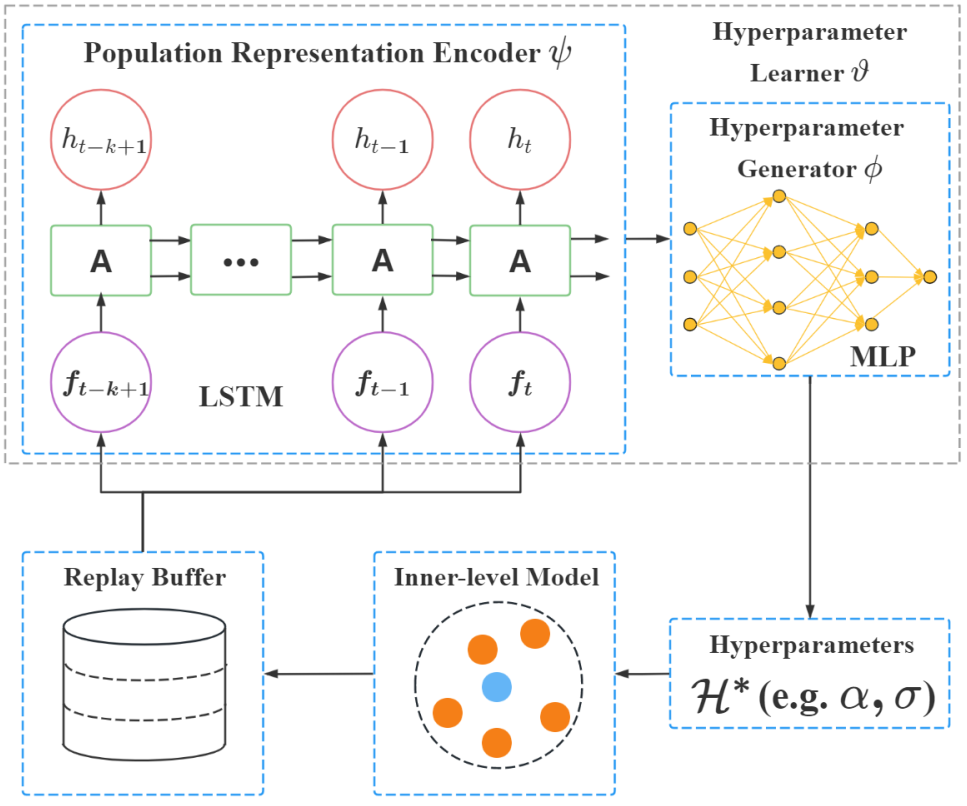}
\caption{The structure of the Meta-level model.}
    \label{fig.2}
\end{figure}

Inspired by this, Figure~\ref{fig.2} presents the proposed network structure of the meta-level model.
We consider the information of all the individuals in the population over a certain time horizon.
In each iteration of the inner-level, a population of $n$ perturbed parameters $\theta_t +\sigma_t \epsilon_i$, $i=1,...,n$ is generated with the fitness function $\boldsymbol{f}_t = [f (\theta_t+\sigma_t\epsilon_1),...,f (\theta_t+\sigma_t\epsilon_n)]$ calculated.
Here, we take this kind of population representation over the past $k$ timesteps $\mathbb{S}_t =[\boldsymbol{f}_{t-k+1}, . . . ,\boldsymbol{f}_t]$ as the input to the meta-level network.
We use a replay buffer to store the inner-level's fitness function over the past $k$ timesteps as $\mathbb{S}_t$, which is fed into an LSTM network $\psi$ to encode the evolving population representation as $\mathbb{X}_{t} = \psi_t (\mathbb{S}_t)$. 
Then, an MLP $\phi$ takes as input the population representation that indicates the history evolving situation and outputs the adaptive hyperparameters for the inner-level ES model, e.g., the noise covariance $\sigma$ or learning rate $\alpha$, as $\mathcal{H}_{t+1} = \phi_t (\mathbb{X}_t)$.

Using the above design, the dimension of the meta-level's input can remain at a small scale, as opposed to using the entire network weights that may reach millions of dimensions~\cite{baik2020meta}.
This is significant for efficiently deploying adaptive hyperparameter optimization on real-world applications.
Then, the evolving situation of the inner-level can be precisely embedded by the population representation encoder that leverages the advances of LSTM.
Finally, using the fitness function as the population representation is agnostic to the inner-level's network design and the task at hand. 
Existing works usually use the entire weights of the inner-level as the input~\cite{baik2020meta}, leading to different meta-network designs for different tasks due to different dimensions of the state-action spaces.
In contrast, our meta-network can keep the same architecture for different tasks by using the task-agnostic fitness function as the population representation.
Hence, the meta-level model is agnostic to the design of the inner-level's network design (other than the algorithm generalization across tasks), which could enjoy better method universality.

\paragraph{Model training.}
The meta-level network is also trained using the ES algorithm to maximize the final returns achieved after inner-level learning.
Let $\vartheta=[\psi, \phi]$ denote the parameters of the meta-level model. 
Evaluating the meta-level fitness function $F (\mathcal{H}_{\vartheta_t}, \theta_t^* (\mathcal{H}_{\vartheta_{t}}))$ is essential for the model training.
The evaluation is based on the final return after the inner-level ES model has been completely optimized with the hyperparameters $\mathcal{H}_{\vartheta_{t}}$ given by the meta-level. 
It is obvious that such a process could largely decrease learning efficiency.

To design a simple and feasible evaluation of $F (\mathcal{H}_{\vartheta_t}, \theta_t^* (\mathcal{H}_{\vartheta_{t}}))$, we use the return after one-step optimization as a truncated estimate of the return after complete inner-level optimization.
Specifically, the meta-level will first sample a group of $m$ perturbed parametric models $\vartheta_{t}^j=\vartheta_{t}+\omega\varepsilon_j, \varepsilon_j\sim\mathcal{N} (0, \bm{I}), j = 1,2,...,m $, where $\omega$ is the meta-level noise covariance.
Each perturbed model $\vartheta_{t}^j$ will output a set of hyperparameter $\mathcal{H}_{\vartheta_{t}^j}$ based on the inner-level's population representation.
Then, the inner-level model $\theta_t (\mathcal{H}_{\vartheta_{t}^j})$ will be updated to $\hat{\theta}_t (\mathcal{H}_{\vartheta_{t}^j})$ with only one-step optimization as
\begin{equation}
\hat{\theta}_t^j \leftarrow \theta_t^j+ \frac{\alpha_t^j}{n\sigma_t^j}\sum_{i=1}^{n}f (\theta_t^j + \sigma_t^j\epsilon_i)\epsilon_i,~~~\epsilon_i\sim\mathcal{N} (0,\bm{I}),
\end{equation}
where $\hat{\theta}_t^j$ and $\theta_t^j$ denote $\hat{\theta}_t (\mathcal{H}_{\vartheta_{t}^j})$ and $\theta_t (\mathcal{H}_{\vartheta_{t}^j})$, respectively, and $\alpha_t^j$ and $\sigma_t^j$ are adaptive hyperparameters generated by $\mathcal{H}_{\vartheta_{t}^j}$.
We repeat this process for $l$ times to obtain a Monte Carlo estimate of the final meta-level fitness function as
\begin{equation}
\begin{aligned}
F\left (\mathcal{H}_{\vartheta_t^j}, \theta_t^* (\mathcal{H}_{\vartheta_t^j})\right) & \approx F\left (\mathcal{H}_{\vartheta_t^j}, \hat{\theta}_t (\mathcal{H}_{\vartheta_t^j})\right) \\
&= \mathbb{E} \left[f\left (\hat{\theta}_t (\mathcal{H}_{\vartheta_t^j})\right)\right] \\
&\approx\frac{1}{l} \sum\limits_{i=1}^{l}[f\left (\hat{\theta}_t^{j,i}\right)].
\end{aligned}
\label{eq.4}
\end{equation}
The search gradient estimation of $\vartheta$ is computed as
\begin{equation}
\begin{aligned}
   \vartheta_{t+1} \leftarrow \vartheta_t + \frac{\beta}{m\omega}\sum\limits_{j=1}^{m}F\left (\mathcal{H}_{\vartheta_t^j}, \theta_t^* (\mathcal{H}_{\vartheta_t^j})\right)\varepsilon_j,
\end{aligned}\label{eq.5}
\end{equation}
where $\beta$ is the learning rate of the meta-level model.
Algorithm~\ref{alg.1} summarizes the training process.

\begin{algorithm}[tb]
  \SetAlgoLined
  \KwIn{Meta-level population size $m$, meta-level learning rate $\beta$, meta-level noise covariance $\omega$, $\theta_t$, $\mathbb{S}_t$, $\vartheta_t = [\phi_t,\psi_{t}]$}
  \KwOut{New meta-level model $\vartheta_{t+1}$}
  Sample $\varepsilon_1, . . . , \varepsilon_m\sim\mathcal{N} (0, \bm{I})$\\
    \For{$j=1,  2,  . . .,m $}{
$\vartheta_{t}^j = \vartheta_{t} + \omega\varepsilon_j$ 
$\mathbb{X}_{t}^j=\psi_t^j (\mathbb{S}_t)$\\
$\mathcal{H}_{\vartheta_{t}^j} = \phi_t^j (\mathbb{X}_{{t}}^j)$\\
Estimate $F (\mathcal{H}_{\vartheta_{t}^j}, \theta_t^* (\mathcal{H}_{\vartheta_{t}}^j))$  by Eq.~(\ref{eq.4})\\
      }
    Update $\vartheta_t$ by Eq.~(\ref{eq.5}) 
  \caption{Training process of the meta-level} 
  \label{alg.1}
\end{algorithm}

\begin{figure}[tb]
\centering
\includegraphics[width=0.72\columnwidth]{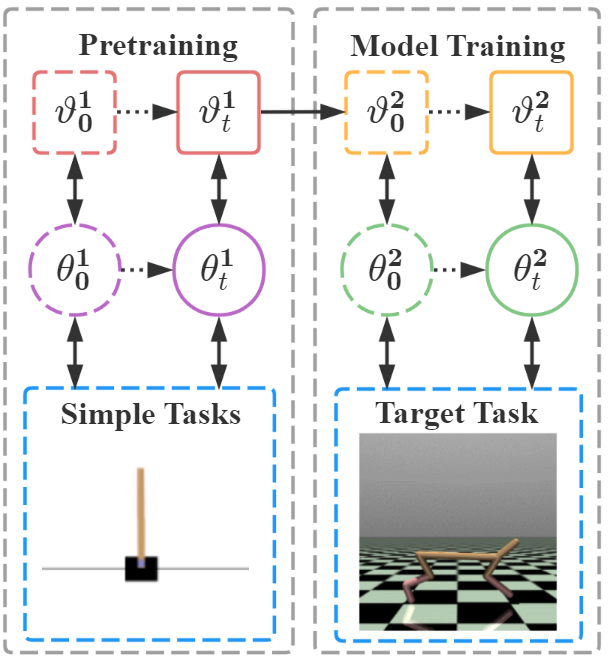}
\caption{Warm starting mechanism.}
    \label{fig.3}
\end{figure}

\paragraph{Warm starting.}
The training of the meta-level model is complicated since it involves an ES update of the inner-level for each individual in the meta-level population.
Inspired by~\cite{ash2020warm}, we introduce a warm staring mechanism to improve the training process.
Instead of random initialization, we pretrain the meta-level model in simple tasks and use it as the initialization of the model in the target task, as shown in Figure~\ref{fig.3}.
Conventional warm-starting mechanisms may require tasks to be relatively similar.
However, as mentioned above, the design of the population representation encoder is agnostic to the task at hand, so it is more feasible for the warm starting mechanism to transfer some knowledge from simple tasks to help train the difficult target task. In practice, we only pretrain the meta-level once in a simple task and save the meta-level for any future initialization.

\paragraph{Nonparametric design.}
To promote BiERL as a more general framework, we also develop a nonparametric design of the meta-level based on BO to learn adaptive hyperparameters. 
Different from the parametric design that requires maintaining a network, the nonparametric model only constructs a dynamic meta-level fitness function $F (\mathcal{H}_{{t}}, \theta_t^* (\mathcal{H}_{{t}}))$  and find the optimal solution with BO periodically. 
This is equivalent to dividing the entire training process into iteratively alternating between training the inner-level ES model and optimizing the inner-level's hyperparameters through BO.
Generally, the nonparametric design with BO is easier to be implemented with a simpler architecture, while the parametric one with the neural network architecture can facilitate more effective joint adaptation on multiple hyperparameters.
Analogous to the estimation approach introduced in the parametric model, the meta-level fitness function $F (\mathcal{H}_t, \theta_t^* (\mathcal{H}_t))$ is also constructed after one-step optimization as
\begin{equation}
F (\mathcal{H}_t, \theta_t^* (\mathcal{H}_t))  \approx \frac{1}{l} \sum_{i=1}^{l}\left[f (\hat{\theta}_t^i)\right].
\label{eq.6}
\end{equation}

\subsection{Integrated Algorithm}
\label{sec4.4}
Based on the above implementations, Algorithm~\ref{alg.2} presents the integrated BiERL framework that jointly updates hyperparameters in parallel to training the ES model within a single agent. 
In the meta-level, we can apply the parametric network with ES update or the nonparametric BO procedure to produce adaptive hyperparameters.
These two designs are suitable for different situations with respective advantages, which will be furtherly demonstrated in the experiments.


\section{Experiments}
\label{sec5}
We conduct extensive experiments to verify the effectiveness of BiERL on a benchmark of continuous control problems in OpenAI Gym~\cite{brockman2016gym,wang2022lifelong}.  
Specifically, we seek to answer the following research questions (RQs).
The source code is available at~\url{https://github.com/chriswang98sz/BiERL}.
\begin{itemize}
\item \textbf{RQ1  (Performance):} Can BiERL improve the performance of basic ERL methods across different RL tasks?
\item \textbf{RQ2  (Scalability):} Can BiERL framework effectively adapt to other hyperparameter optimization? 
\item \textbf{RQ3  (Ablation):} What are the respective contributions of different modules to overall performance?
\end{itemize}
  
  \begin{algorithm}[tb]
  \SetAlgoLined
  \KwIn{Population size $n$, interval $k$}
  \KwOut{Optimal inner-level policy parameters $\theta^*$}
  Initialize $\theta_0$, $\mathcal{H}_0$,  Replay Buffer $\mathcal{R}$
  
  Warm starting $\vartheta_{0} = [\phi_0,\psi_{0}]$
  
  \For{$t=0,  1,  2,  . . . $}{
    \uIf ({\ ~~~// Meta-level}) {Use parametric model}{
Update $\vartheta_{t}$ by Algorithm~\ref{alg.1}}
 \ElseIf{Use nonparametric model}{
Construct $F (\mathcal{H}_{{t}}, \theta_{t}^* (\mathcal{H}_{{t}}))$ by Eq.~(\ref{eq.6})\\
Use BO to find $\mathcal{H}_{{t+1}}^*$} 
    
    \For ({\ ~~~// Inner-level}){$i=0,  1,  2,  . . .,k-1 $}{
    Use ERL to update $\theta_{t+1,i}$ with $\mathcal{H}_{t+1,i}$\\
    \If{Use parametric model}{
      Store $\boldsymbol{f}_{t+1,i}$ to $\mathcal{R}$\\
      $\mathbb{S}_{t+1,i} =[\boldsymbol{f}_{t,i+1},... ,\boldsymbol{f}_{t+1,i}]$\\
    $\mathbb{X}_{t+1,i}=\psi_{t+1} (\mathbb{S}_{t+1,i})$\\
    $\mathcal{H}_{t+1,i+1} =  \phi_{t+1} ( \mathbb{X}_{{t+1,i}})$\\
      }
    }
   } 
  \caption{BiERL}
  \label{alg.2}
\end{algorithm}

  \begin{figure*}[h!]
  
			\centering 
		\subfigure[Ant-v2]{\includegraphics[width=0.230\textwidth]{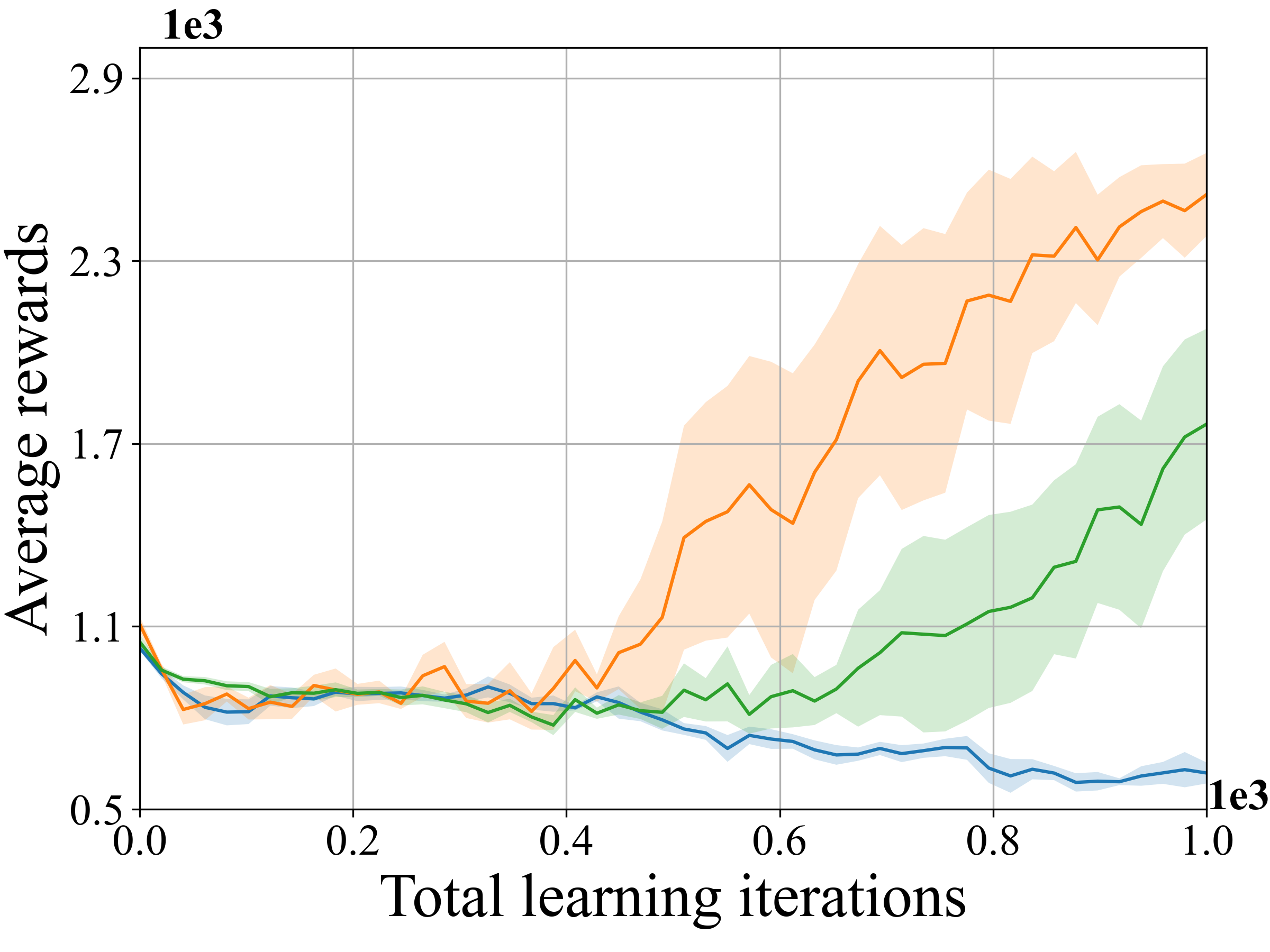}}
		\subfigure[HalfCheetah-v2]{\includegraphics[width=0.230\textwidth]{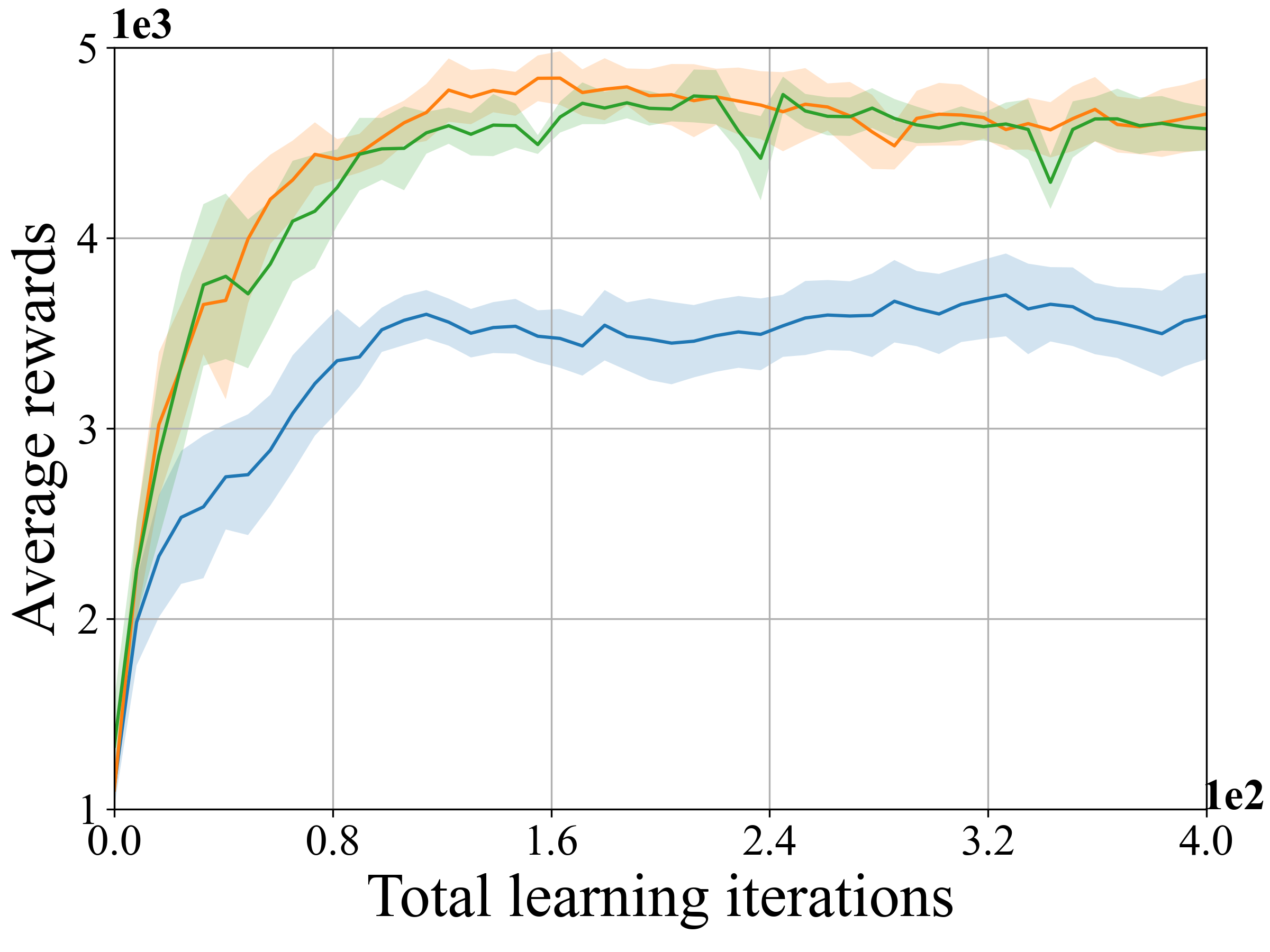}}
		\subfigure[Walker2d-v2]{\includegraphics[width=0.230\textwidth]{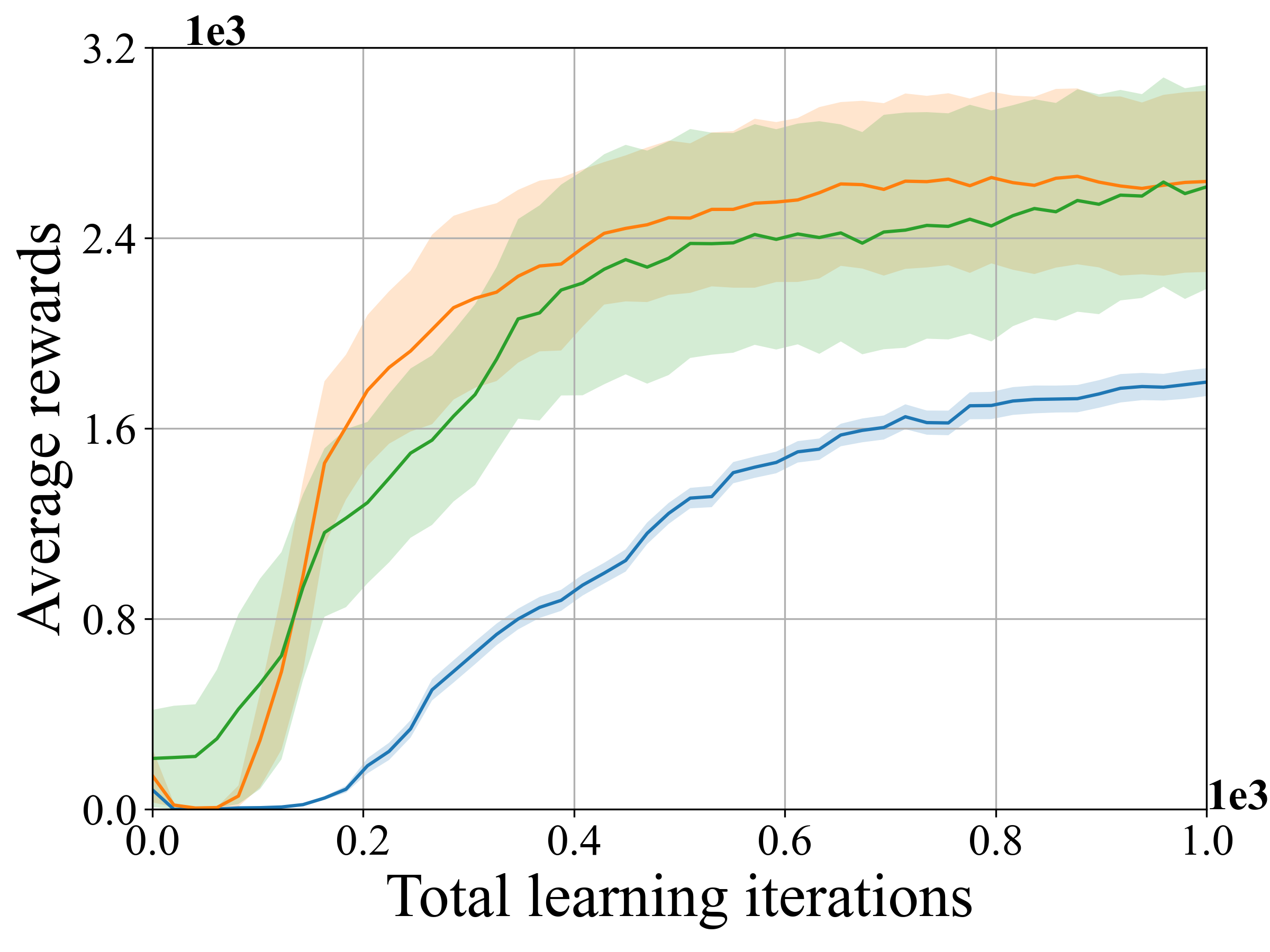}}
		\subfigure[LunarLanderContinuous-v2]{\includegraphics[width=0.230\textwidth]{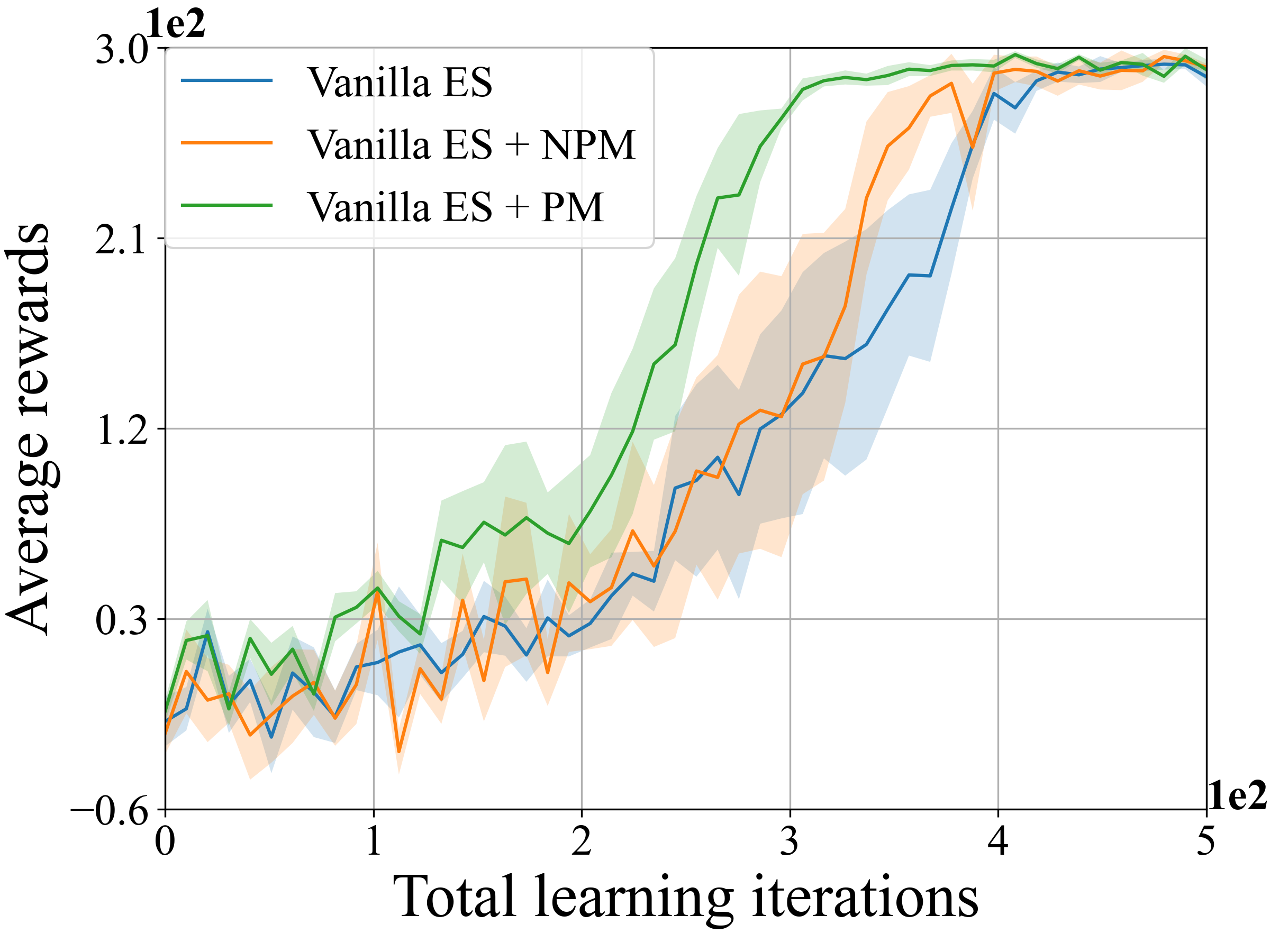}}
		\subfigure[Ant-v2]{\includegraphics[width=0.230\textwidth]{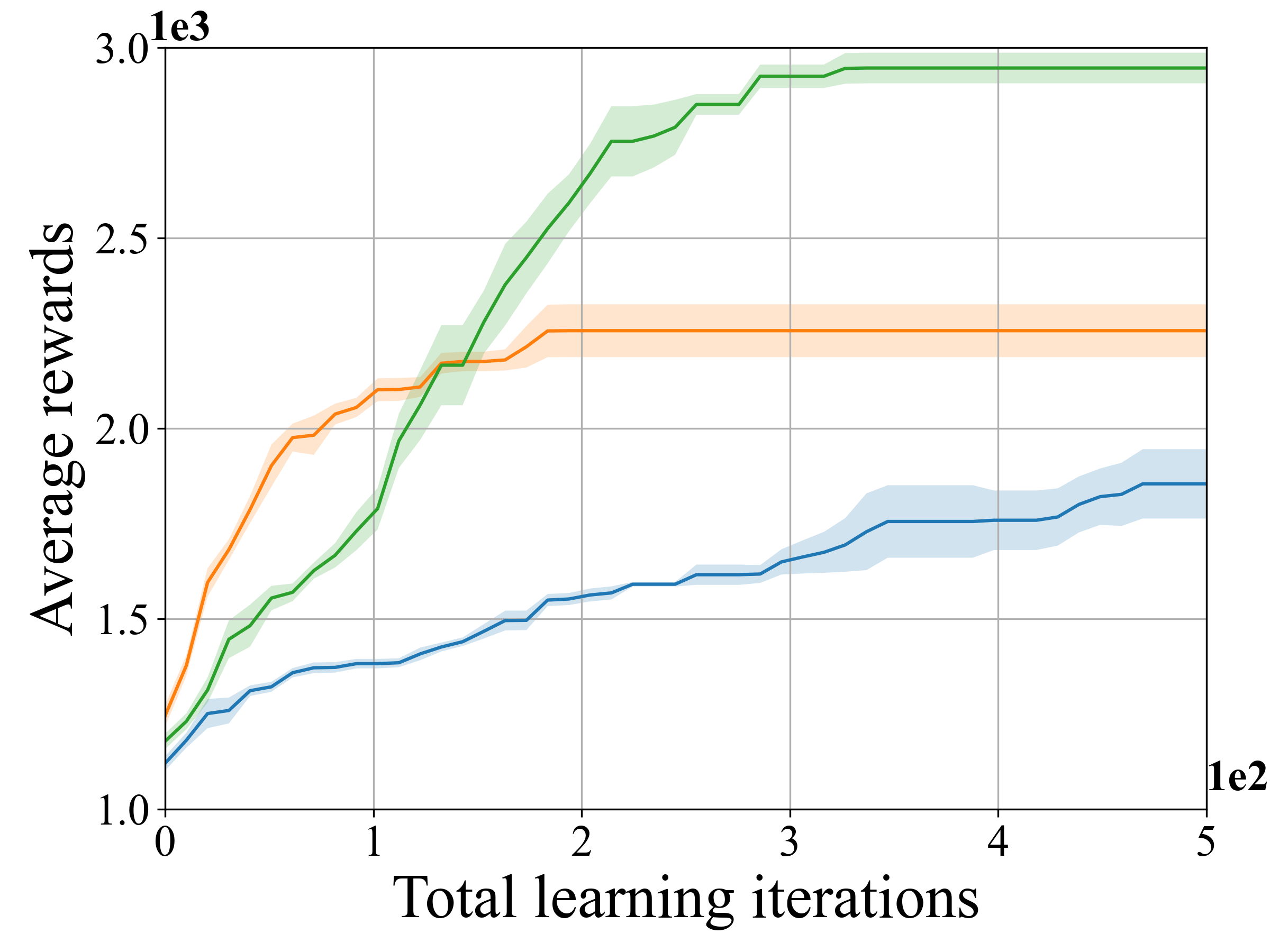}}
		\subfigure[HalfCheetah-v2]{\includegraphics[width=0.230\textwidth]{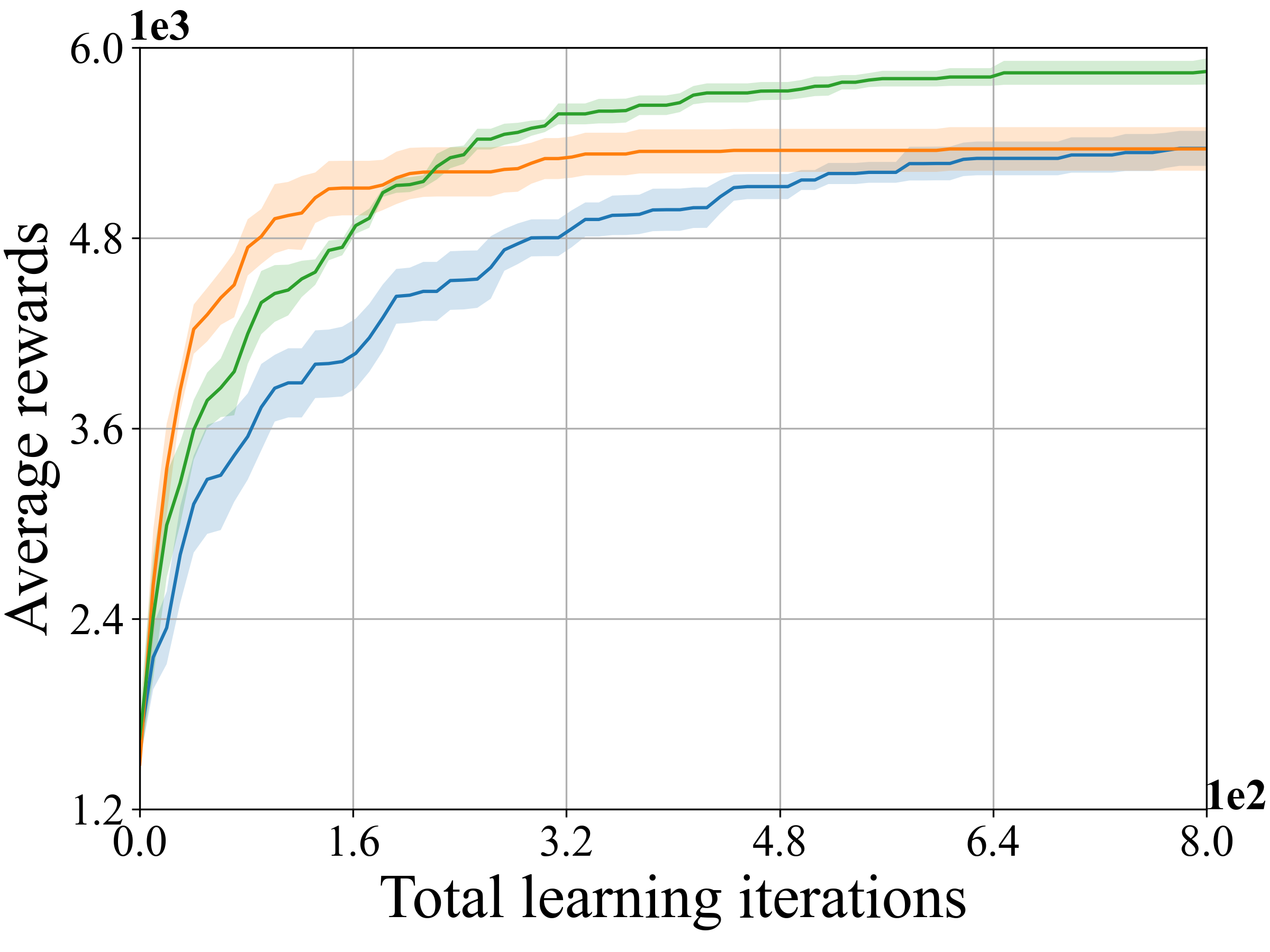}}
		\subfigure[Walker2d-v2]{\includegraphics[width=0.230\textwidth]{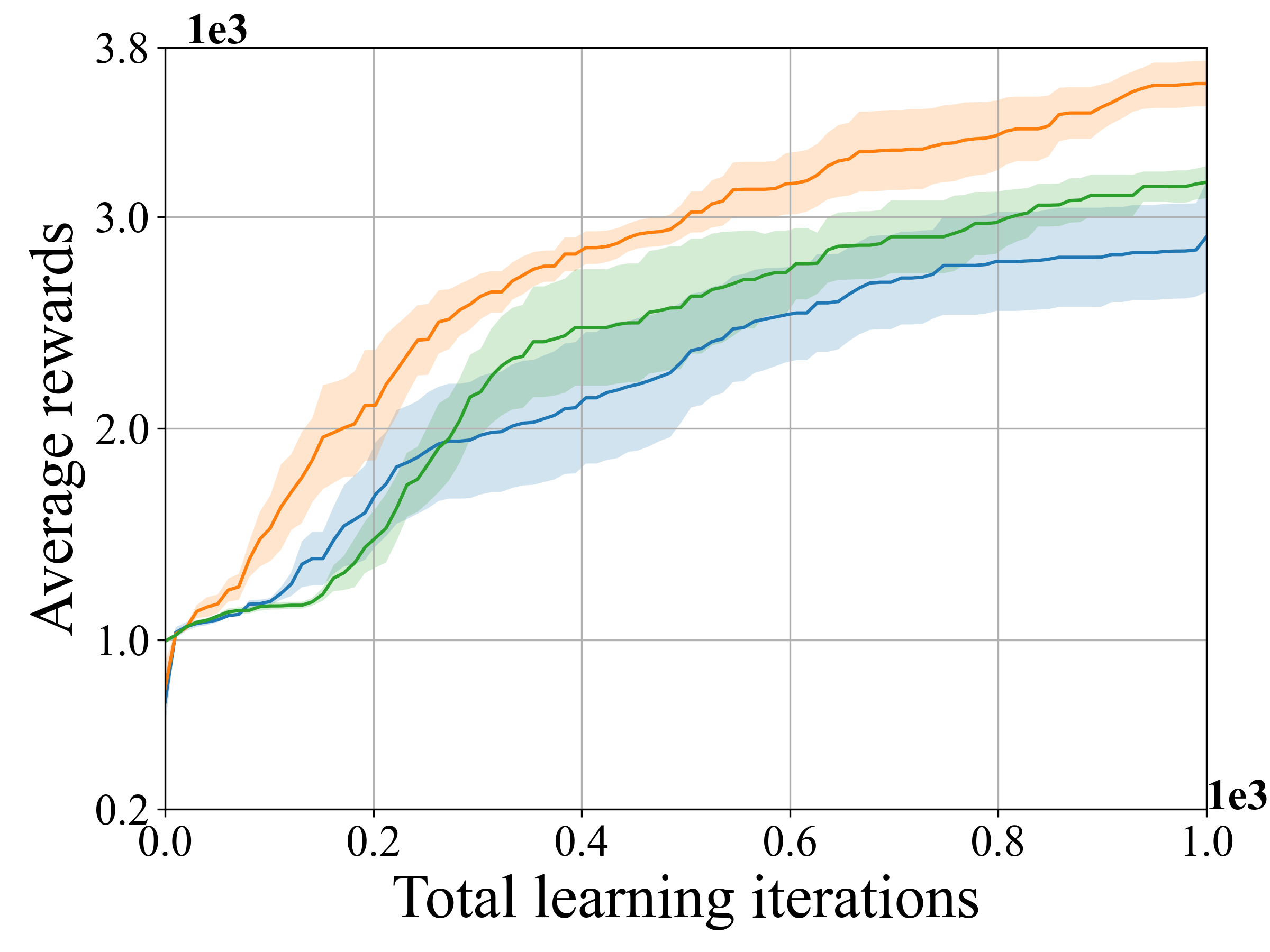}}
		\subfigure[LunarLanderContinuous-v2]{\includegraphics[width=0.230\textwidth]{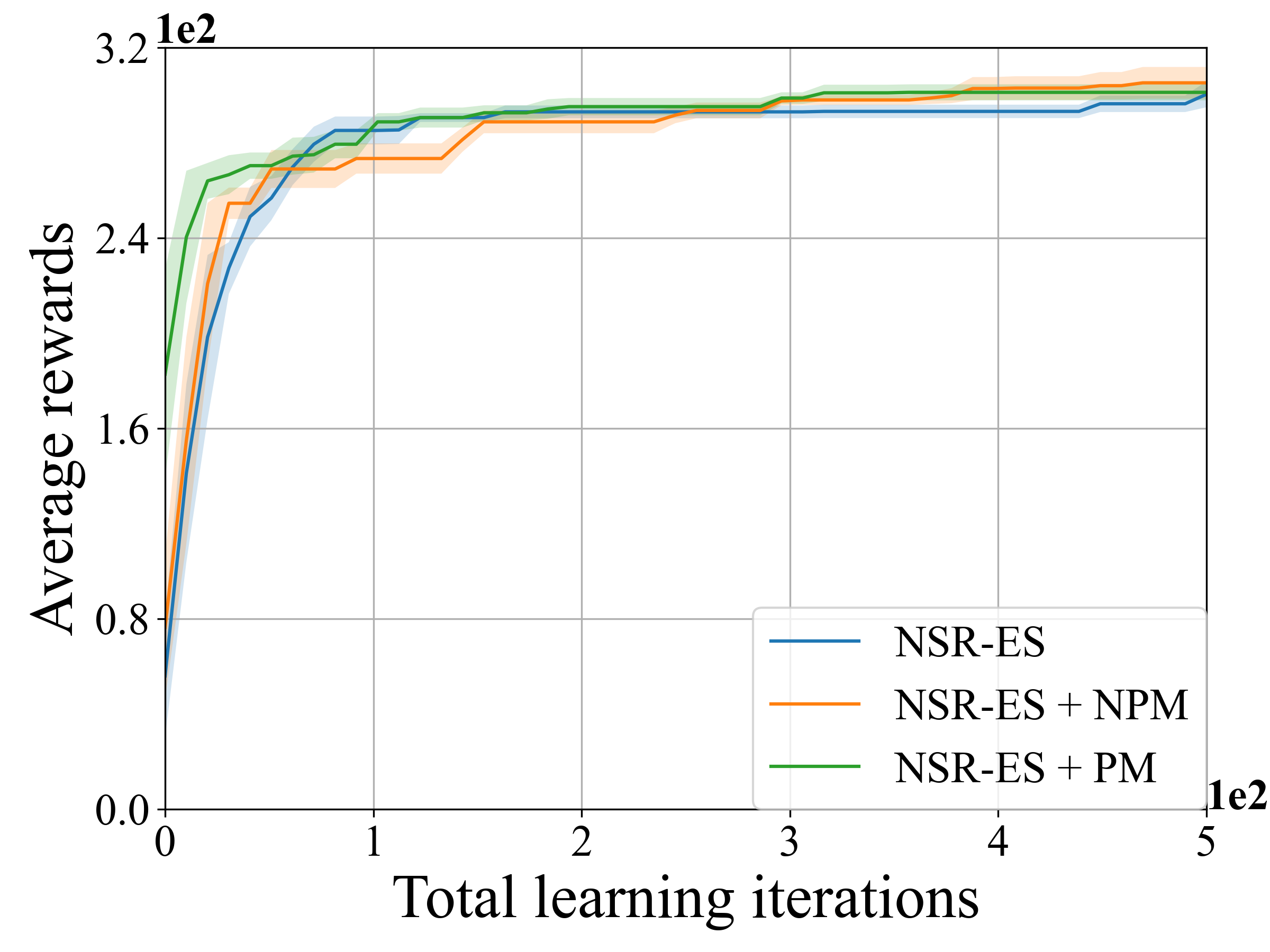}}
		\subfigure[Ant-v2]{\includegraphics[width=0.230\textwidth]{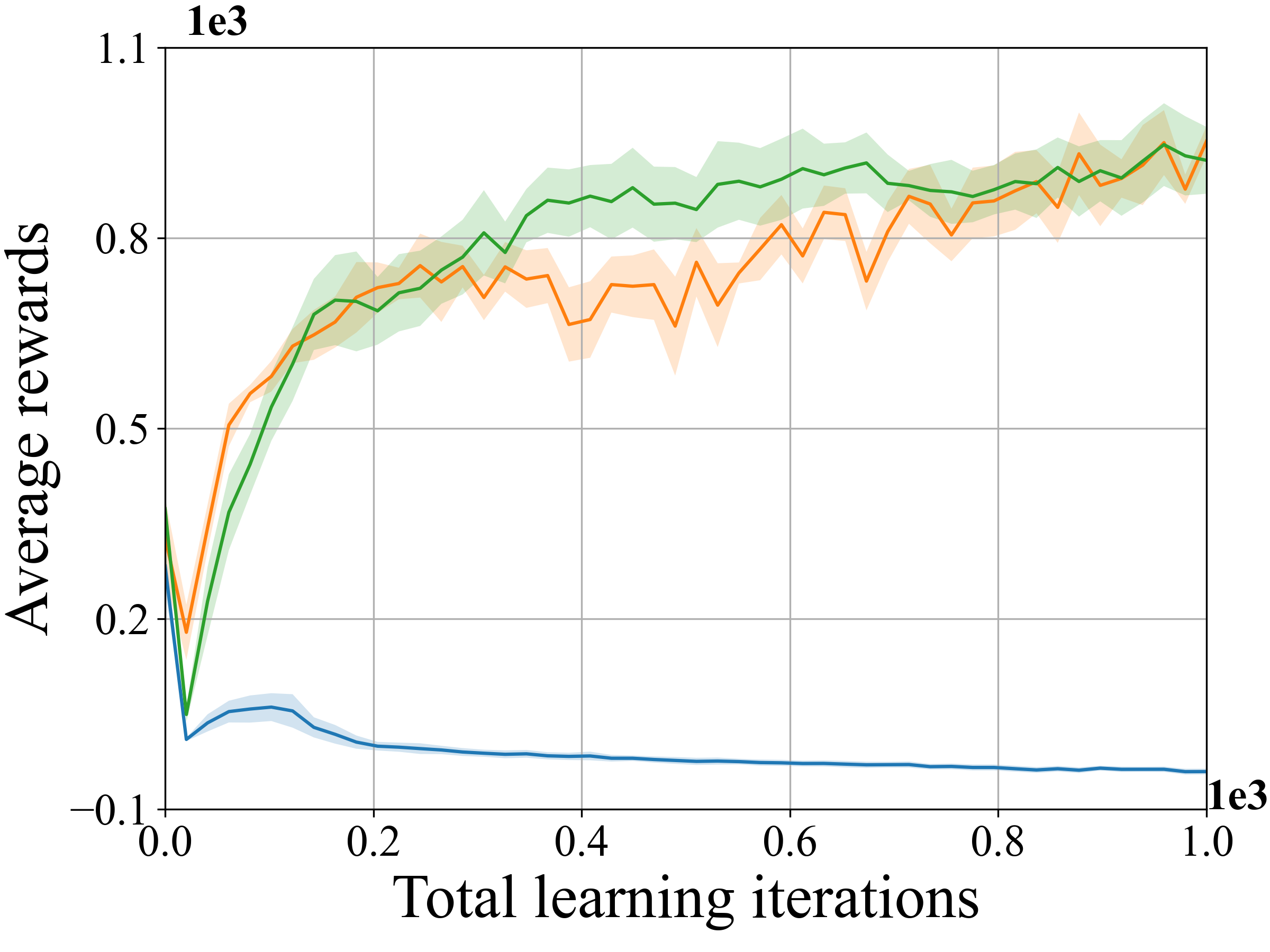}}
		\subfigure[HalfCheetah-v2]{\includegraphics[width=0.230\textwidth]{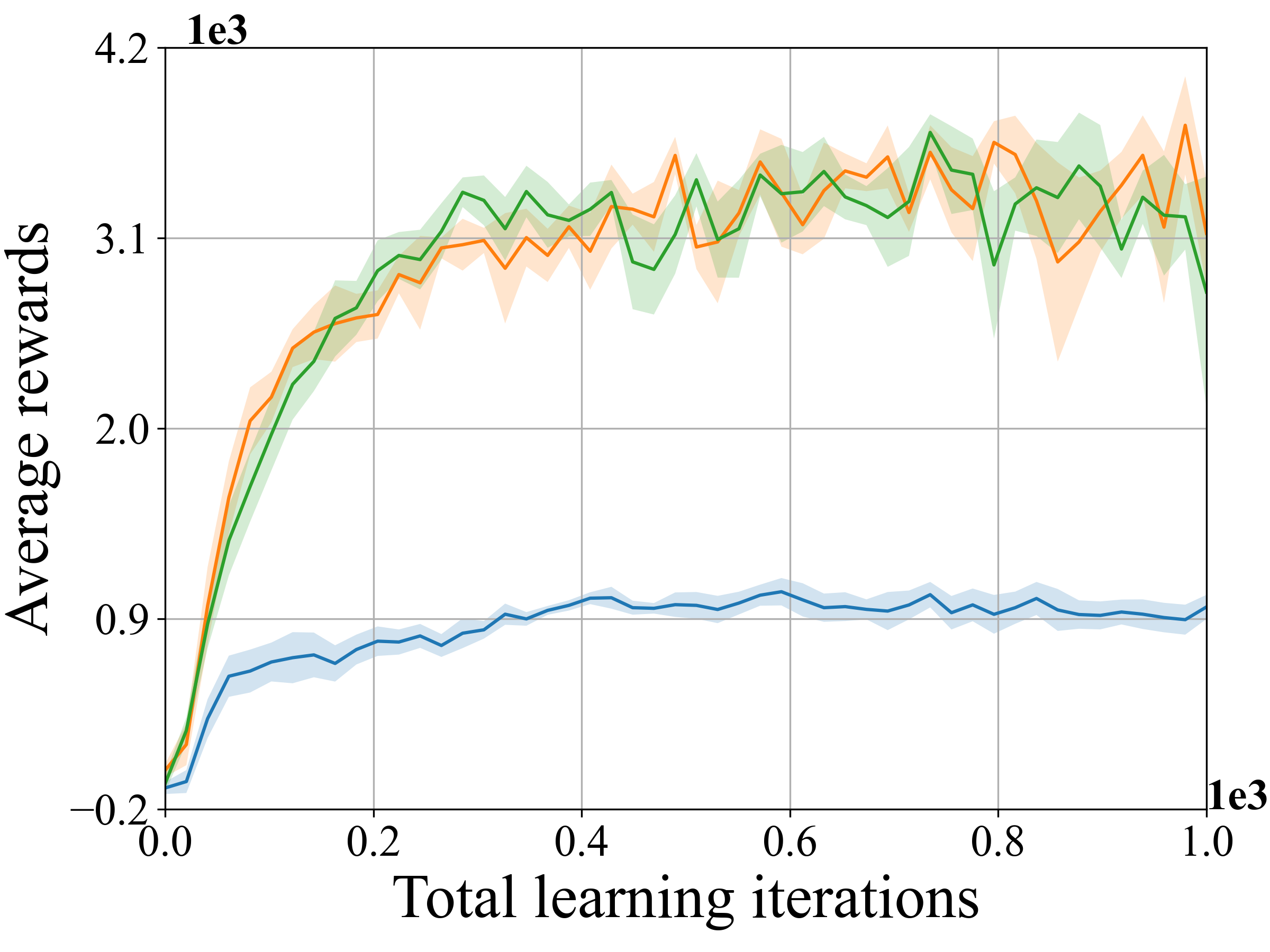}}
		\subfigure[Walker2d-v2]{\includegraphics[width=0.230\textwidth]{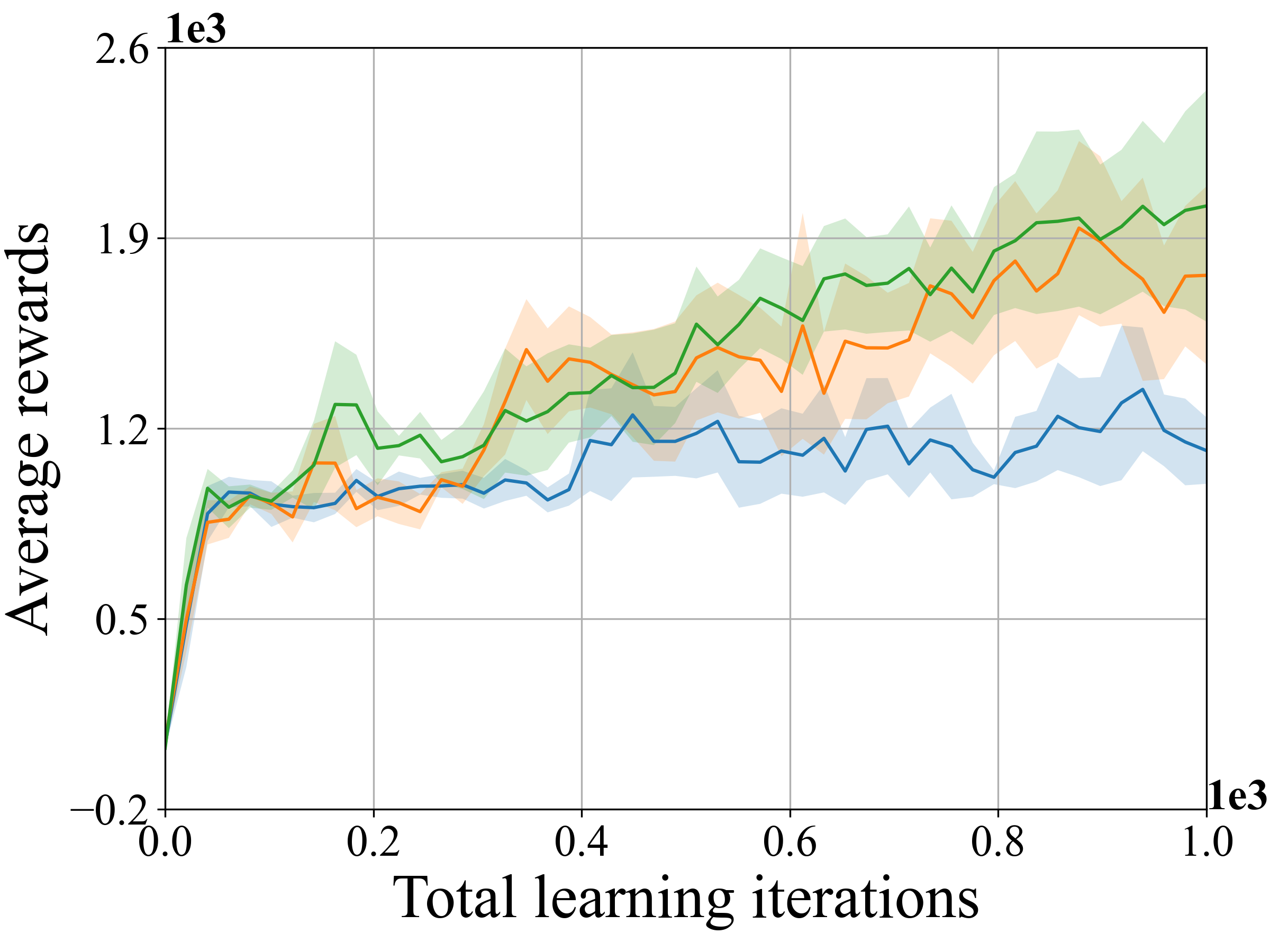}}
		\subfigure[LunarLanderContinuous-v2]{\includegraphics[width=0.230\textwidth]{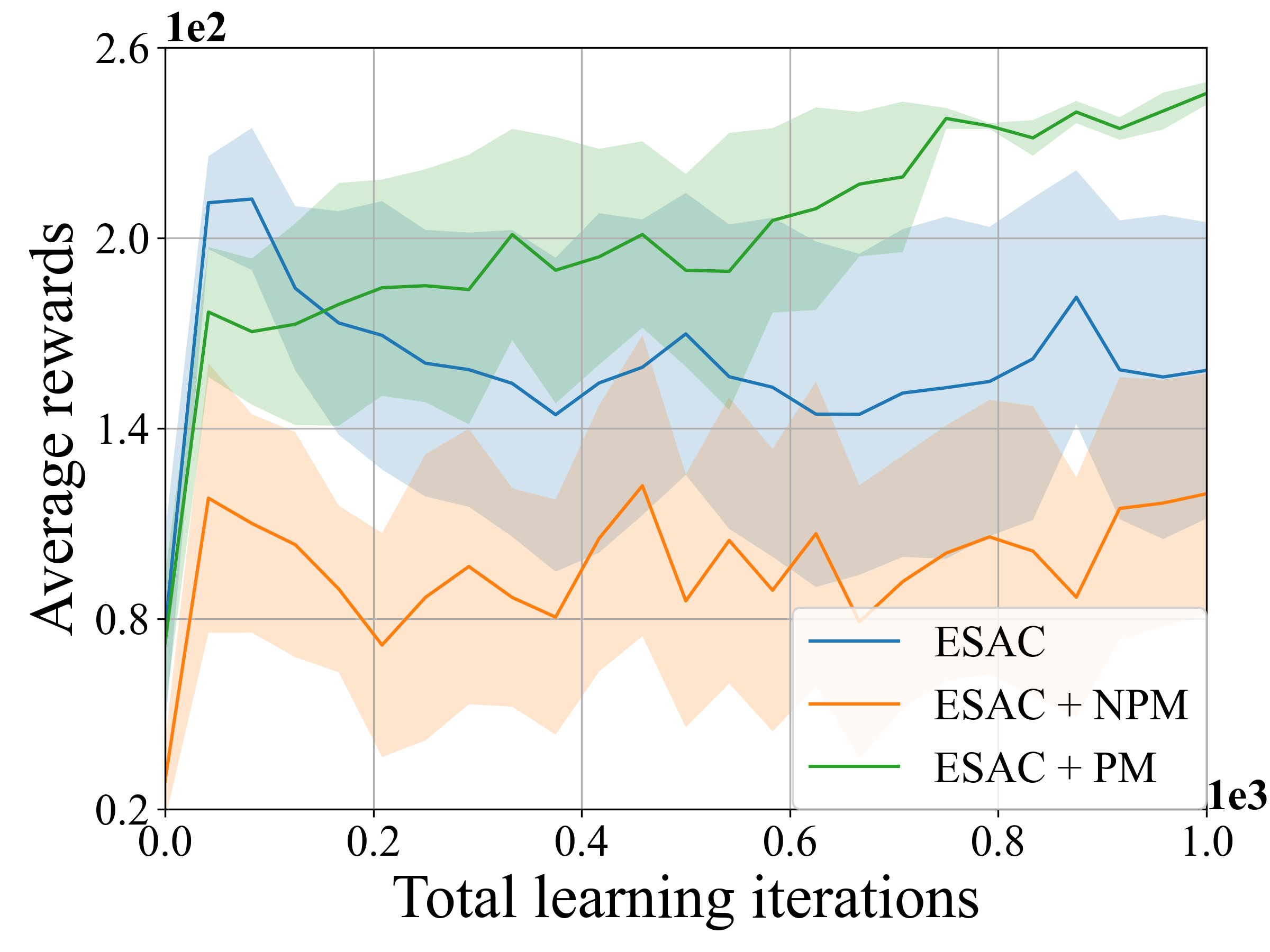}}
			\caption{Performance of different algorithms in three standard MuJoCo and one Box2D environments.
			Figures (a)-(d), (e)-(h), and (i)-(l) show the learning curves of Vanilla ES, NSR-ES, and ESAC by plugging our PM and NPM methods, respectively.
			All the curves are averaged over 5 different random seeds, and the standard deviation is shown as a shaded region.}
                \label{fig.4}
\end{figure*}

\begin{figure*}[tb]

\centering 
\subfigure[Ant-v2]{\includegraphics[width=0.230\textwidth]{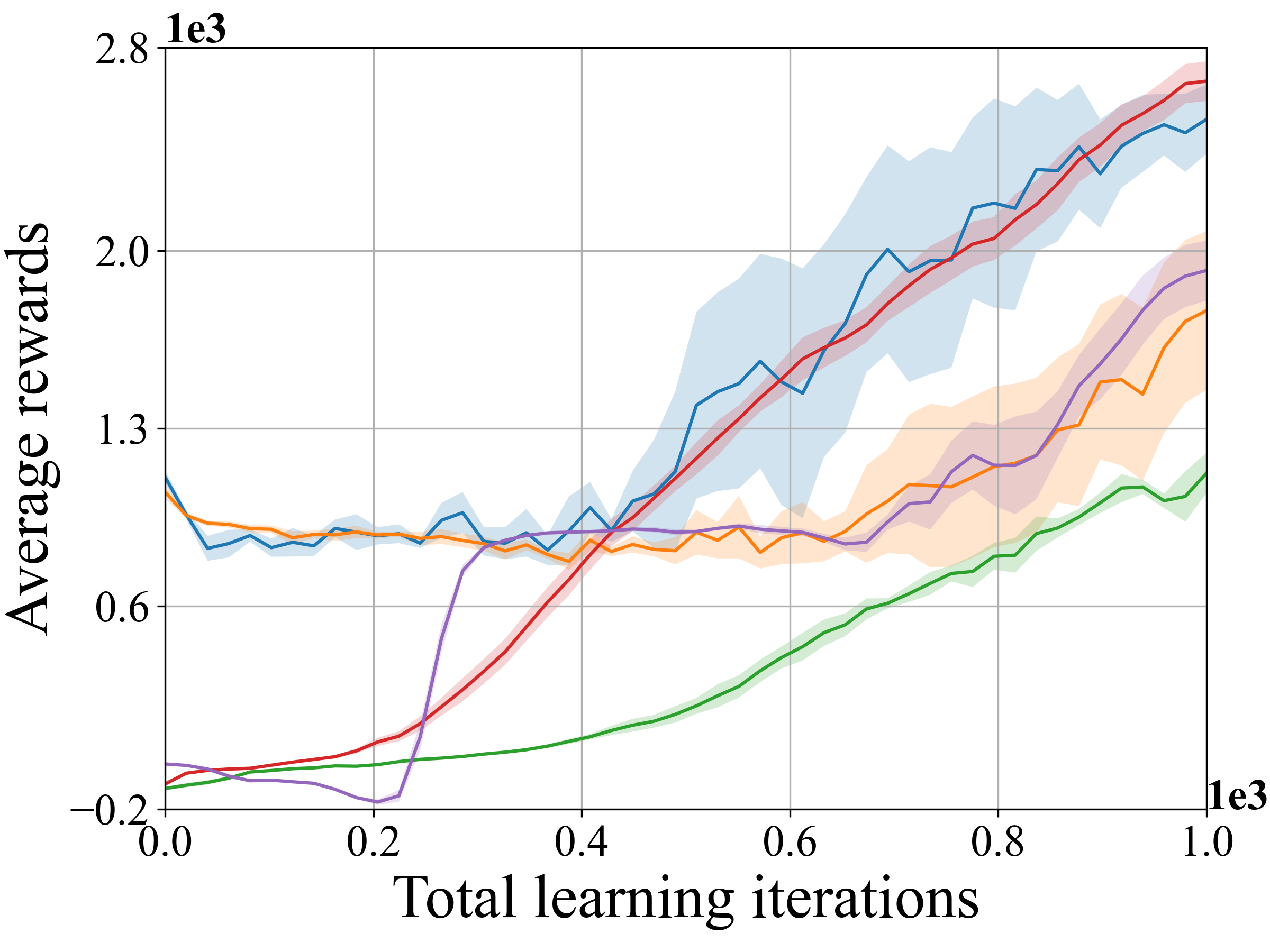}}
\subfigure[HalfCheetah-v2]{\includegraphics[width=0.230\textwidth]{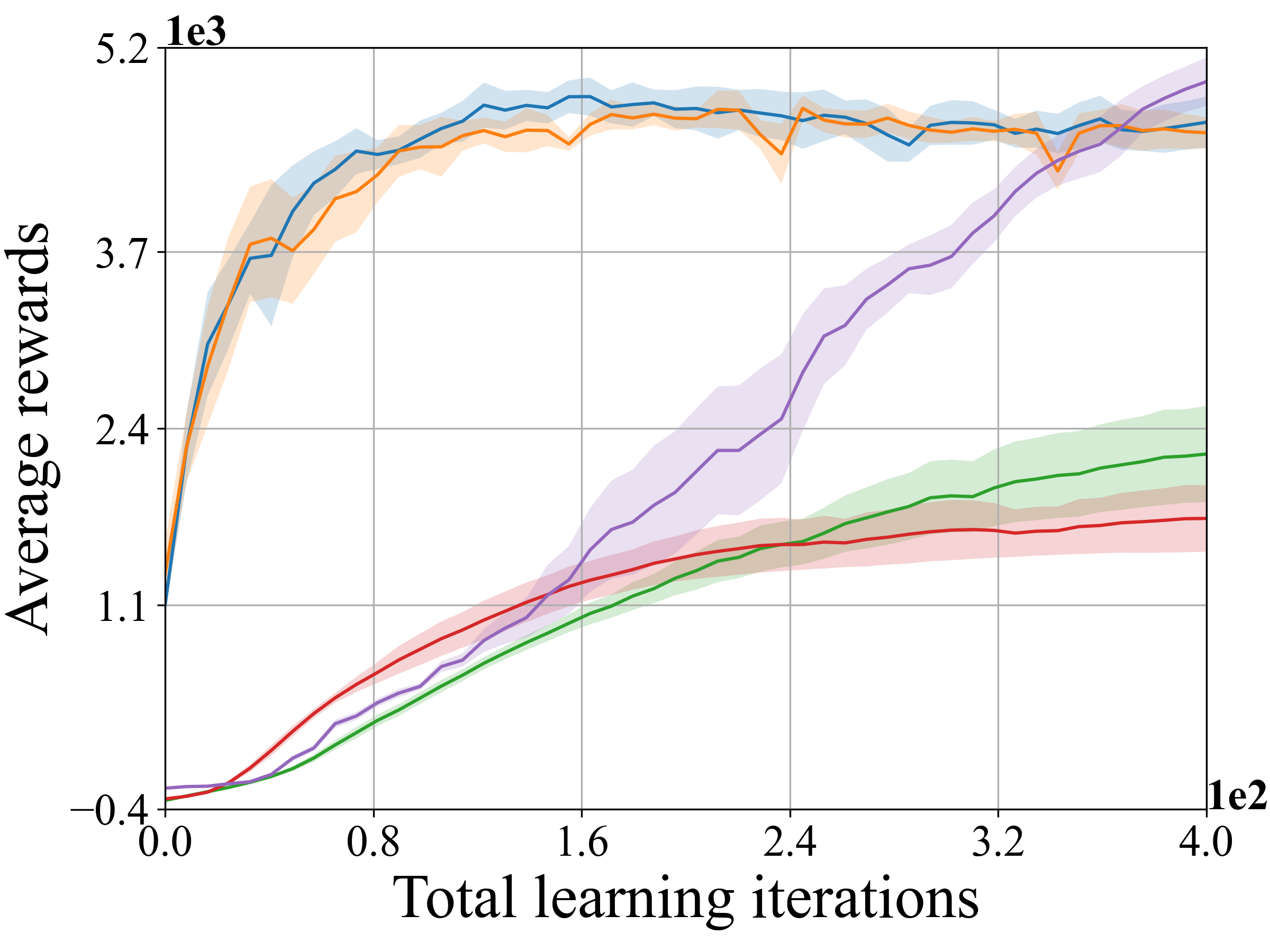}}
\subfigure[Walker2d-v2]{\includegraphics[width=0.230\textwidth]{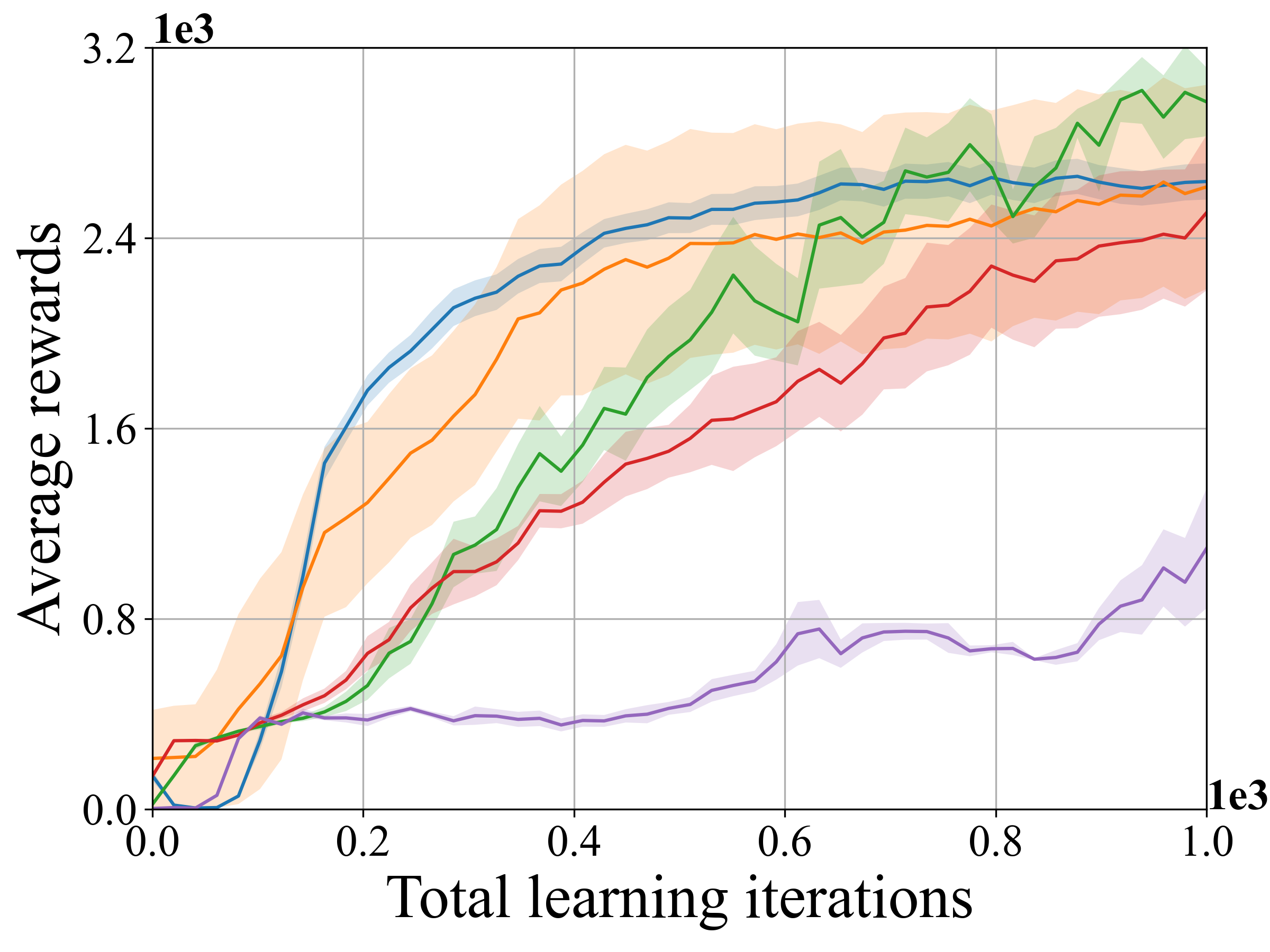}}
\subfigure[LunarLanderContinuous-v2]{\includegraphics[width=0.230\textwidth]{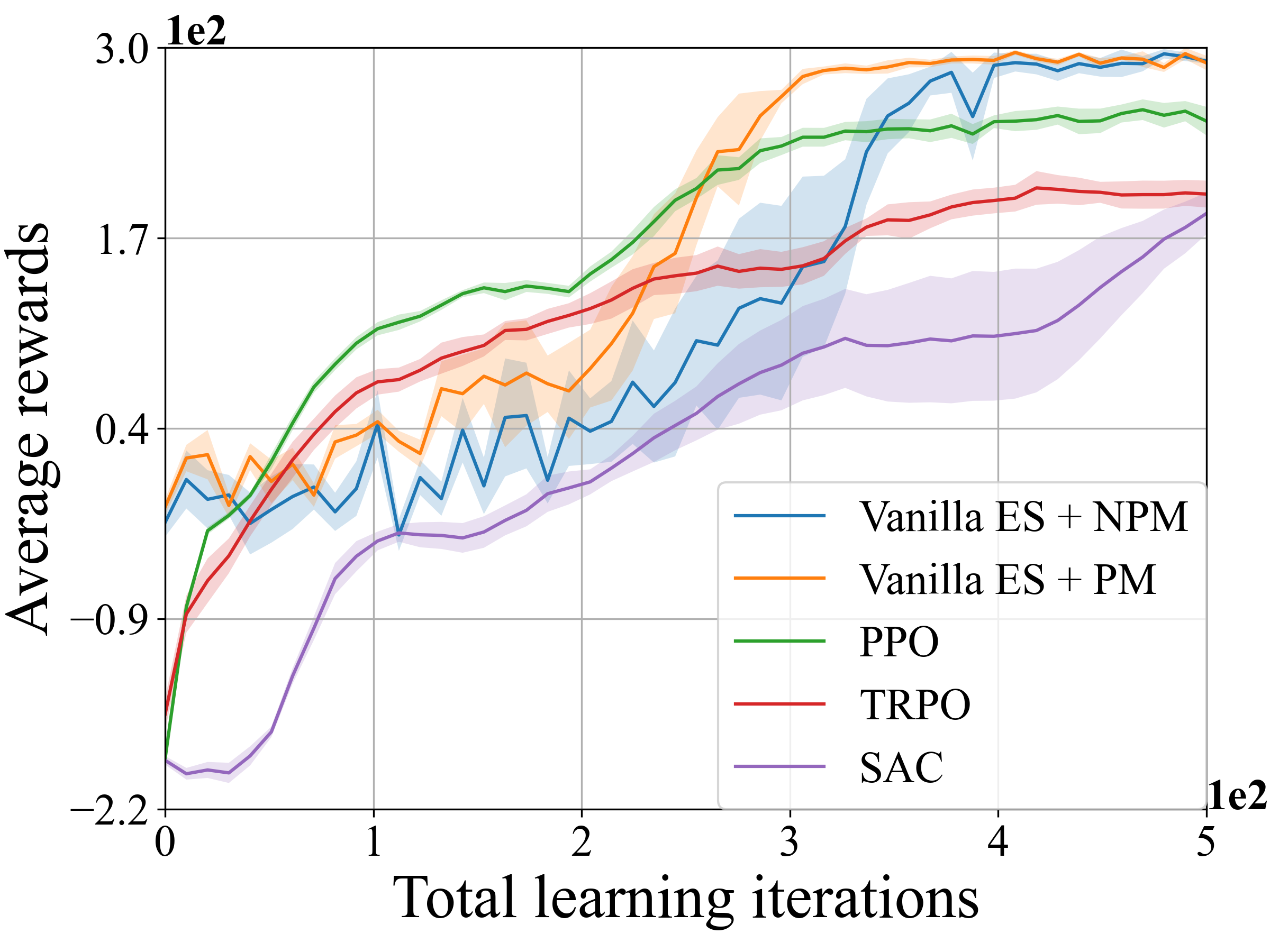}}
\caption{Performance of our method and state-of-the-art DRL methods in three standard MuJoCo and one Box2D environments.
All the curves are averaged over 5 different random seeds, and the standard deviation is shown as a shaded region.}
\label{fig.9}
\end{figure*}
  \subsection{Performance  (RQ1)}
  \label{sec5.1}

\begin{figure}[tb]
\centering 
\subfigure[Ant-v2]{\includegraphics[width=0.230\textwidth]{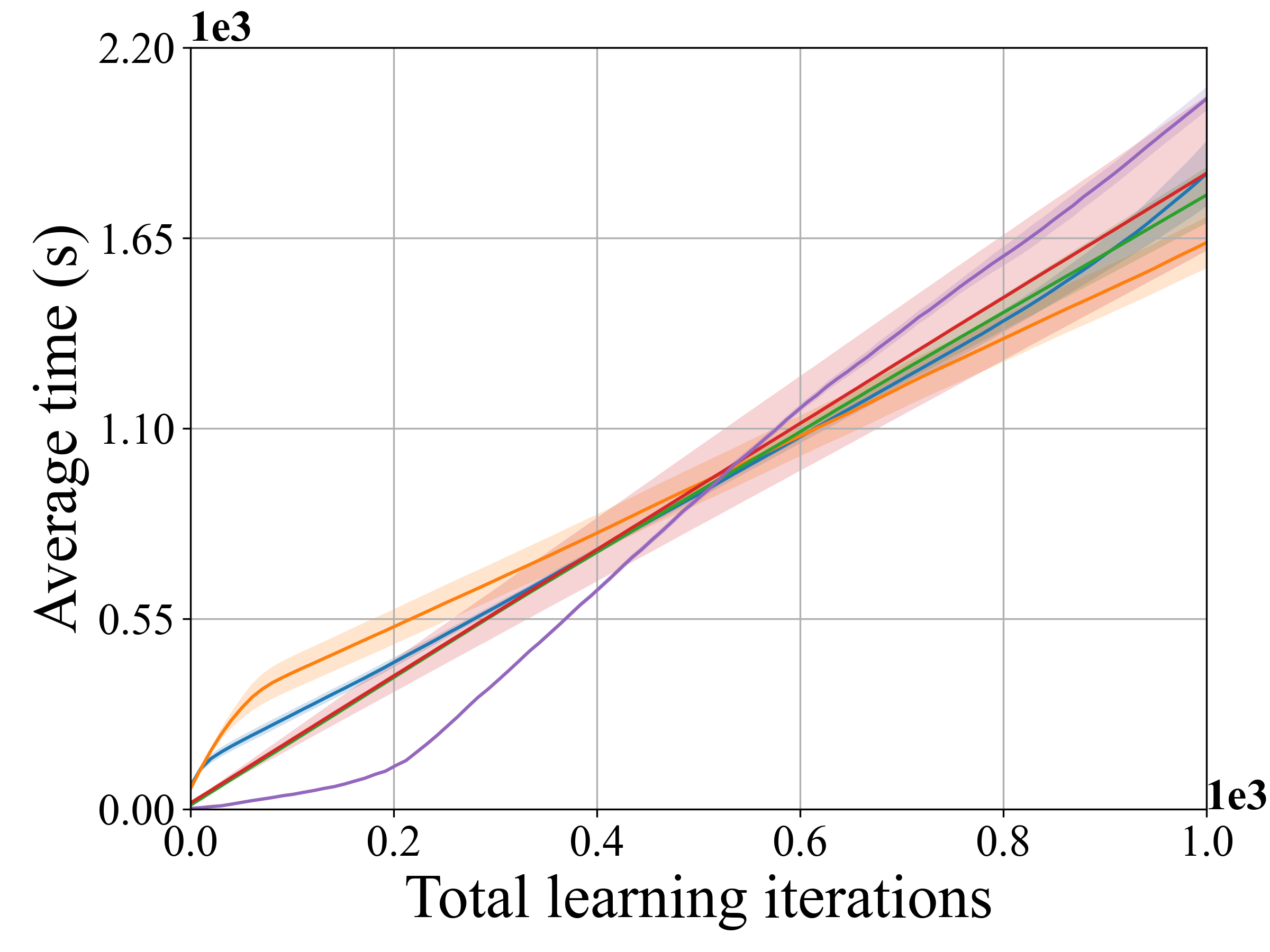}}
\subfigure[LunarLanderContinuous-v2]{\includegraphics[width=0.230\textwidth]{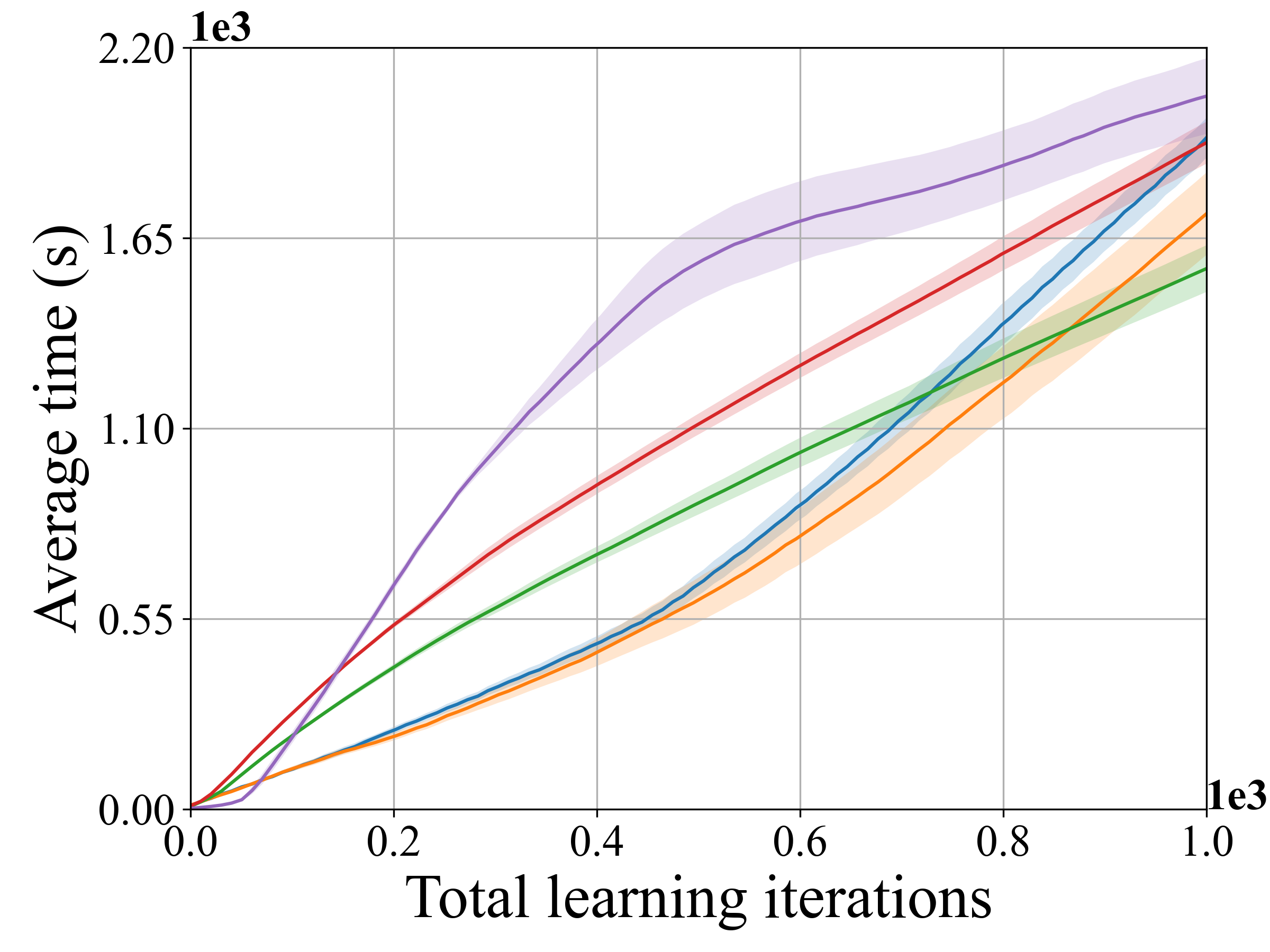}}
\caption{Running time regarding the learning iterations of BiERL, Vanilla ES, and DRL methods evaluated on the two example tasks.}
\label{fig.10}
\end{figure}

To demonstrate and illustrate the outperformance of our method, we compare BiERL with three representative ERL methods including Vanilla ES~\cite{salimans2017evolution}, NSR-ES~\cite{conti2018improving}, and ESAC~\cite{suri2022off} on a total of $6$ MuJoCo~\cite{todorov2012mujoco} and $2$ Box2D tasks in the OpenAI Gym~\cite{brockman2016gym}.
We apply our BiERL architecture to the three baselines as the meta-level optimizer with the parametric model  (PM) and nonparametric model (NPM) for optimizing the noise covariance $\sigma$, respectively.
Each algorithm runs with five random seeds and is reported in mean $\pm$ standard deviation.
Further implementation details are given in Appendix.

The comparisons across BiERL, Vanilla ES, NSR-ES, and ESAC baselines are shown in Figure~\ref{fig.4}.
We make a conclusion from the results: BiERL $>$ NSR-ES $>$ Vanilla ES $>$ ESAC\footnote{Here, we use the binary comparison operators to indicate the performance order of these algorithms.}.
Specifically, (1) applying the PM and NPM to three baselines can significantly improve the performance, and lead to high learning efficiency, which validates the effectiveness of BiERL.
It implies that adaptive hyperparameters for changing the noise covariance $\sigma$ can help the agent learn more efficiently;
(2) NSR-ES obtains better performance than Vanilla ES and ESAC by plugging the PM or NPM of BiERL. 
The reason may be that NSR-ES is able to avoid local optima encountered by combining exploration with reward maximization;
(3) Vanilla ES achieves comparable performance with the ESAC across all the tasks, which implies that there are potentials for rapid improvements since Vanilla ES reproduces one policy in each iteration and may update the same policy for several iterations;
(4) plugging PM into baselines achieves more gains over NPM most of the time, which validates that PM could capture long-term dependencies from the population representation using the advance of LSTM;
(5) BiERL can achieve more statistically significant advantages in complex tasks  (e.g., Ant-v2 and HalfCheetah-v2) than the simple ones (e.g., Walker2d-v2 and LunarLanderContinuoous-v2) across all baselines, which shows the superiority of BiERL in challenging tasks.
It should benefit from the meta-level generating adaptive hyperparameters, leading to adequate exploration to avoid local optima and discover a better policy.

Since our focus is a general framework that realizes efficient meta-parameter optimization for general ERL algorithms, we mainly investigate the performance improvement of deploying BiERL on a diversity of typical ERL algorithms.
In addition, we also present experimental results with typical state-of-the-art DRL methods, including TRPO, PPO, and SAC, as is shown in Figure~\ref{fig.9}.
It can be observed that our method can outperform the state-of-the-art DRL methods in these tasks, especially at the beginning of the training process, which further validates the effectiveness of adapting the hyperparameters during learning and the efficiency of our method.


Next, Figure~\ref{fig.10} shows the detailed running time regarding the learning iterations of BiERL, Vanilla ES, and DRL methods evaluated on the example Ant-v2 and LunarLanderContinuous-v2 tasks.
It can be observed that the required running time of BiERL is roughly at the same scale as both baselines of ES and DRL methods.
Owing to the simple and feasible approximation of the meta-level evaluation and the parallelizability of ERL methods, the general framework of BiERL does not significantly increase the time complexity of ERL algorithms, while it is capable of consistently improving the learning performance for various ERL algorithms.
In addition to performance comparison, high parallelization is another promising advantage of ERL methods, including BiERL.
We can use distributed computing environments to further reduce the computation time, and obtain greater superiority compared to DRL methods.


In summary, BiERL can consistently improve the learning performance of various ERL methods with little increase in the computation time.
More experiments conducted on other MuJoCo and Box2D tasks in OpenAI Gym are included in the Appendix. 

\begin{figure}[tb]
			\centering 
		\subfigure[HalfCheetah (Performance)]{\includegraphics[width=0.230\textwidth]{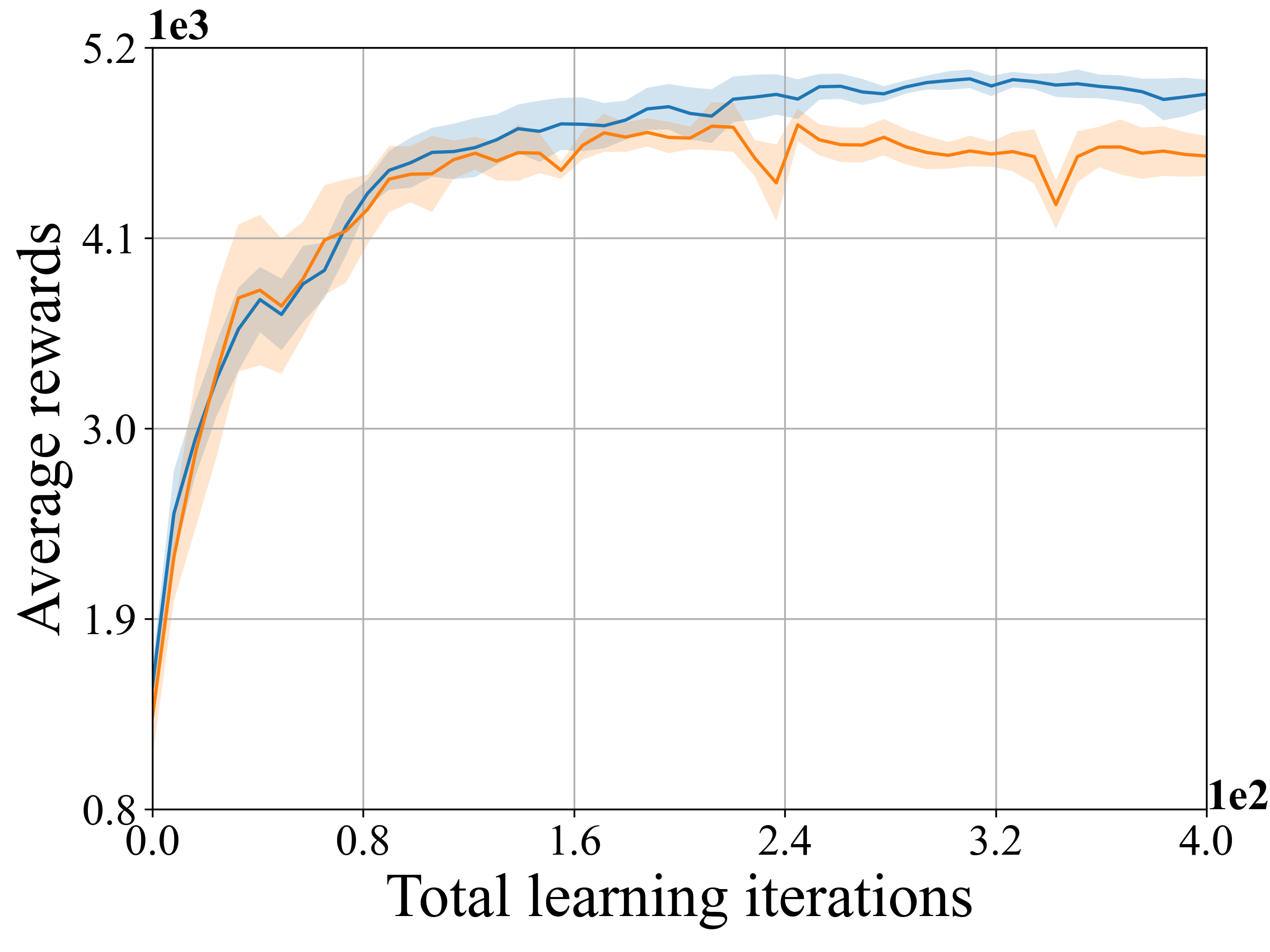}}
		\subfigure[Walker2d (Performance)]{\includegraphics[width=0.230\textwidth]{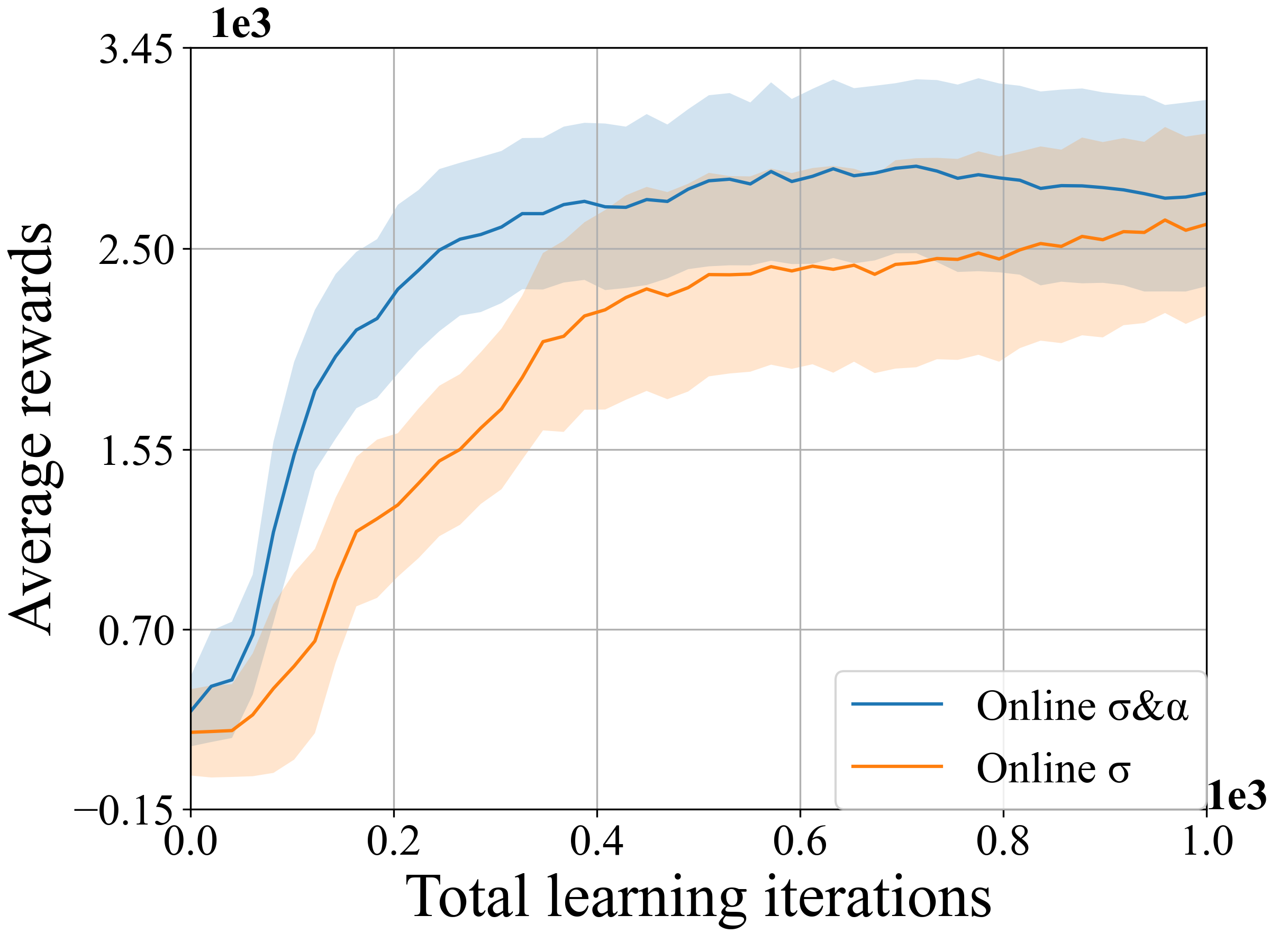}}
j		\subfigure[HalfCheetah ($\sigma$\&$\alpha$)]{\includegraphics[width=0.230\textwidth]{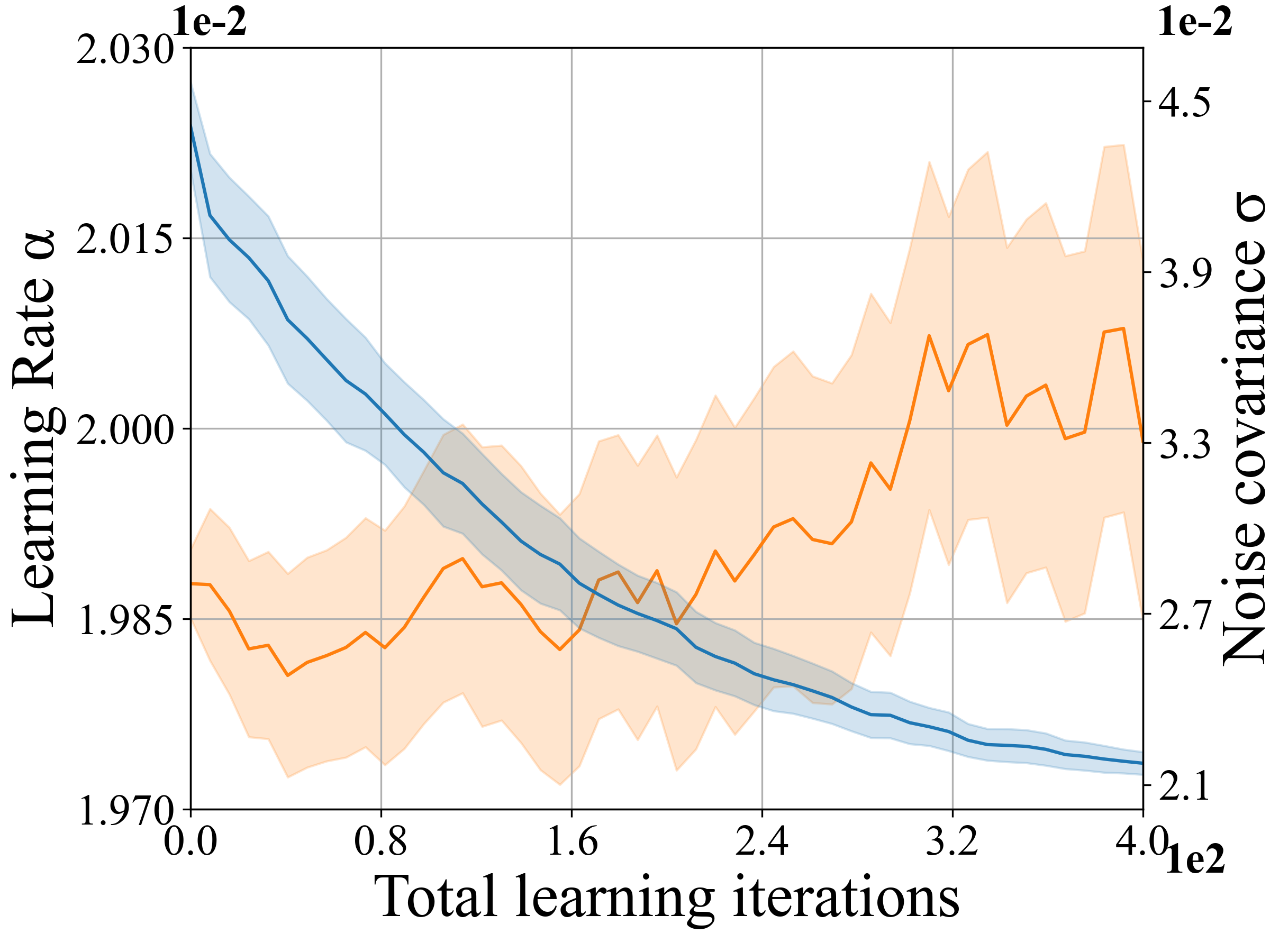}}
		\subfigure[Walker2d ($\sigma$\&$\alpha$)]{\includegraphics[width=0.230\textwidth]{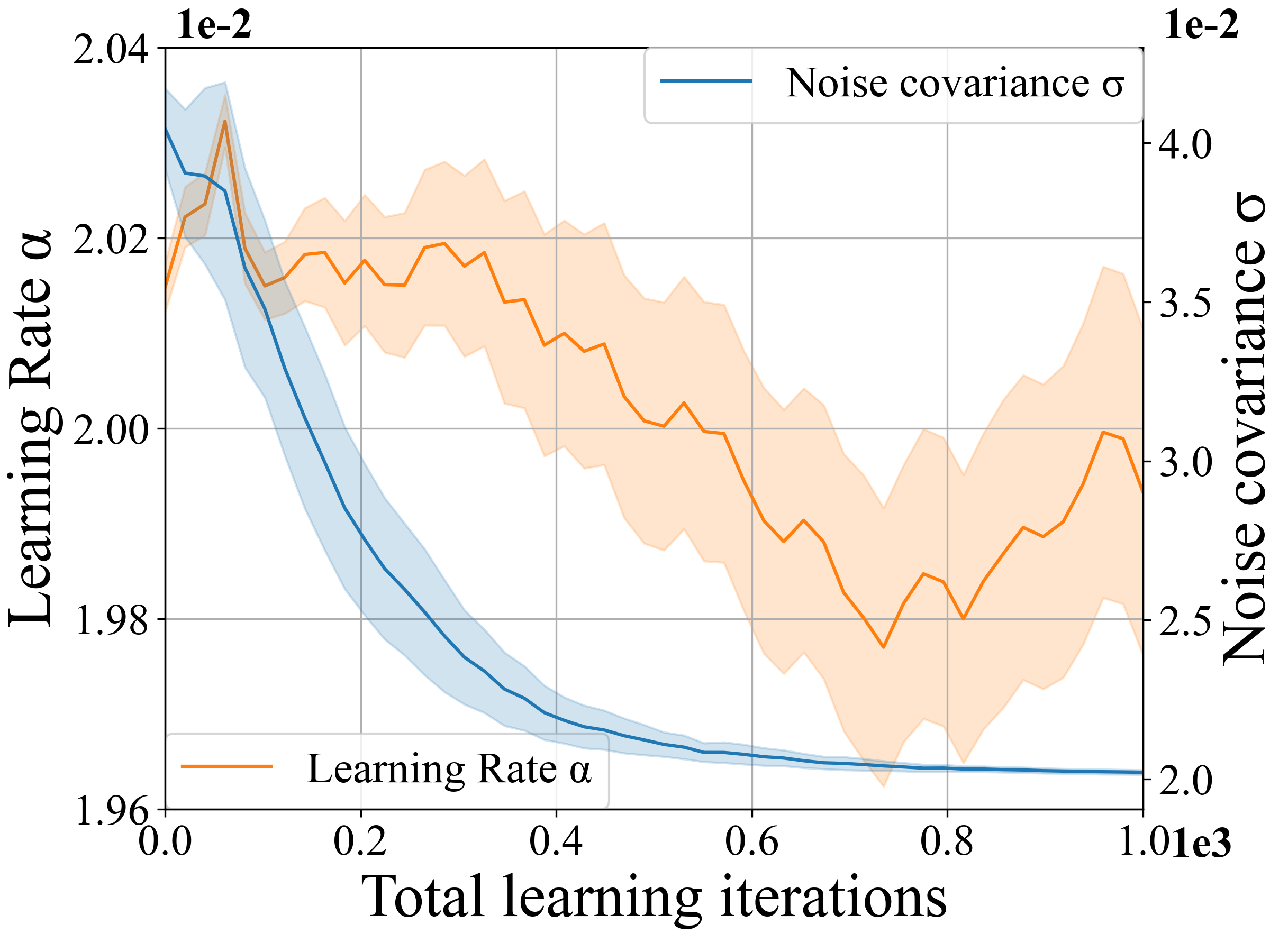}}
			\caption{Scalability analysis on the learning rate $\alpha$ for BiERL.  (a)-(b) show learning curves, and (c)-(d) show the numerical values of hyperparameters during learning.}
                \label{fig.5}
\end{figure}

\subsection{Scalability (RQ2)}
\label{sec5.2}
To answer RQ2, we assess the scalability of BiERL.
We extend the learning rate $\alpha$ as joint optimization hyperparameters by the meta-level to study whether BiERL can consistently produce high performance with a better mechanism of adaptive learning rate.
To ensure the positivity of the learning rate, we clip the parameterization $\alpha\in[0.016,0.024]$ and denote it as online $\sigma\&\alpha$.

 Figure~\ref{fig.5} presents the scalable nature of BiERL.
 From Figure~\ref{fig.5}-(a) and -(b), by comparison with fixed learning rate $\alpha$, we find that the online $\sigma\&\alpha$ method obtains improved performance for both tasks. 
 It implies that our BiERL can provide prominent performance boosts and scalability.
 Further, we study the variation of the learning rate and noise covariance during the learning process.
 As shown in Figure~\ref{fig.5}-(c) and~\ref{fig.5}-(d), noise covariance $\sigma$ becomes small as the policy tends to converge to the robust policy during the learning process.
 It is required that a smaller noise covariance $\sigma$ can produce more stable behaviors, which can reduce catastrophic outcomes with less randomized behaviors.
 It can also be observed that the learning rate $\alpha$ is generated in a dynamic range for each inner-loop step.
 An interesting behavior to note is that the ranges of generated hyperparameter values are small.
 We believe that such a phenomenon could be owed that the learning rate $\alpha$ allows the model to focus on learning stability.

\begin{figure}[tb]
\centering 
\subfigure[HalfCheetah ($n$)]{\includegraphics[width=0.230\textwidth]{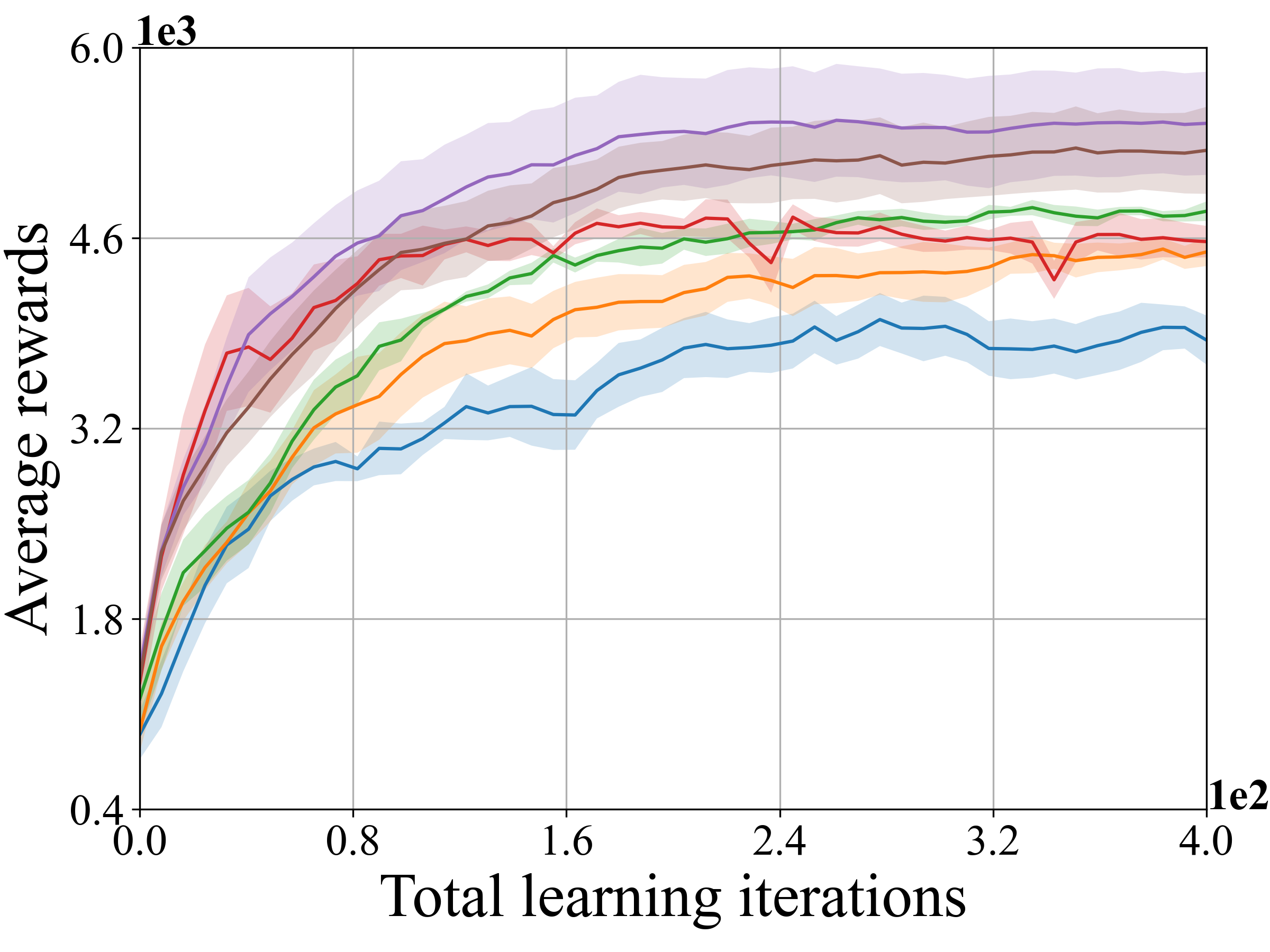}}
\subfigure[Walker2d ($n$)]{\includegraphics[width=0.230\textwidth]{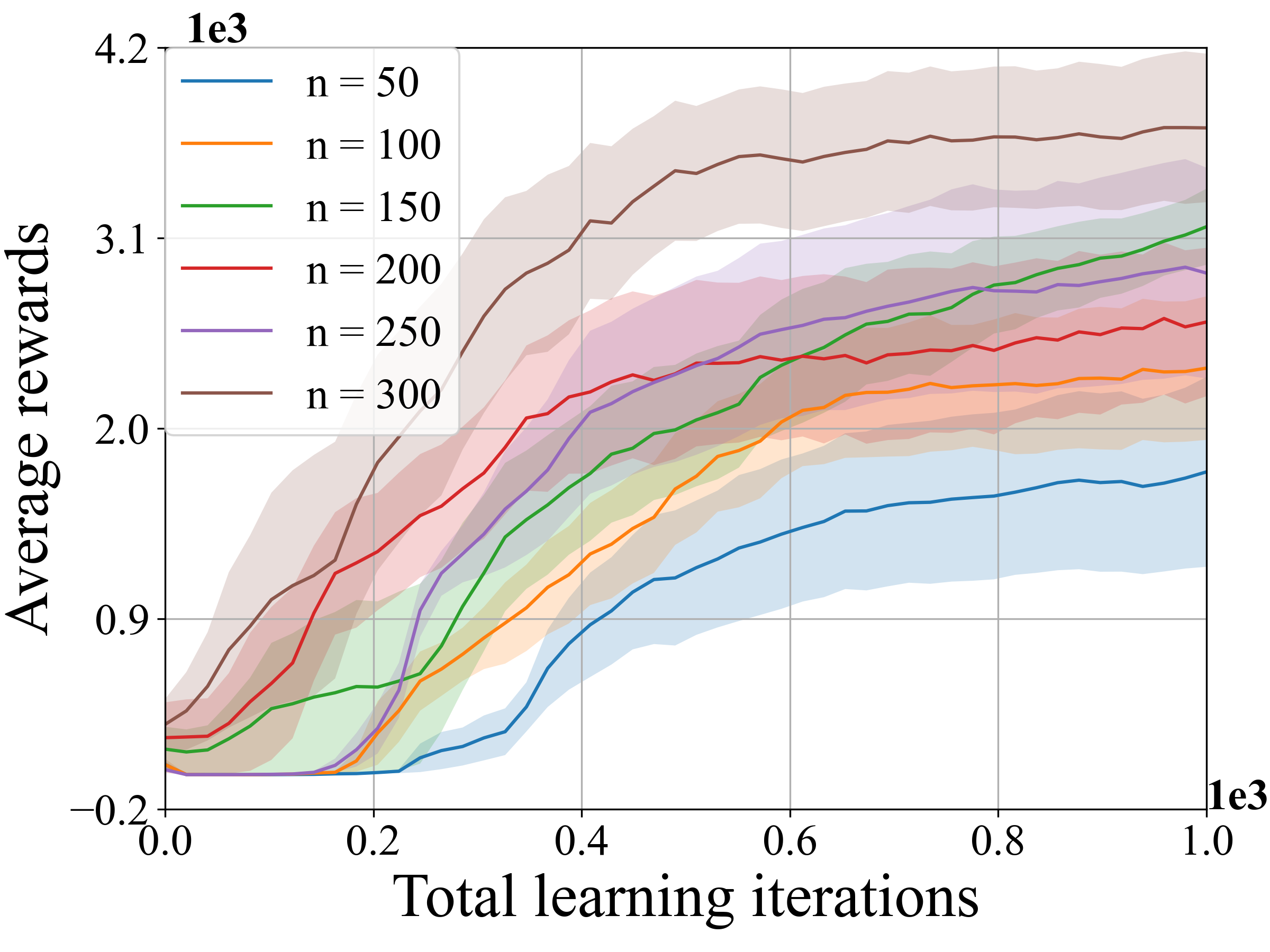}}
\subfigure[HalfCheetah ($m$)]{\includegraphics[width=0.230\textwidth]{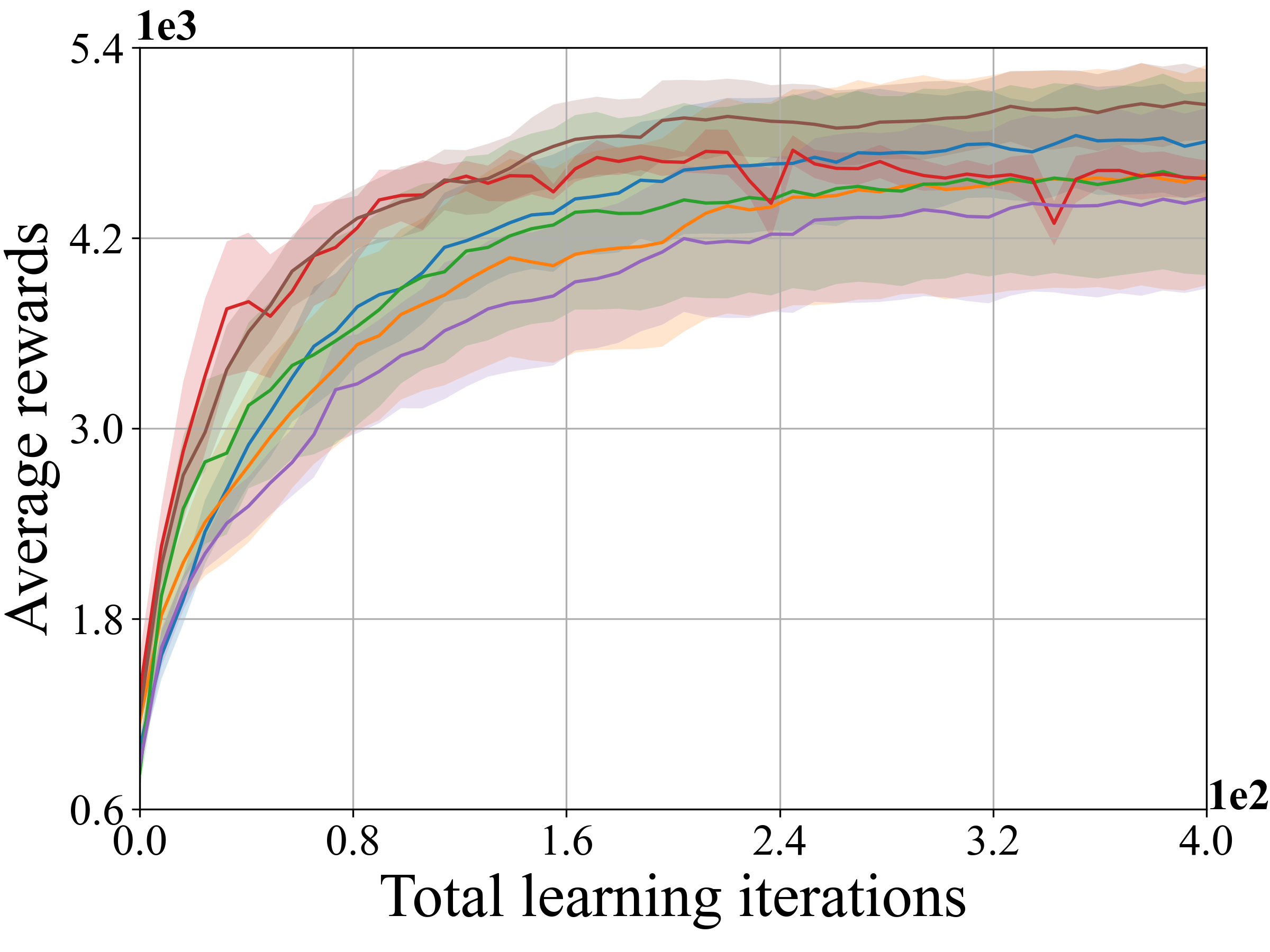}}
\subfigure[Walker2d ($m$)]{\includegraphics[width=0.230\textwidth]{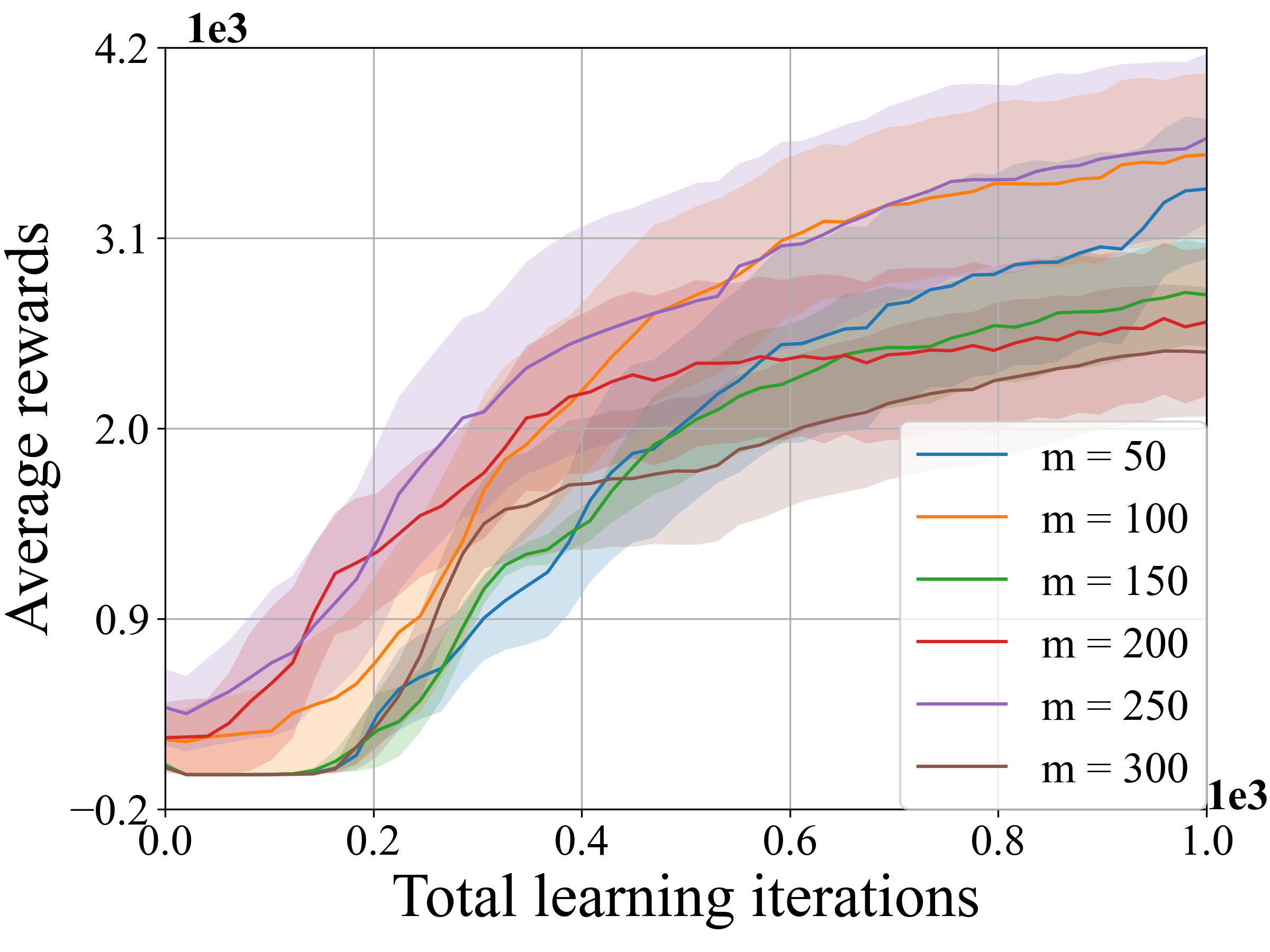}}
\caption{Performance of BiERL with different population sizes $n$ or $m$, given the same number of iterations.}
\label{fig.7}
\end{figure}

 Moreover, we study the influence of another two hyperparameters, the population size of the inner-level $n$ and the meta-level $m$.
 These experimental results are provided in Figure~\ref{fig.7}. We find that a larger population size tends to achieve high performance.
 However, when the population size is too large, some policies with similar behavior will be selected and updated, harming the efficiency of BiERL.
 Thus, an appropriate population size (e.g., $150$ or $200$) can better trade off between performance and efficiency.  
 
  \begin{figure}[tb]
			\centering 
		\subfigure[HalfCheetah]{\includegraphics[width=0.230\textwidth]{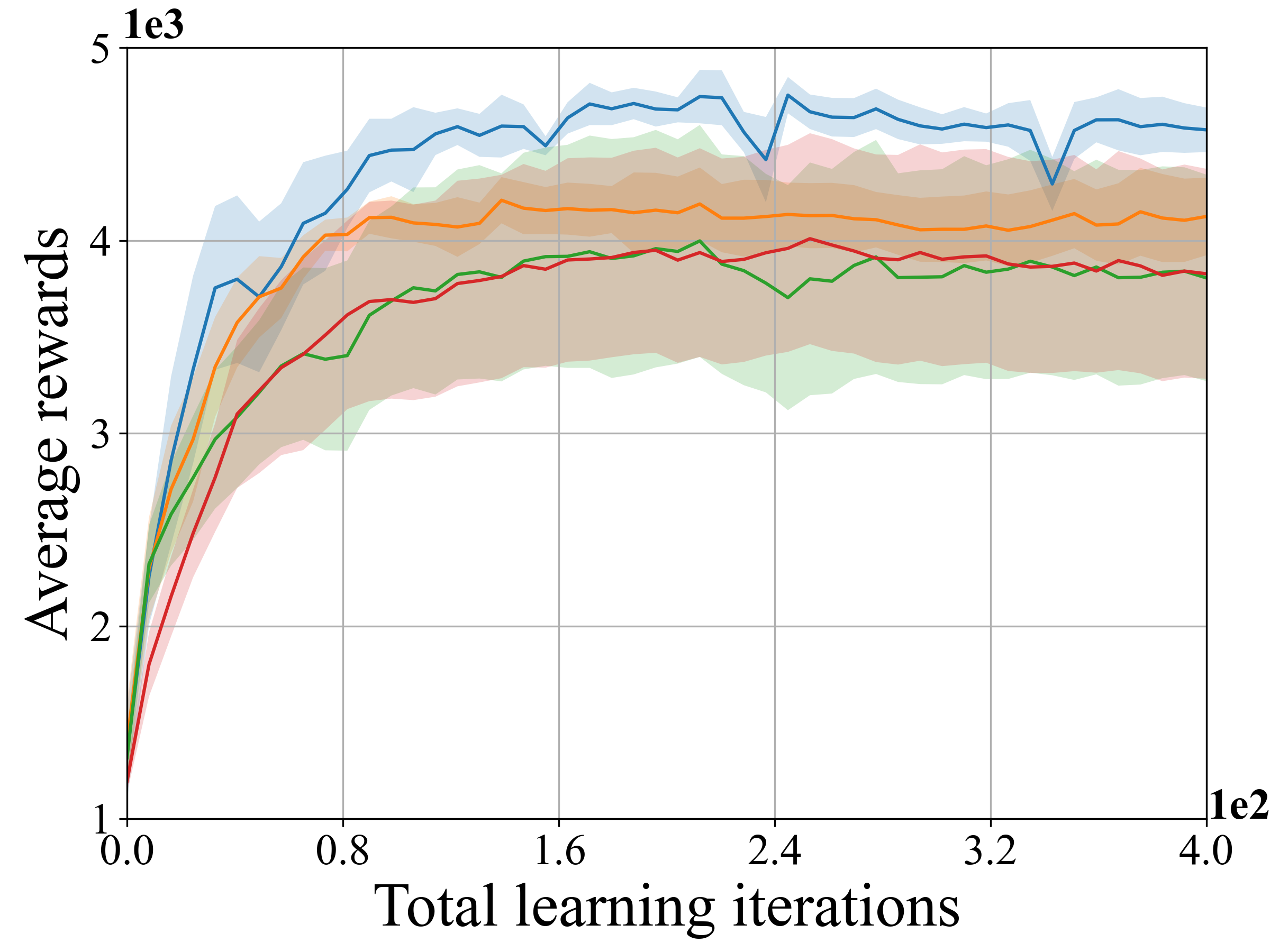}}
		\subfigure[Walker2d]{\includegraphics[width=0.230\textwidth]{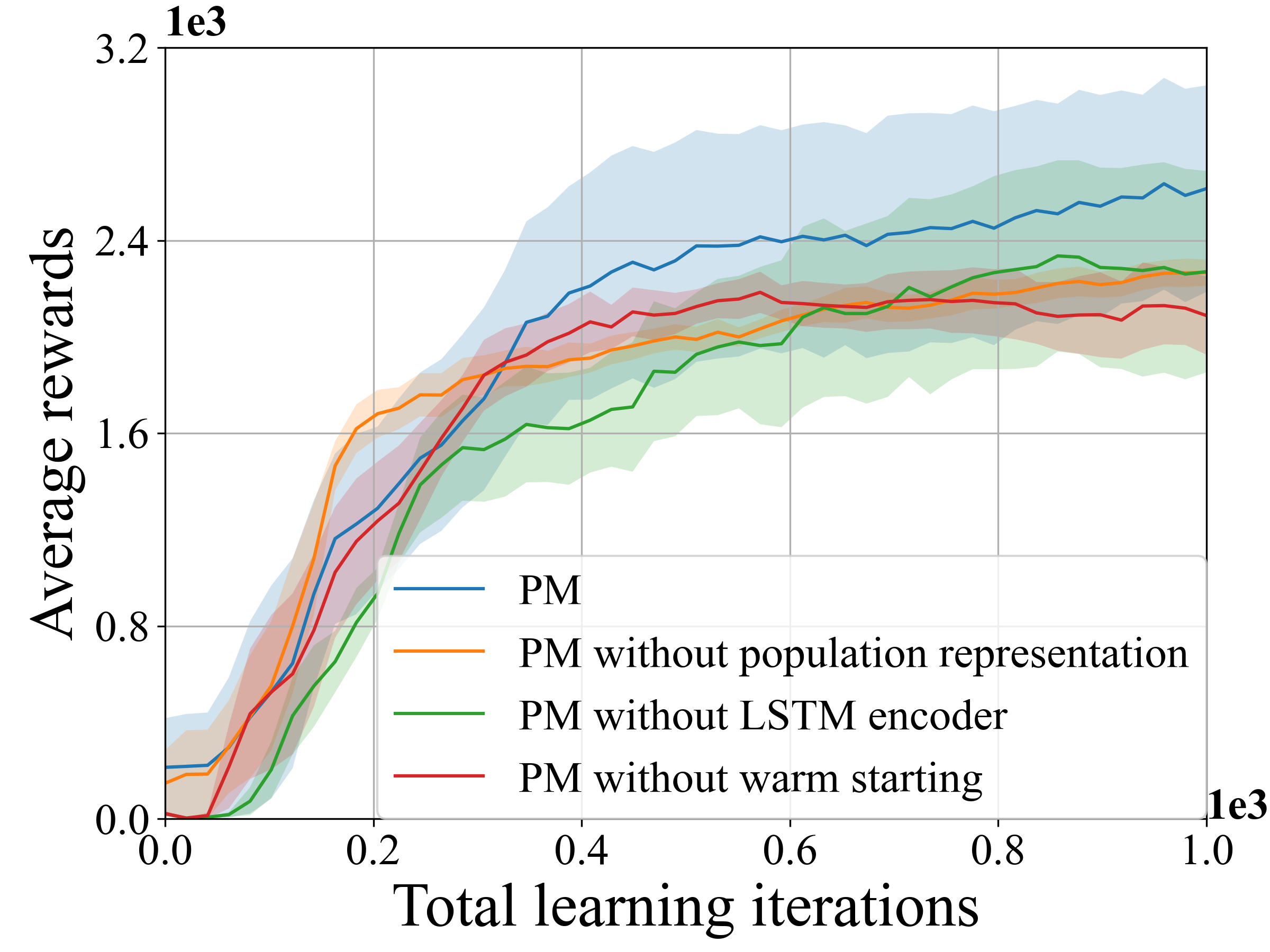}}

			\caption{Ablation experiments where the population representation, LSTM encoder, and warm starting are removed separately.
   }
                \label{fig.6}
\end{figure}
  \subsection{Ablation study (RQ3)}
  To investigate the role of each part of the parametric meta-level, we perform a series of ablation studies including (1) \textbf{PM without population representation} represents that the input of the meta-level solely considers the unperturbed parameters $f (\theta_t)$ instead of population information;
  (2) \textbf{PM without LSTM encoder} represents that the meta-level solely removes the LSTM encoder;
  (3) \textbf{PM without warm starting} represents that randomly initializing the meta-level instead of using the warm starting mechanism.
  
 Figure~\ref{fig.6} presents the ablation results of the three parts.
 The parametric model conditioned on each part of BiERL still exhibits performance improvement over both tasks, suggesting that all parts play important roles.
While PM without the population representation learns faster during the beginning learning stage, it finally obtains lower returns, which could be fallen into the deceptive trap without sharing information with the exploration in the parameter space.
When removing the LSTM encoder module leads to a degradation in learning performance across both tasks, it implies that the LSTM encoder module can help the agent efficiently capture the history information and obtain better performance.
We also observe that PM without warm starting learns inefficiently at the early stage, yet its overall performance is slightly worse than PM.
This result demonstrates the importance of a warm starting in complex domains, where pretraining in a simple task only yields successful results for finetuning in the complex target task.
In summary, our proposed PM model conditioned on all parts gives the best performance, indicating that population representation, the LSTM encoder, and the warm starting parts are complementary parts of BiERL.

\section{Conclusion }
\label{sec6}

We propose a general BiERL framework to realize efficient meta-parameter optimization without prior domain knowledge or costly optimization procedure.
We introduce an elegant meta-level architecture that efficiently adjusts hyperparameters according to the informative representation of the inner-level's evolving experience, and we design a simple and feasible evaluation of the meta-level fitness function to facilitate learning efficiency.
Empirically, our method provides significant performance gains over representative baselines on a range of RL tasks.
We believe that our work can initiate several interesting future research directions. 
For instance, one can explore different types of meta-level architecture, other than the LSTM and MLP networks used here. 
Another interesting future work could be combining BiERL with representation learning in behavior space and analyzing the performance of BiERL theoretically.

\section*{Acknowledgements}
The work was supported in part by the National Natural Science Foundation of China under Grant 62006111, Grant 62073160, and Grant 62106172, and in part by the National Key R\&D Program of China under Grant 2022ZD0116402.

\bibliography{ecai}

\clearpage
\appendix

\onecolumn

\setlength{\baselineskip}{2em}

\noindent \vspace{0.5em} \rule{\textwidth}{0.15em}
\centerline{\LARGE \textbf{Supplementary Materials}} \\
\centerline{\LARGE \textbf{BiERL: A Meta Evolutionary Reinforcement Learning Framework}} \\
\centerline{\LARGE \textbf{via Bilevel Optimization}}
\vspace{0.5em} \rule{\textwidth}{0.15em}

\vspace{1em}
\centerline{\large \textbf{Junyi Wang$^{1}$, Yuanyang Zhu$^{1}$, Zhi Wang$^{1}$, Yan Zheng$^2$, Jianye Hao$^2$, and Chunlin Chen$^1$}}

\centerline{\textbf{$^1$Department of Control Science and Intelligent Engineering, Nanjing University}}
\centerline{\textbf{$^2$College of Intelligence and Computing, Tianjin University}}

\setlength{\baselineskip}{1.5em}

\vspace{1em}
\section{Experimental Configuration}
\label{appendix1}
We adopt the Vanilla ES framework~[6] to implement our method and the other two baselines, NSR-ES~[1] and ESAC~[7].
For a fair comparison with the ES algorithms, we implement the inner-level of BiERL as well as its baselines with the architectures utilized in Vanilla ES~[6]:
an MLP with two 64-unit hidden layers separated by tanh nonlinearities. 
The hyperparameters of the inner-level of BiERL are the same as the baselines.
The input is the current observation while the output is the action distribution.

For the meta-level architecture,  we use an LSTM encoder and an MLP generator containing one hidden layer. A Sigmoid activation is applied to the output layer of the generator. The dimension of the output depends on the number of adaptive hyperparameters.  
We set the maximum episode length to 1000. We train the inner-level model for 400 to 1000 iterations and use a time interval of $k=10$ for updating the meta-level, while the training time of each task is about 2 to 5 hours, depending on the different meta-level optimizers and the difficulty of the target task.
The experiment configuration and the network architecture of our method are illustrated in Tables \ref{tab.1}, \ref{tab.2}, and \ref{tab.3}, respectively.

We implement all the methods with PyTorch 1.12.1 framework in Python 3.9 running on Ubuntu 18.04 with 2 AMD EPYC 7H12 64-Core CPU Processors and 1 NVIDIA GeForce RTX 3080 GPU. The experimental tasks are from Gym 0.15.7~[2] and MuJoCo 2.0~[8].

\begin{table}
	\renewcommand\arraystretch{1.5}
	\centering
	\begin{tabular}{c|c|c|c|c|c}
		
		\hline 
		\multicolumn{3}{c|}{Inner-level}    &   \multicolumn{3}{c}{Meta-level} \\ 
		\hline 
		Hyperparameters   & Notation & Value & Hyperparameters   & Notation & Value\\ 
		\hline 
		Optimizer        &   -    & SGD       & Optimizer        &   -    & SGD         \\
		Maximum Episode Length  & - & 1000   &  Warm Starting Iterations  & -  & 10\\
		Decay Factor     & $\gamma$ & 1  & Interval& $k$      & 10\\       
		
		Learning Rate    & $\alpha$ & 0.02    & Learning Rate    & $\beta$  & 0.006    \\
		Noise Covariance & $\sigma$ & 0.05 & Noise Covariance & $\omega$ & 0.05  \\
		Population Size  & $n$      & 200   & Population Size  & $m$      & 200  \\
		
		\hline

	\end{tabular}

	\caption{The experiment configuration of BiERL.}
	\label{tab.1}
\end{table}


\begin{minipage}[c]{0.4\textwidth}
	\centering
	\renewcommand\arraystretch{1.5}
	\begin{tabular}{c|cc|c}
		\hline
		& \multicolumn{2}{c|}{MLP} &\\  \hline
		Input & Hidden \#1  & Hidden \#2  & Output\\ \hline
		$\lvert \mathcal{S} \rvert$ & 64        & 64    & $\lvert \mathcal{A} \rvert$   \\
		\hline
	\end{tabular}
	\captionof{table}{The inner-level's architecture.}
	\label{tab.2}
\end{minipage}
\begin{minipage}[c]{0.6\textwidth}
	\centering
	\renewcommand\arraystretch{1.5}
	\begin{tabular}{c|cc|c|c}
		\hline
		& \multicolumn{2}{c|}{LSTM Encoder $\psi$} & {MLP Generator $\phi$} &\\  \hline
		Input      &  Sequence Length    & Hidden  & Hidden  & Output \\ \hline
		$n$          & $k$            & 1024        & 32         & $\lvert \mathcal{H} \rvert$  \\
		\hline
	\end{tabular}
	
	\captionof{table}{The meta-level's architecture.}
	\label{tab.3}
\end{minipage}

\clearpage 
\section{Extended Results of Comparative Experiments}
\label{appendixc}
To thoroughly compare the performance of BiERL with the three baselines and the state-of-the-art RL methods, we also run experiments on other 3 MuJoCo (Hopper-v2, InvertedDoublePendulum-v2, and Swimmer-v2)  and 1 Box2D (BipedalWalker-v3) tasks as Figure \ref{fig.8} and Figure \ref{fig.s9}.
The basic settings of these experiments are consistent with the previous ones introduced in Section 4.1 and Appendix \ref{appendix1}. From the results, we can observe that BiERL with PM or NPM can also outperform the three baselines in these tasks, which further validates that BiERL can converge to better policies much faster.

Additionally, to demonstrate the results of the comparative experiments more intuitively,  the numerical
results of mean $\pm$ standard deviation over all episodes in all 8 tasks are reported in Table \ref{tab.4}. The best performance of each group of comparative experiments is indicated in bold font.

\begin{figure*}[tb]
	\setcounter{figure}{9}
	\centering 
	\subfigure[Hopper-v2]{\includegraphics[width=0.230\textwidth]{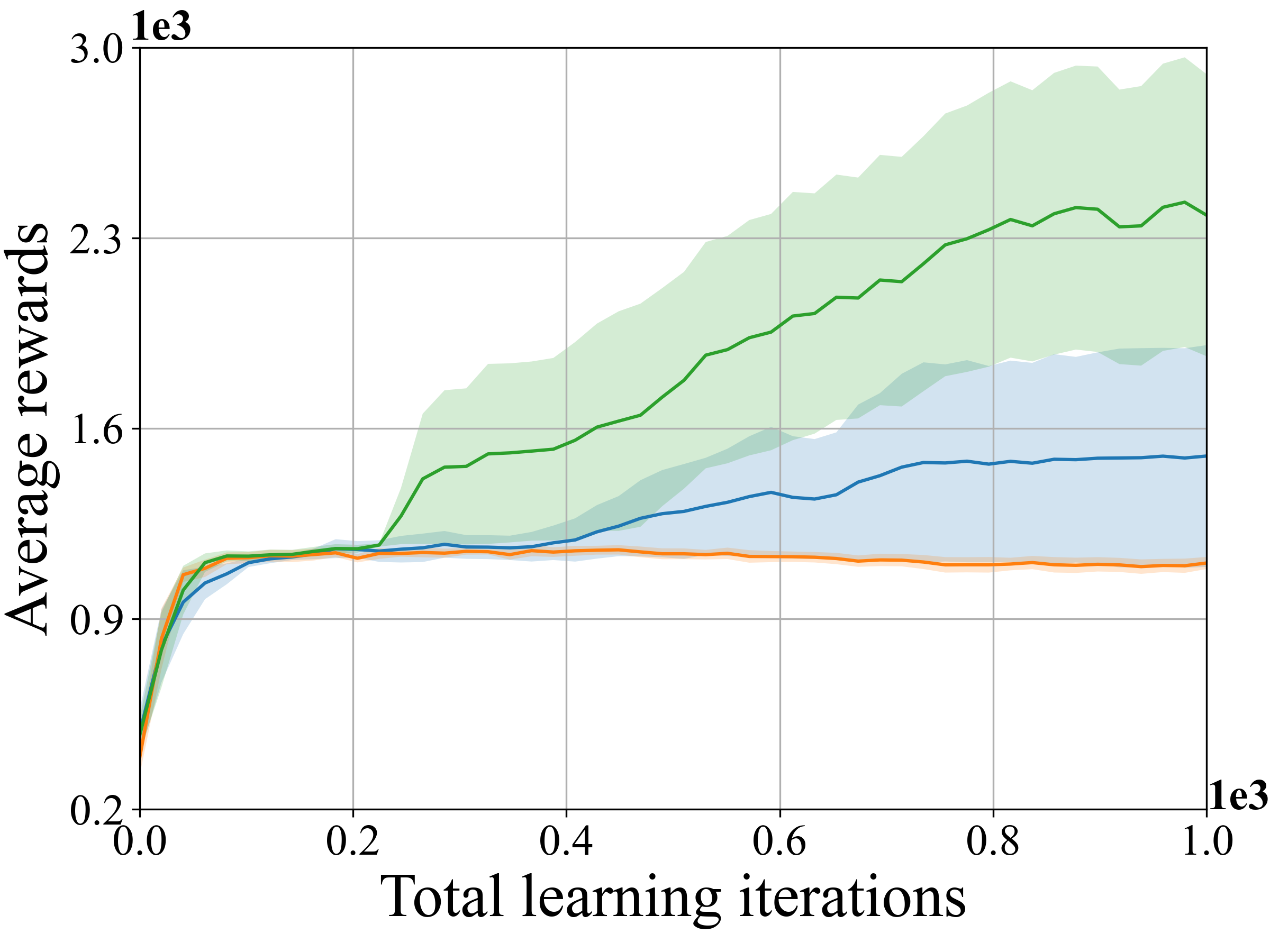}}
	\subfigure[InvertedDoublePendulum-v2]{\includegraphics[width=0.230\textwidth]{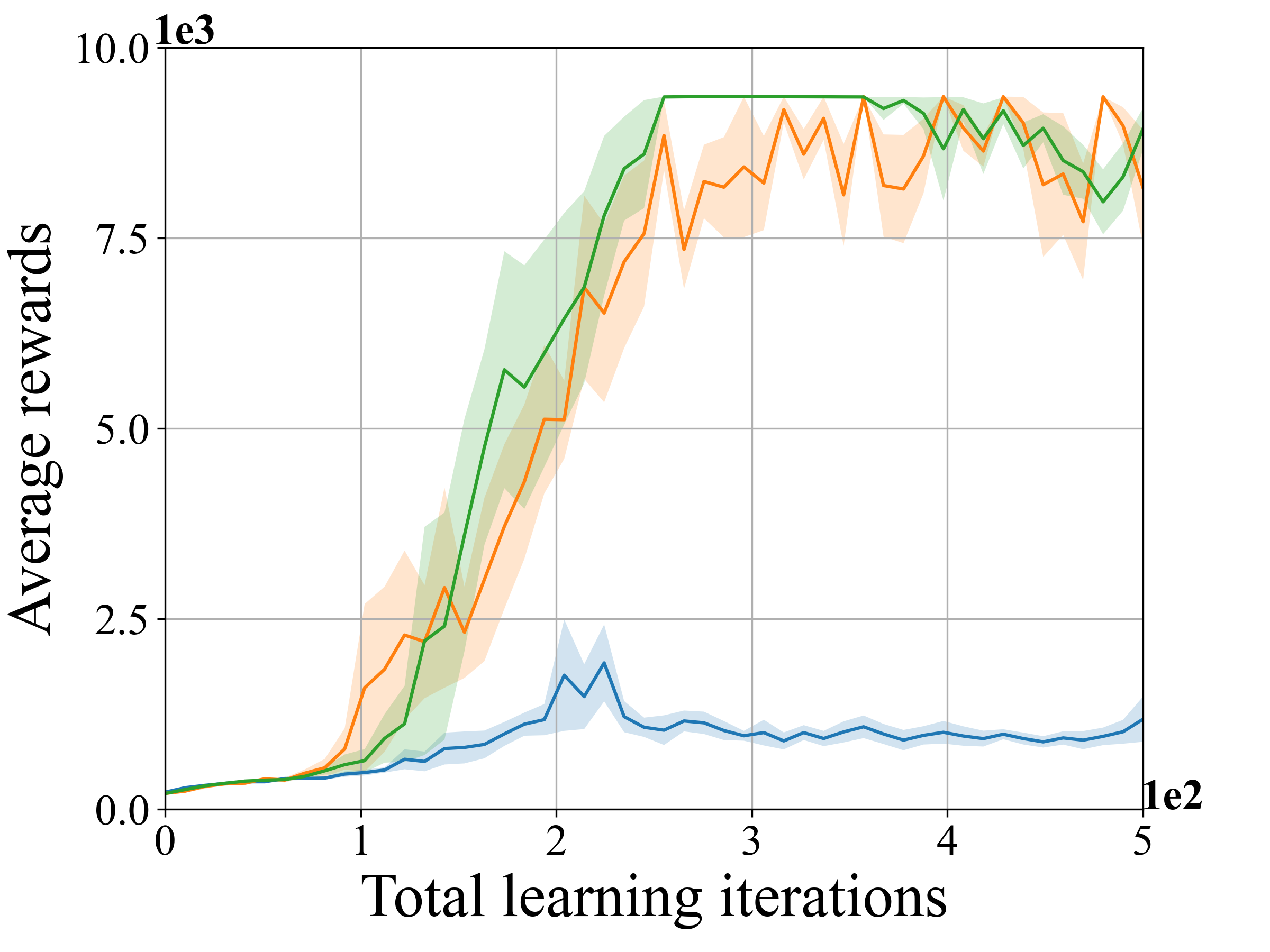}}
	\subfigure[Swimmer-v2]{\includegraphics[width=0.230\textwidth]{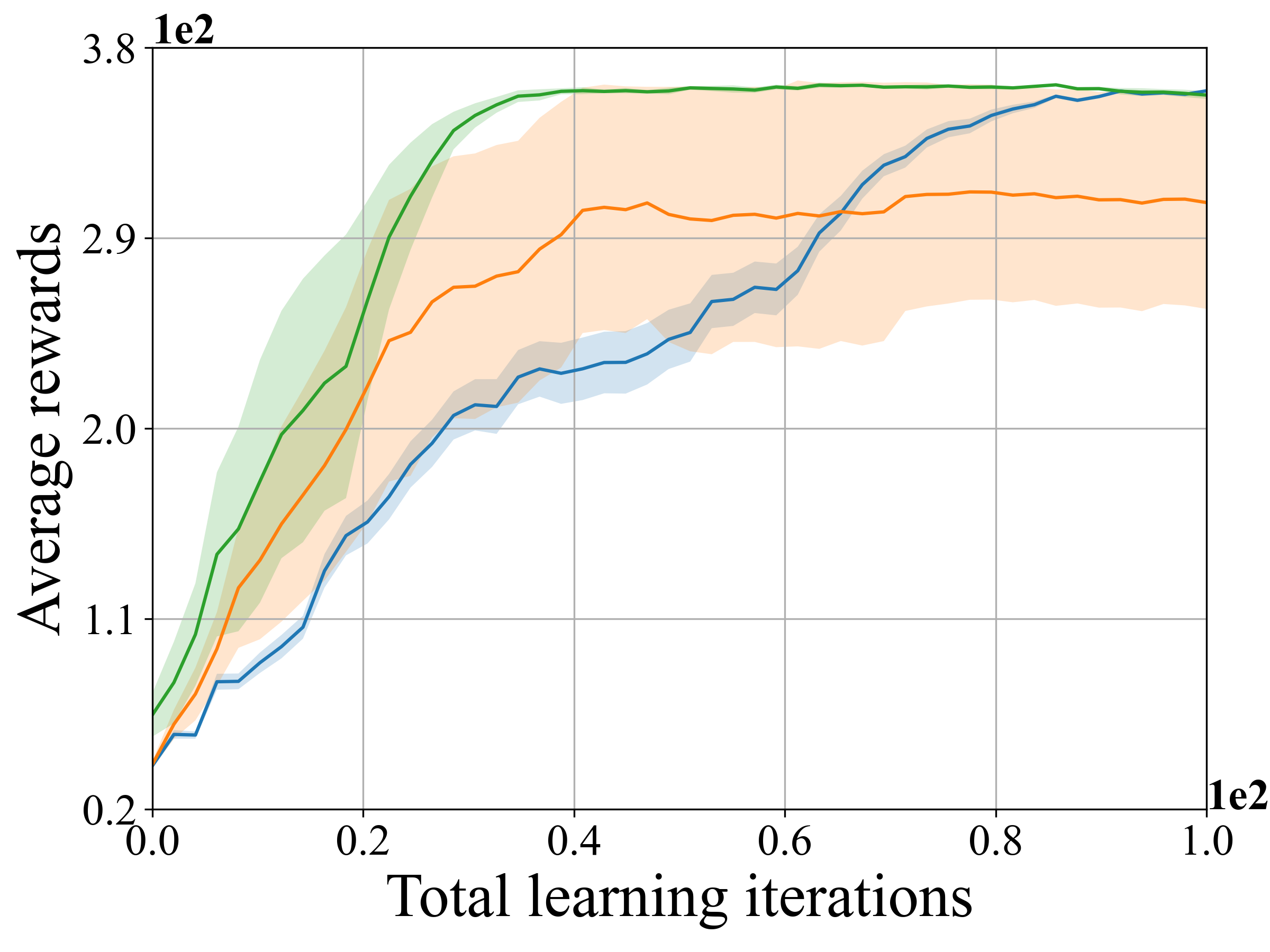}}
	\subfigure[BipedalWalker-v3]{\includegraphics[width=0.230\textwidth]{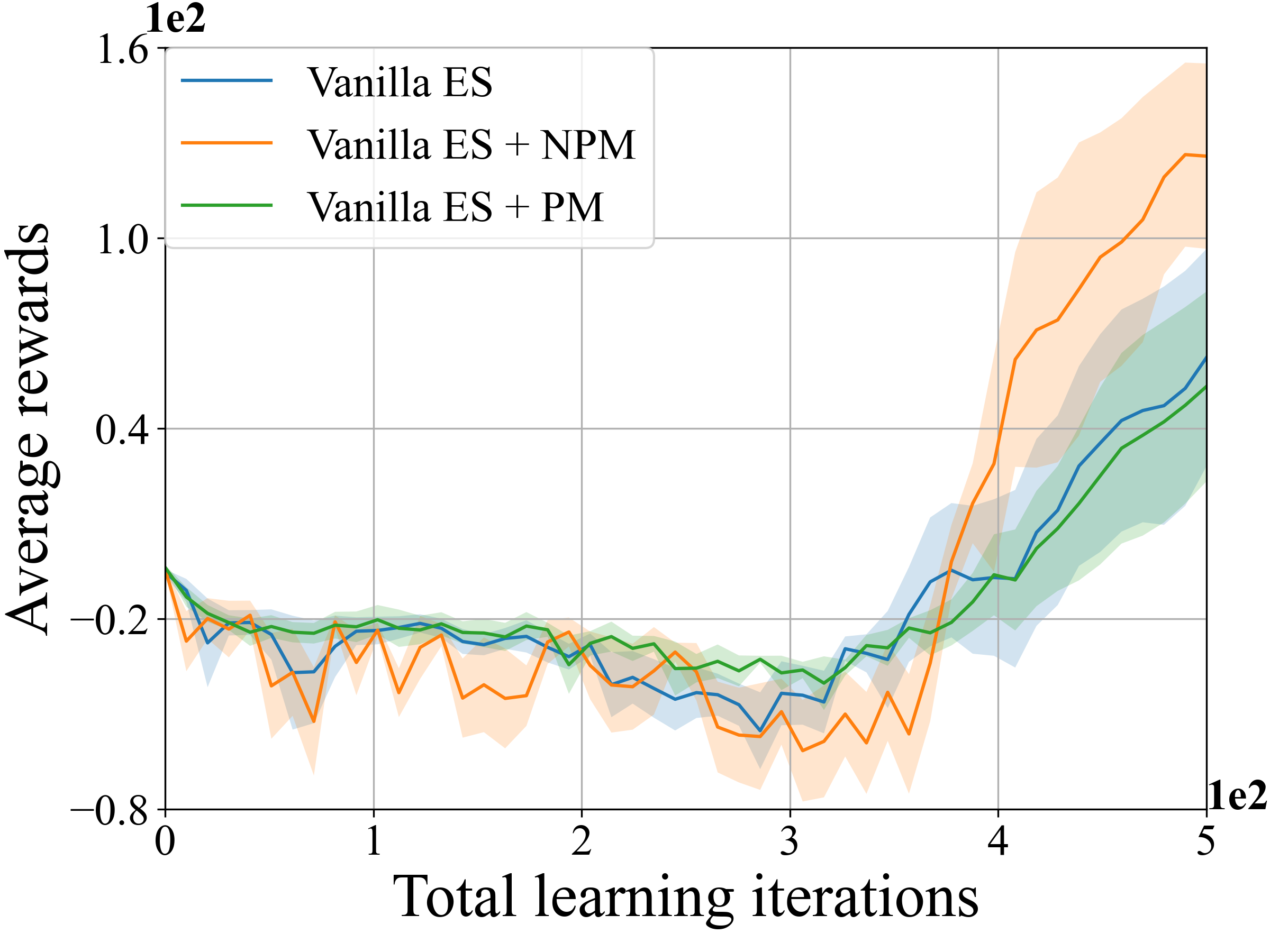}}
	\subfigure[Hopper-v2]{\includegraphics[width=0.230\textwidth]{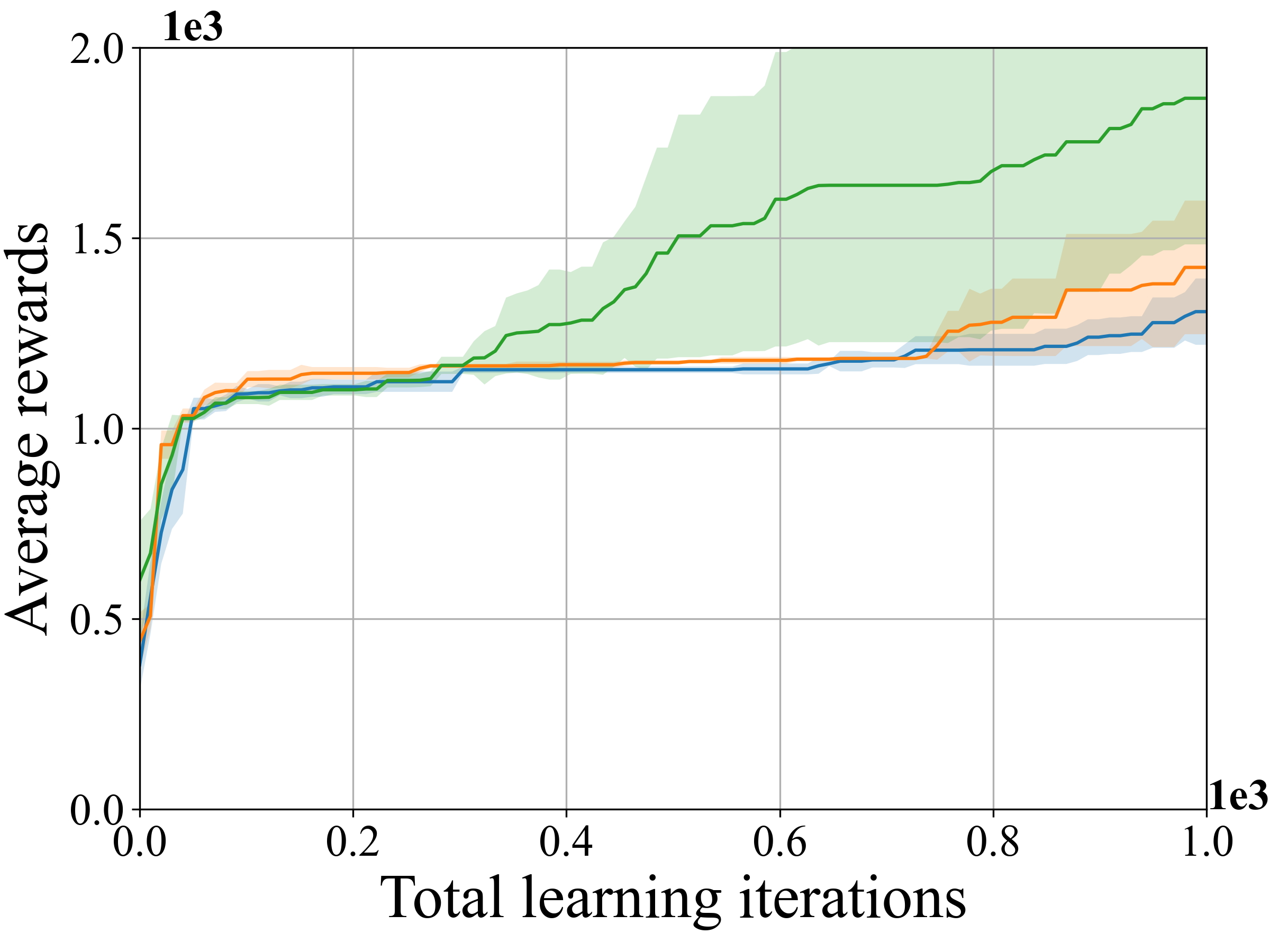}}
	\subfigure[InvertedDoublePendulum-v2]{\includegraphics[width=0.230\textwidth]{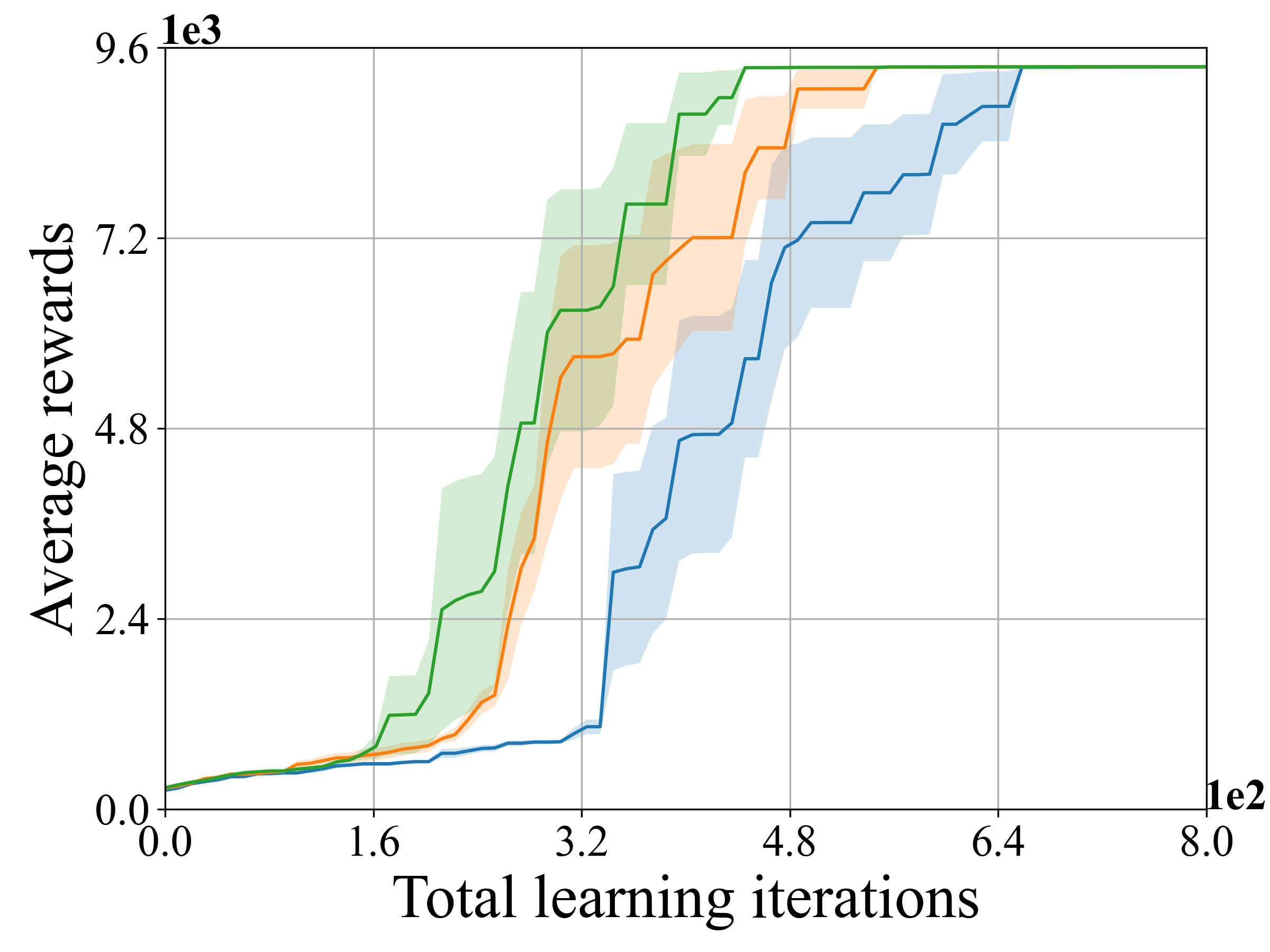}}
	\subfigure[Swimmer-v2]{\includegraphics[width=0.230\textwidth]{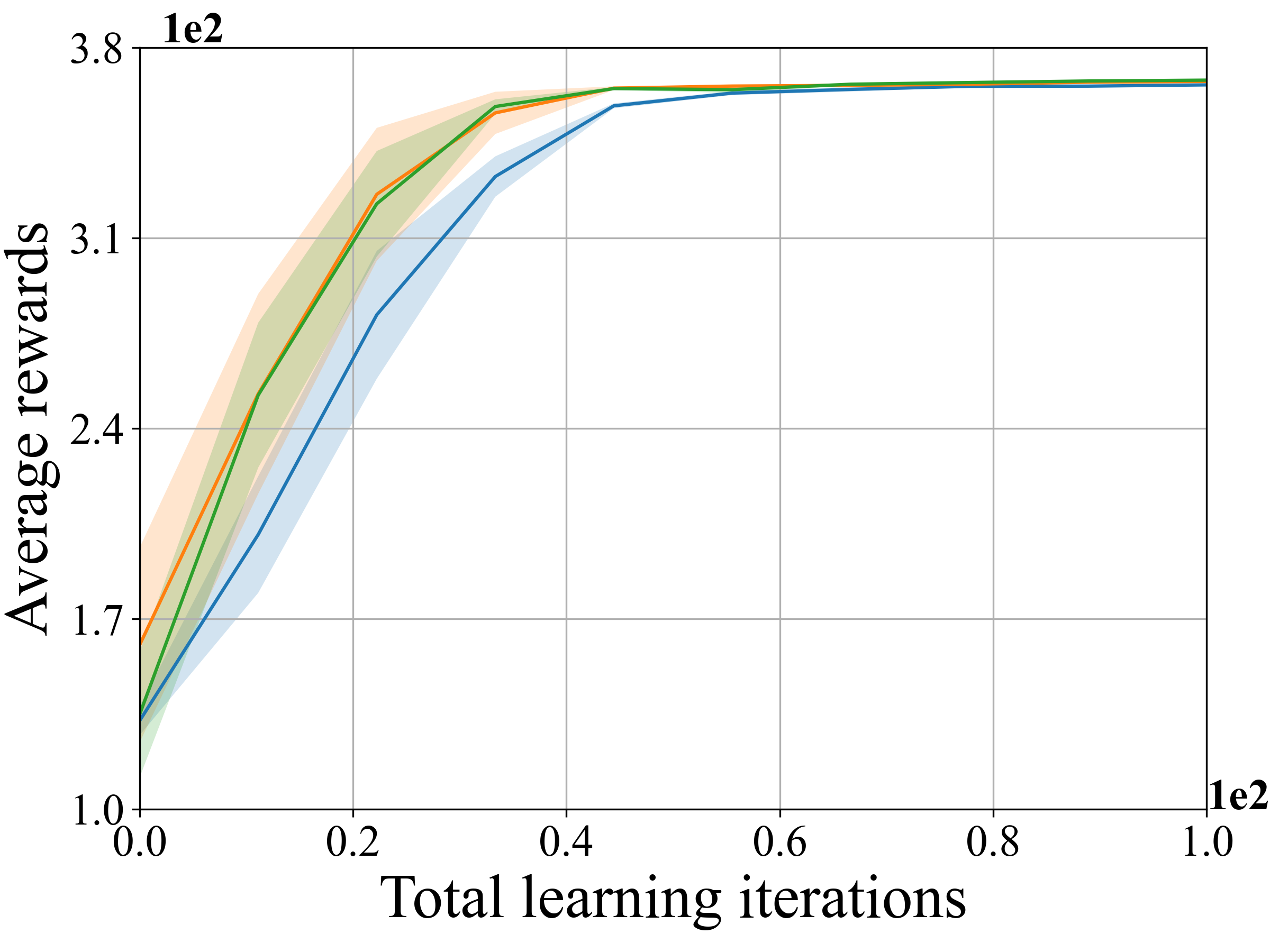}}
	\subfigure[BipedalWalker-v3]{\includegraphics[width=0.230\textwidth]{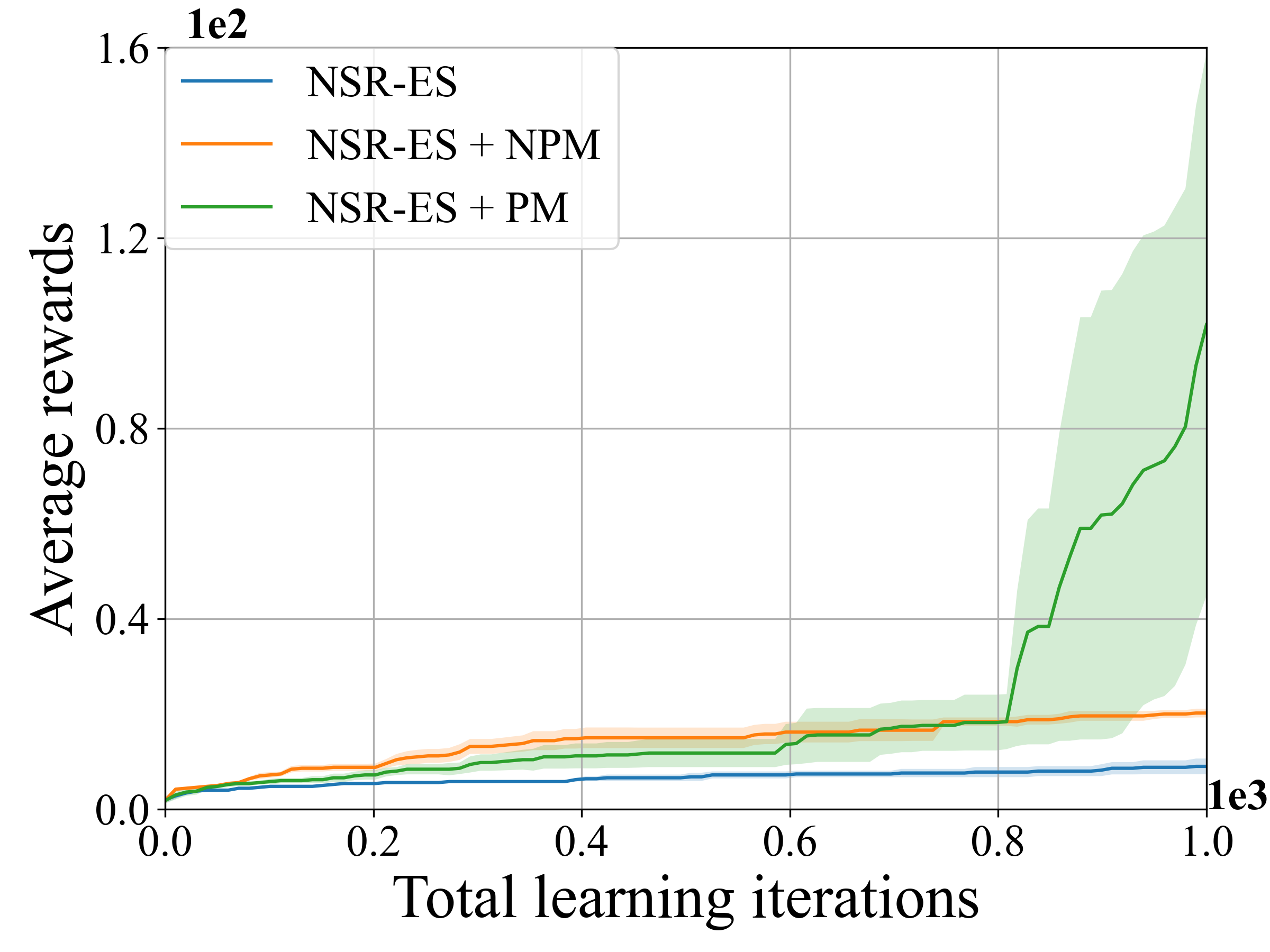}}
	\subfigure[Hopper-v2]{\includegraphics[width=0.230\textwidth]{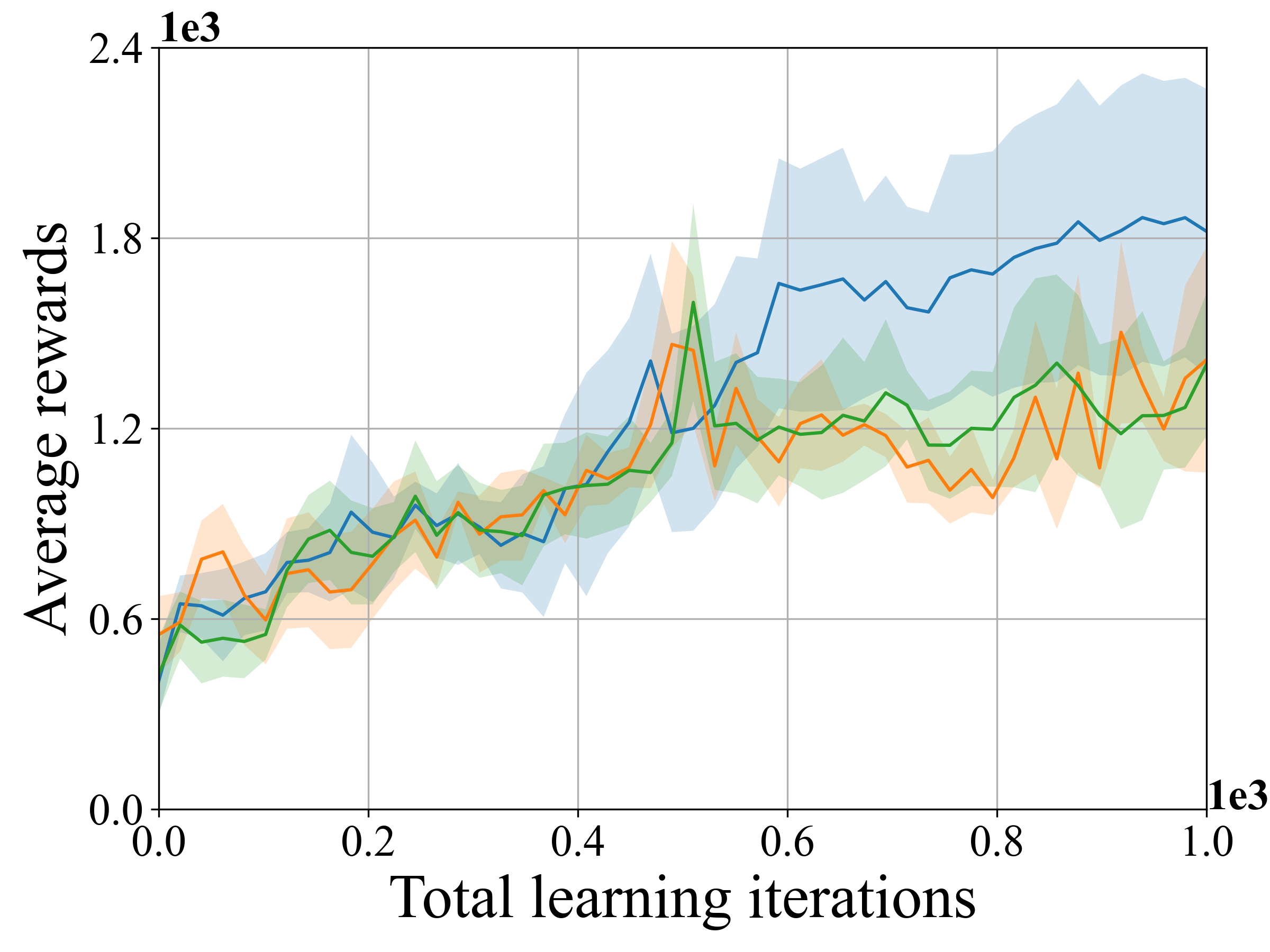}}
	\subfigure[InvertedDoublePendulum-v2]{\includegraphics[width=0.230\textwidth]{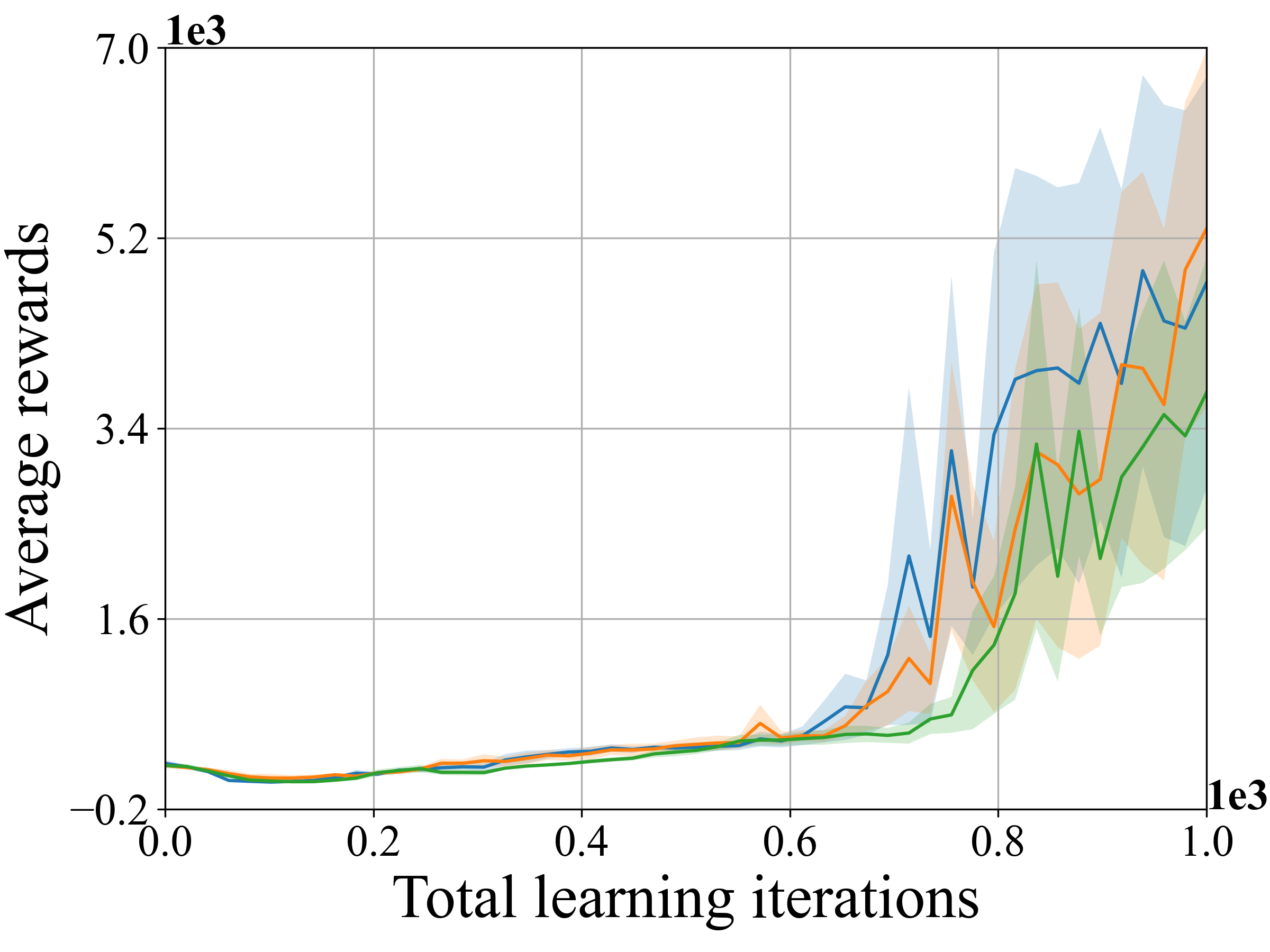}}
	\subfigure[Swimmer-v2]{\includegraphics[width=0.230\textwidth]{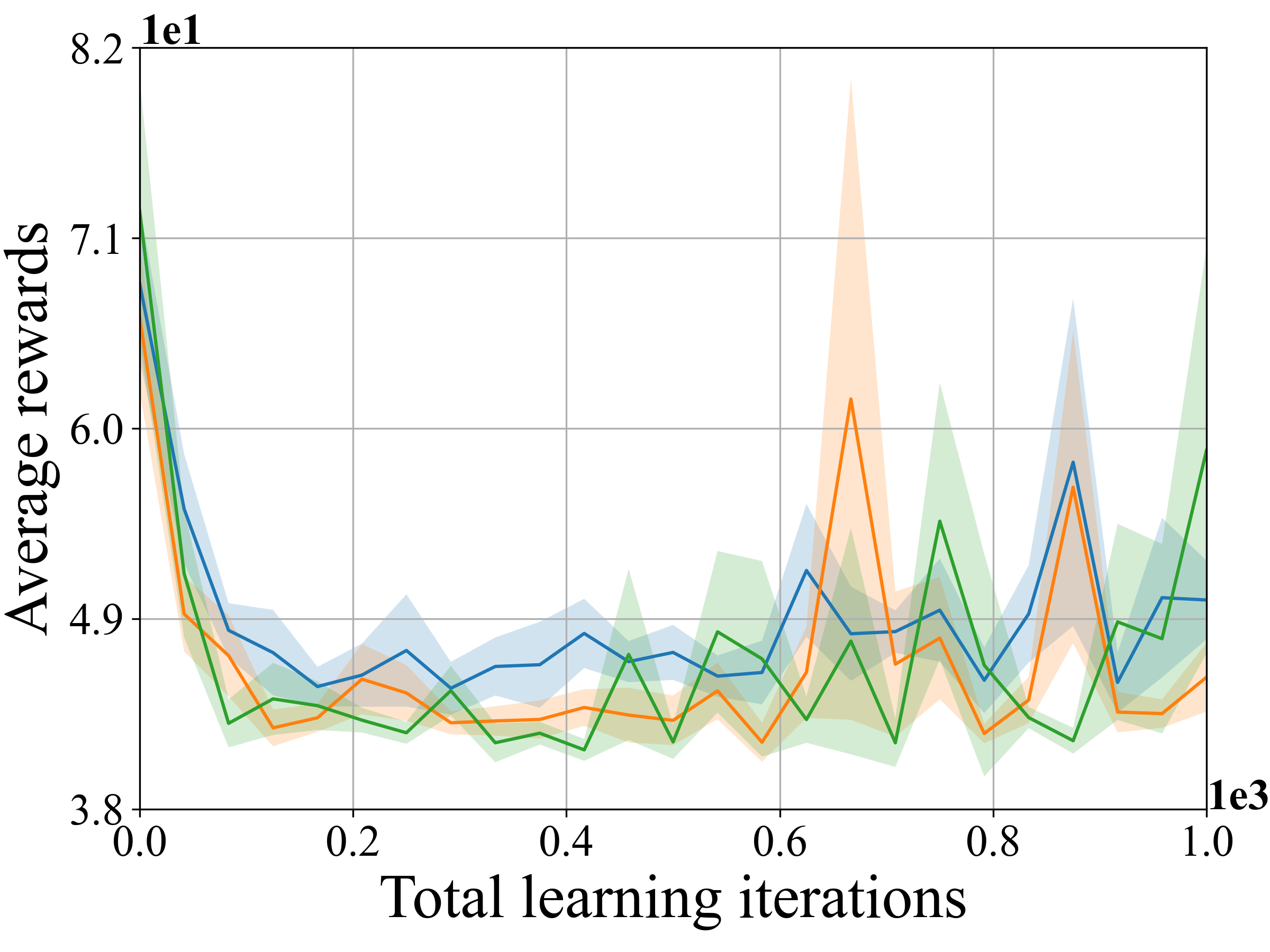}}
	\subfigure[BipedalWalker-v3]{\includegraphics[width=0.230\textwidth]{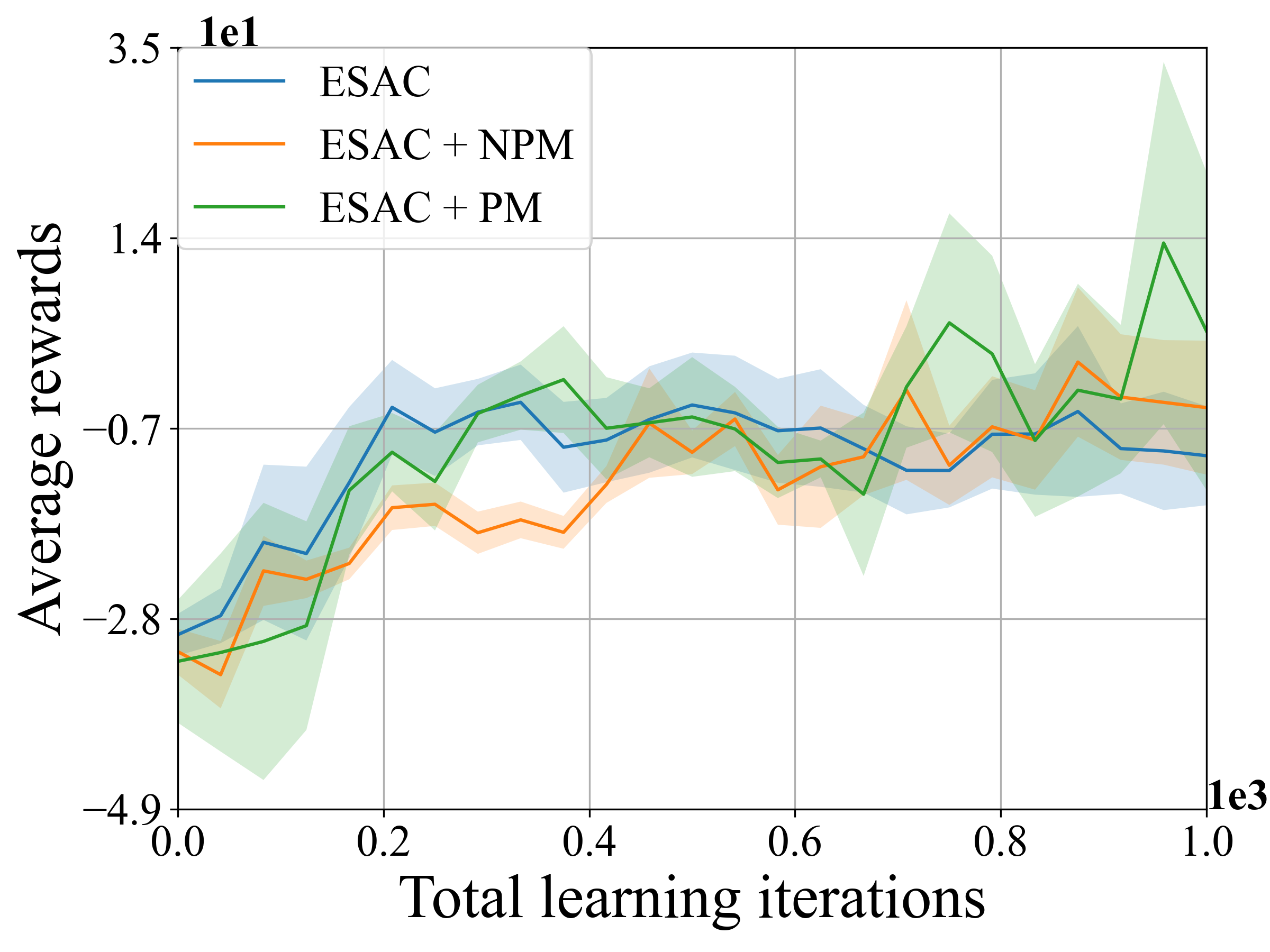}}
	\caption{Learning curves of extended comparative experiments.}
	\label{fig.8}
\end{figure*}
\begin{figure*}[tb]
	
	\centering 
	
	\subfigure[Hopper-v2]{\includegraphics[width=0.230\textwidth]{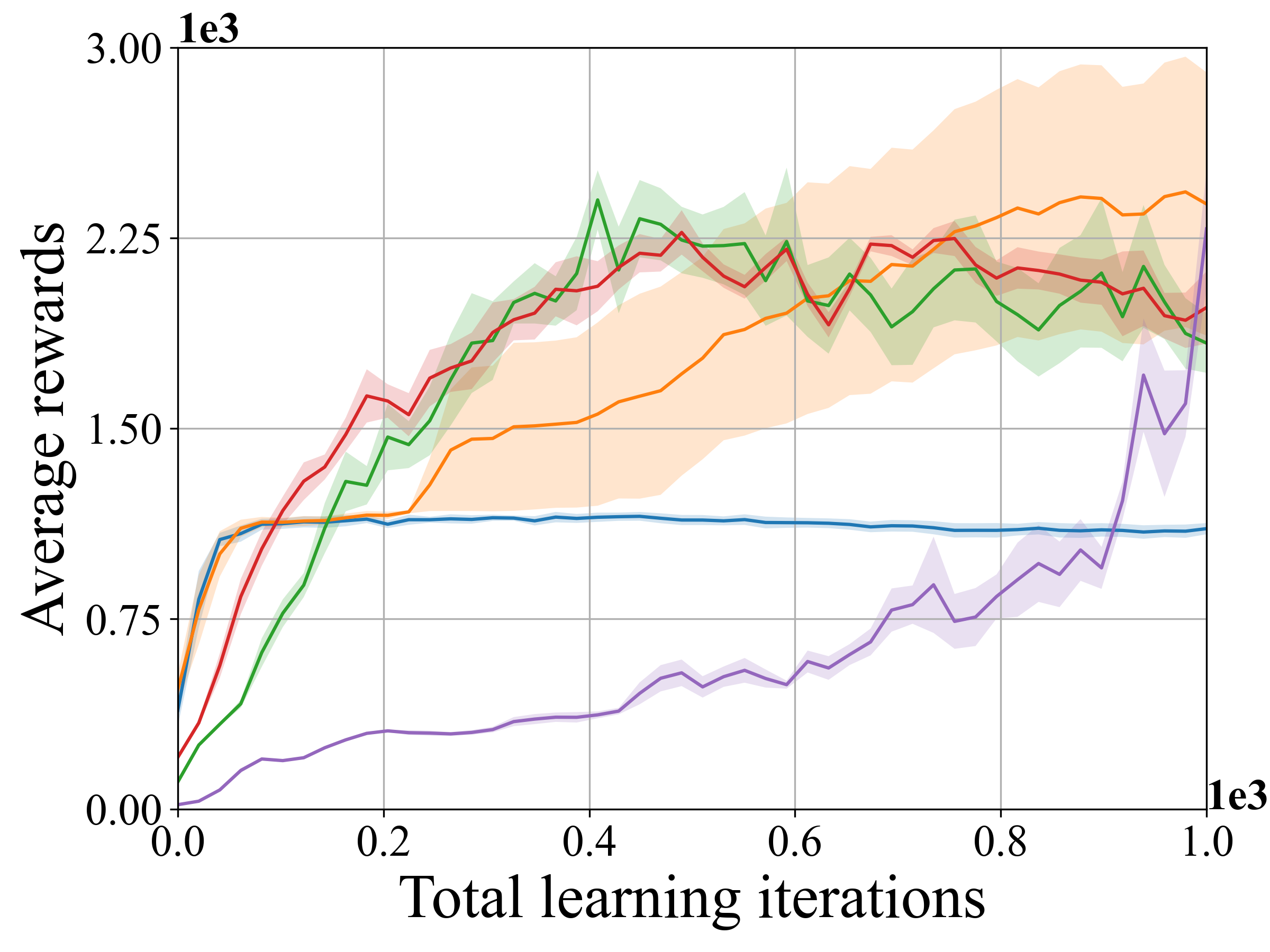}}
	\subfigure[InvertedDoublePendulum-v2]{\includegraphics[width=0.230\textwidth]{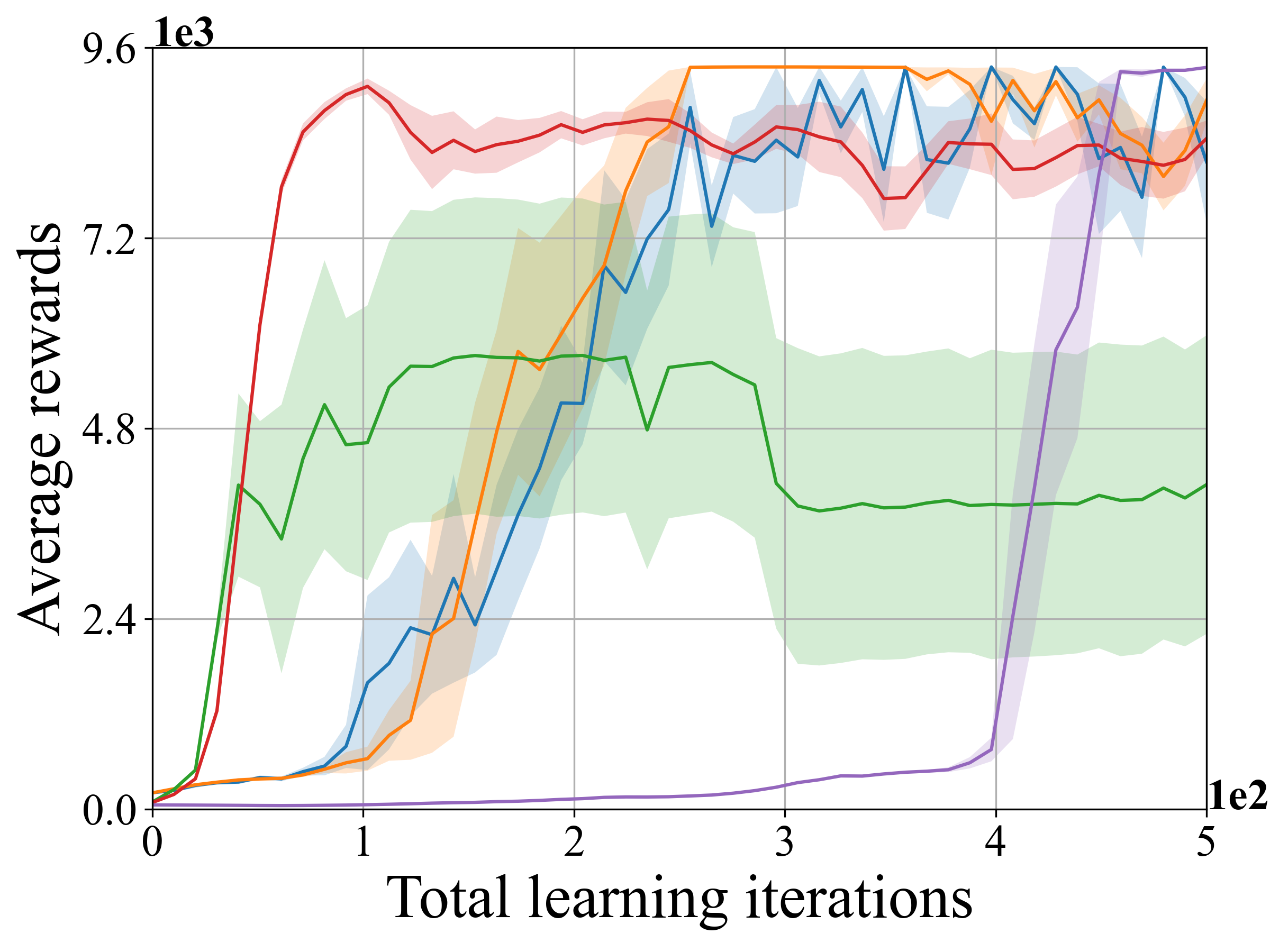}}
	\subfigure[Swimmer-v2]{\includegraphics[width=0.230\textwidth]{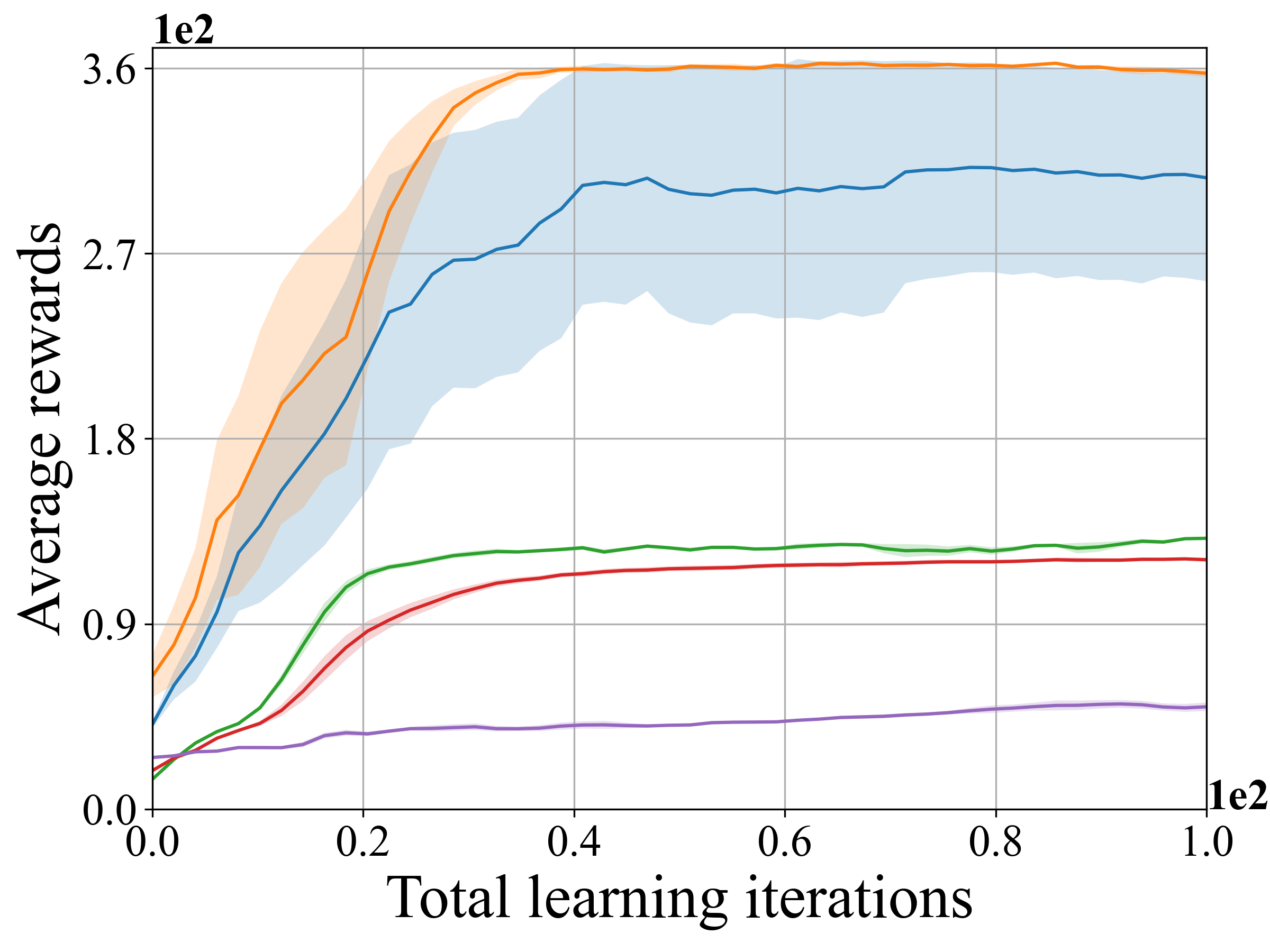}}
	\subfigure[BipedalWalker-v3]{\includegraphics[width=0.230\textwidth]{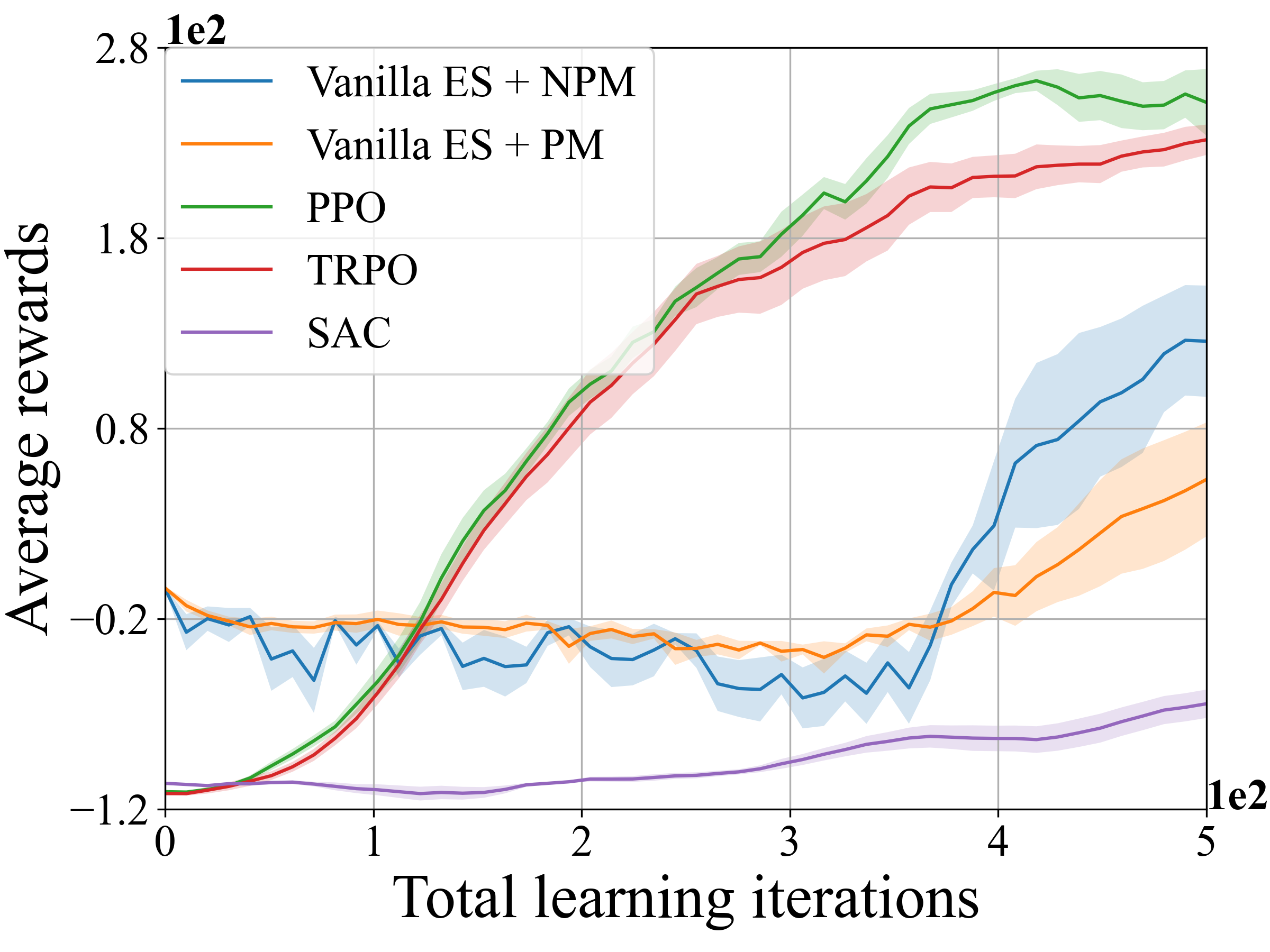}}
	
	\caption{Learning curves of extended comparative experiments with state-of-the-art RL methods.}
	\label{fig.s9}
\end{figure*}

\section{Extended Results of Comparative Experiments with other ERL methods}
\label{appendixd}
We investigate two representative Evolutionary Reinforcement Learning (ERL) approaches for optimizing hyperparameters in the context of AutoRL.
In detail, we conduct experiments with PB1~[3] and PB2~[5] to evaluate their effectiveness.
PB1 and PB2 depend on genetic algorithms (GAs) to evolve a population of gradient-based RL agents, such as PPO, with the goal of optimizing the network.
For similar reasons, other advanced ERL methods like ERL-Re$^2$~[4] are compatible with our methods, so we did not consider them baselines.

\begin{figure*}[tb]
	
	\centering 
	
	\subfigure[Ant-v2]{\includegraphics[width=0.230\textwidth]{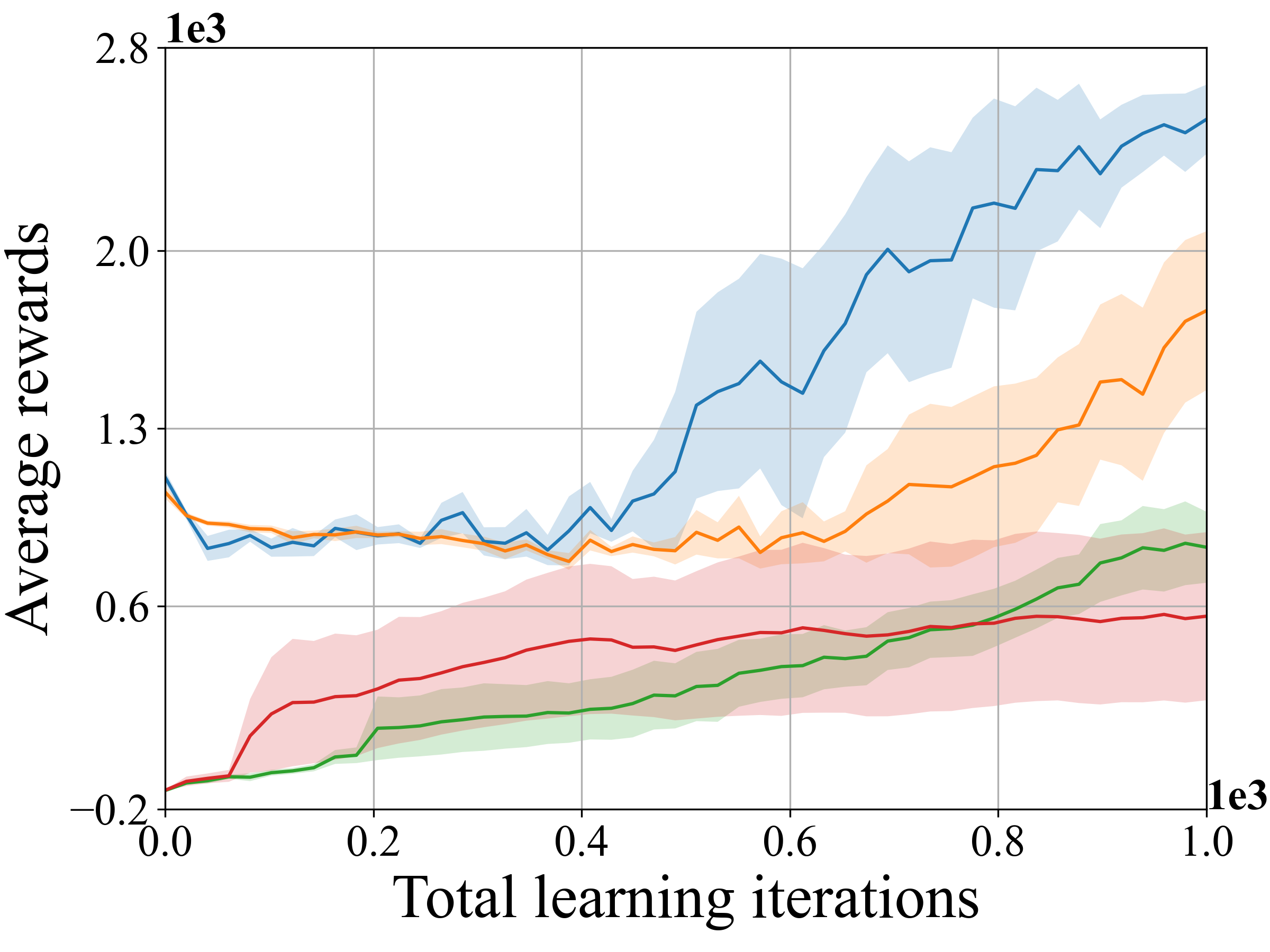}}
	\subfigure[HalfCheetah-v2]{\includegraphics[width=0.230\textwidth]{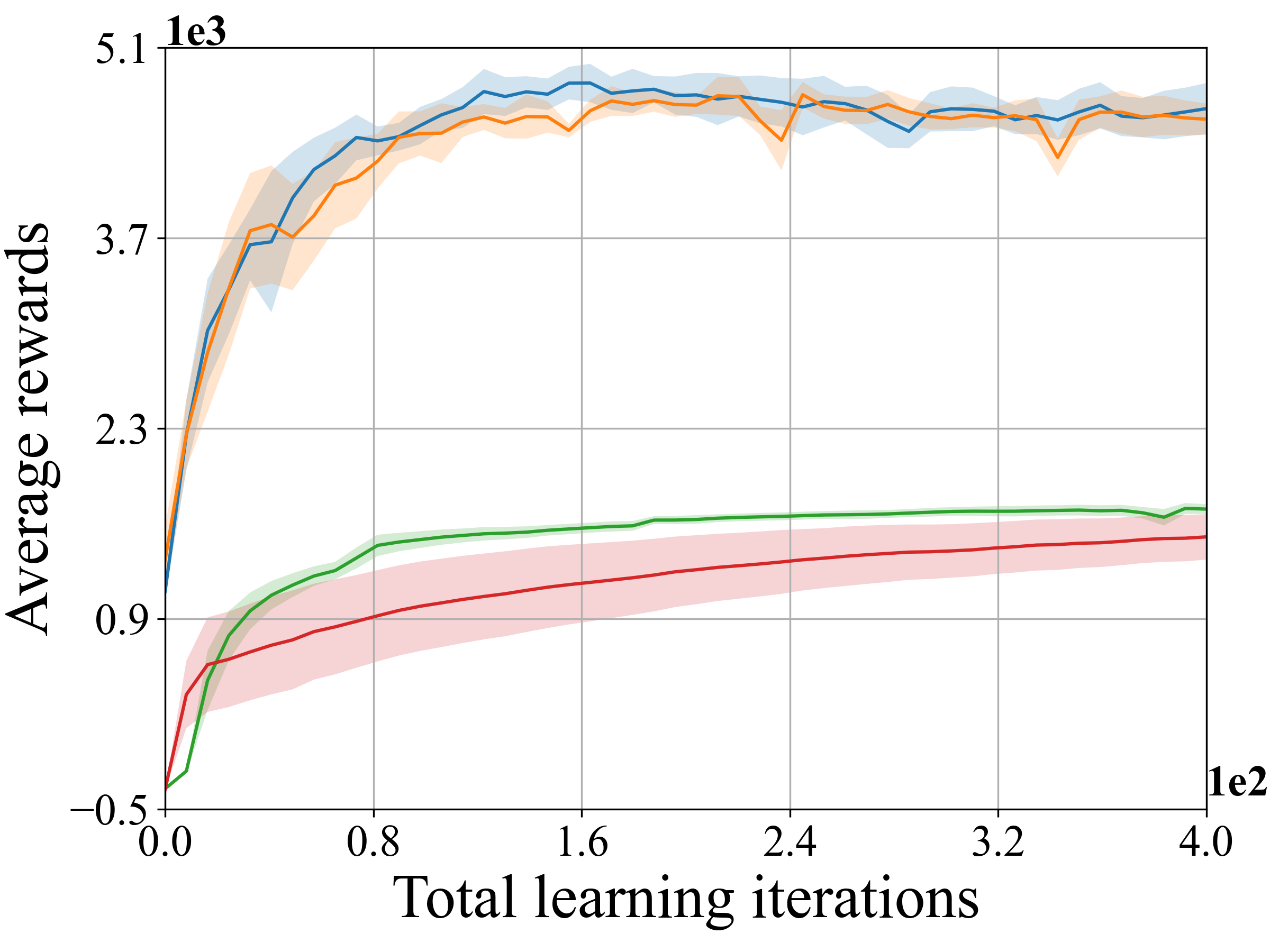}}
	\subfigure[Walker2d-v2]{\includegraphics[width=0.230\textwidth]{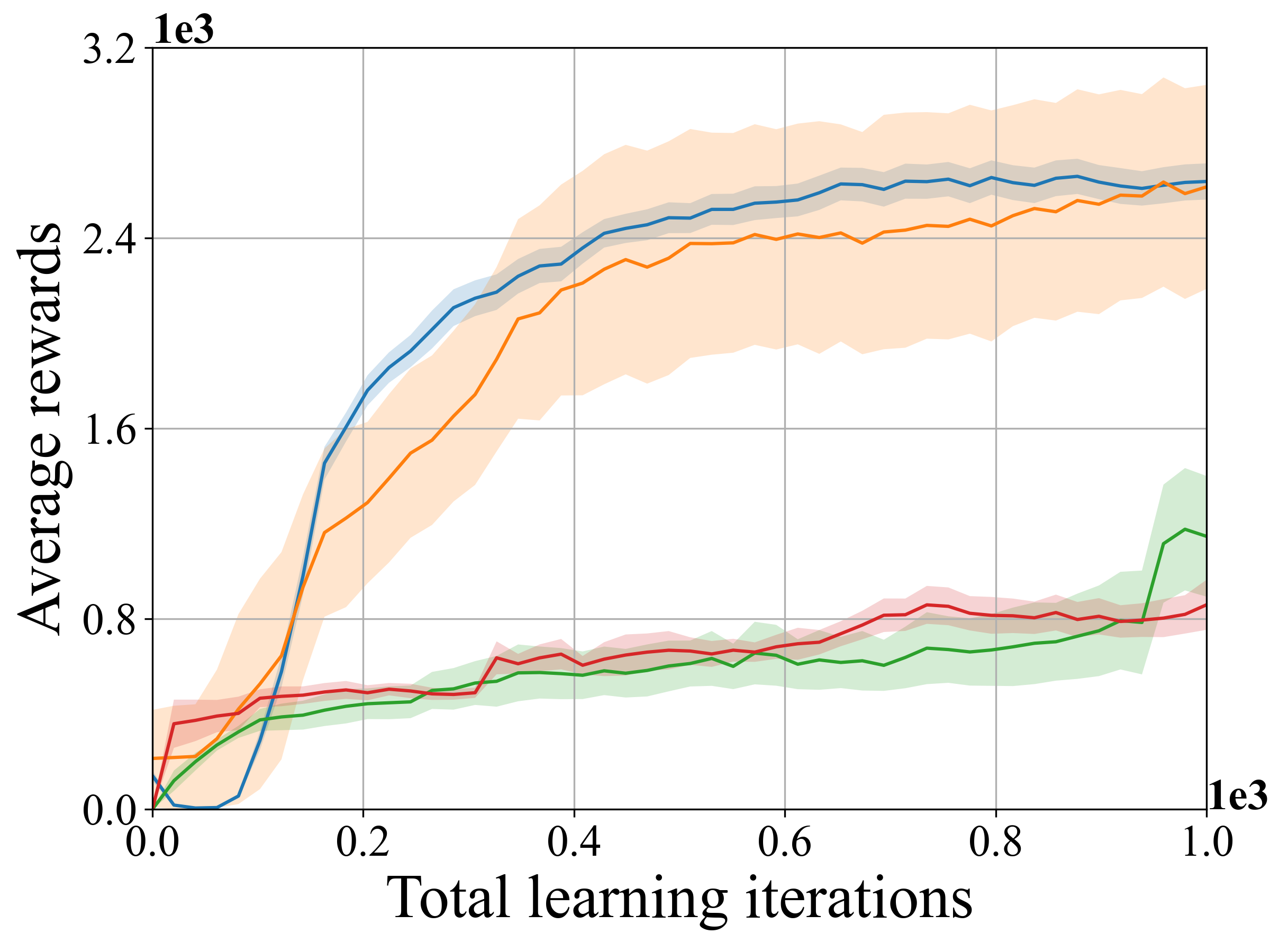}}
	\subfigure[LunarLanderContinuous-v2]{\includegraphics[width=0.230\textwidth]{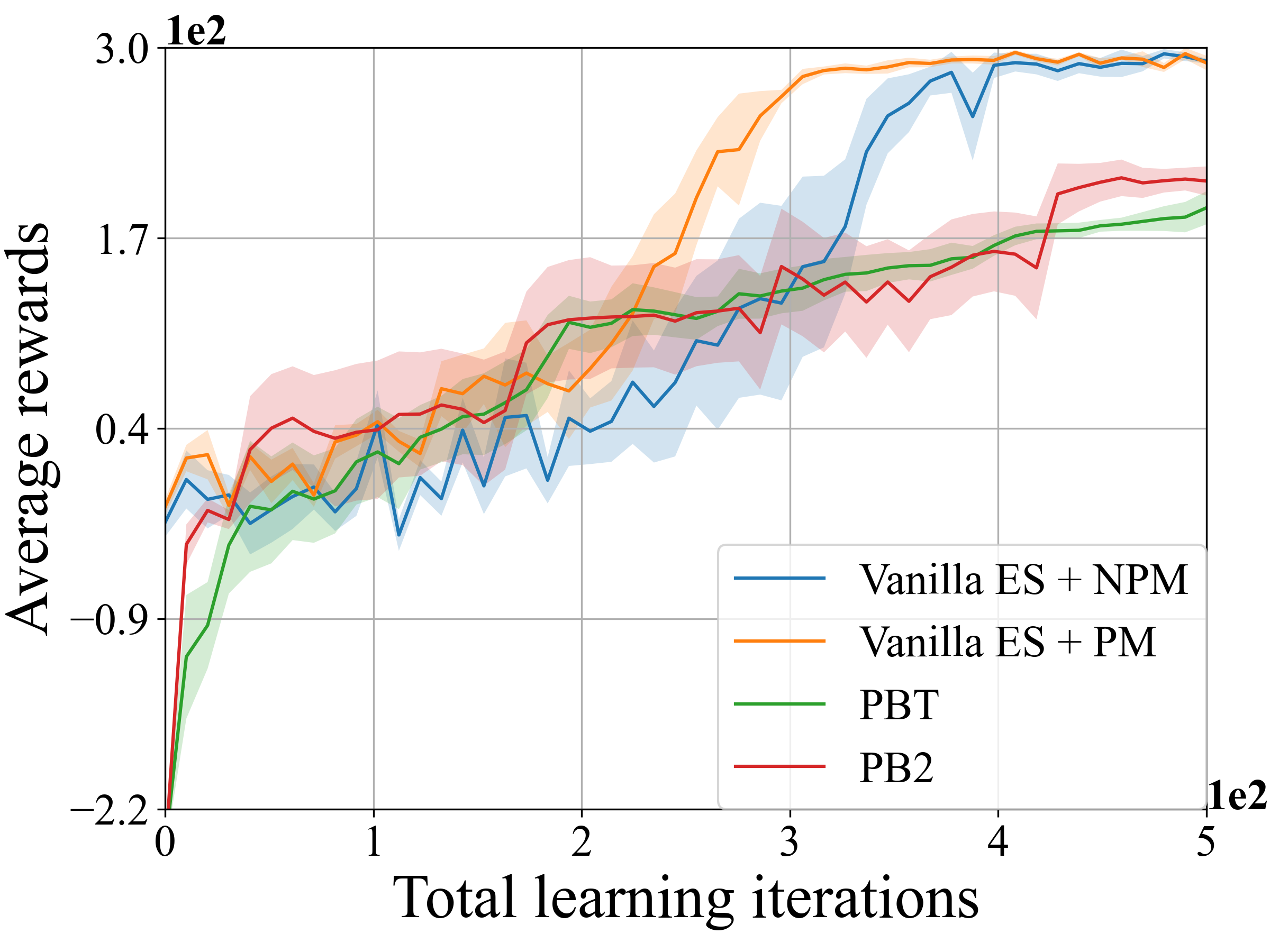}}
	\subfigure[Hopper-v2]{\includegraphics[width=0.230\textwidth]{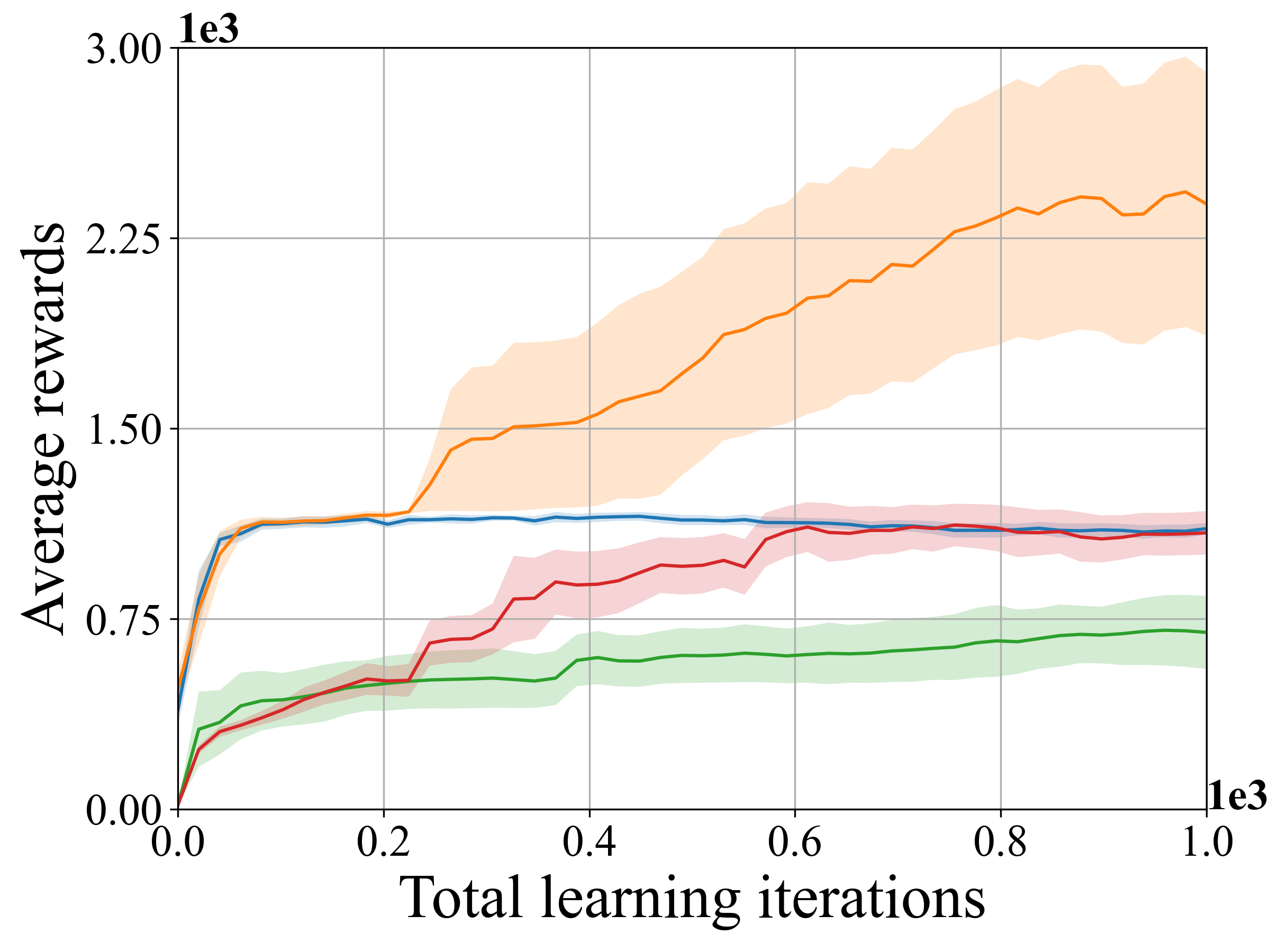}}
	\subfigure[InvertedDoublePendulum-v2]{\includegraphics[width=0.230\textwidth]{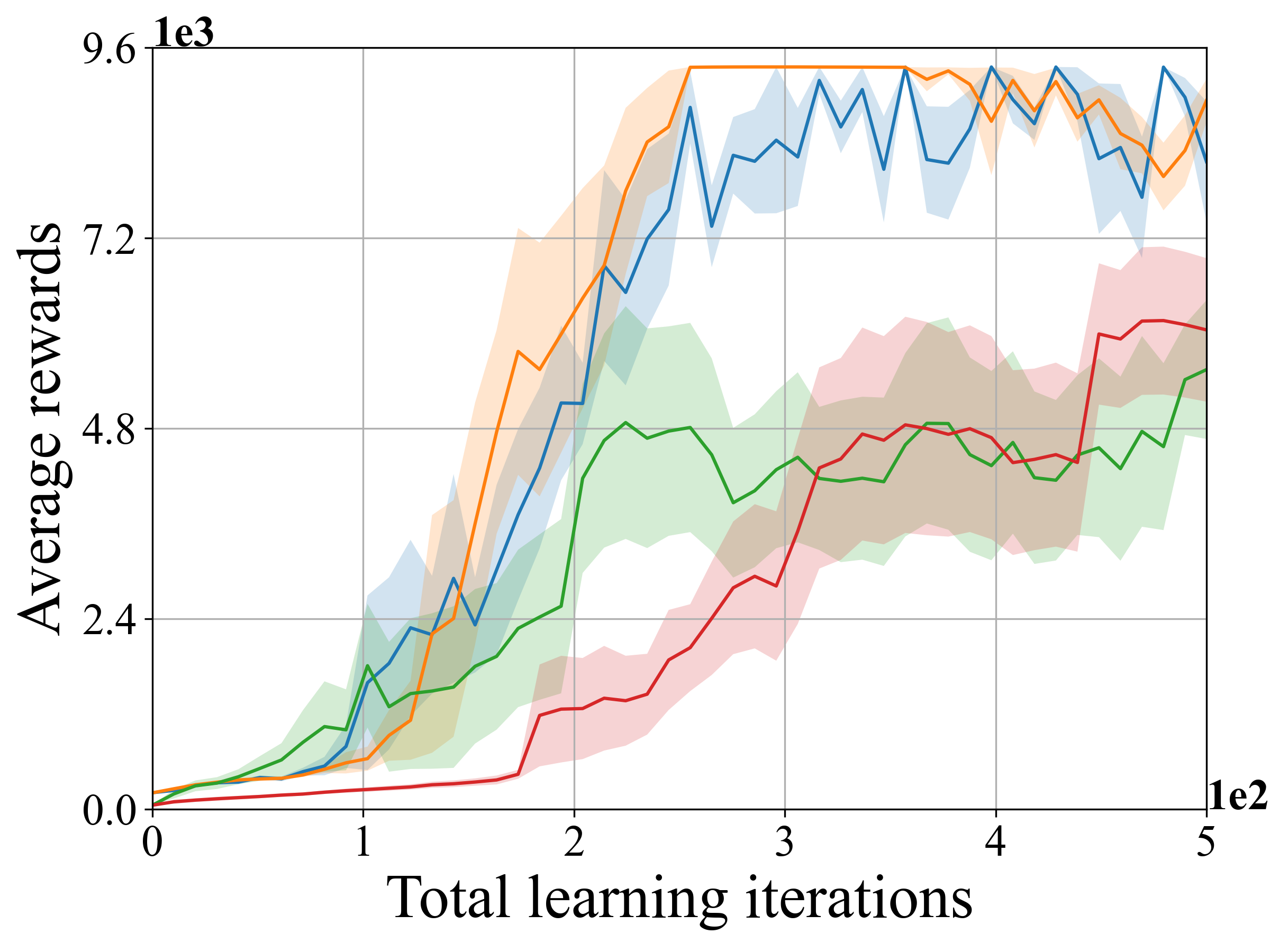}}
	\subfigure[Swimmer-v2]{\includegraphics[width=0.230\textwidth]{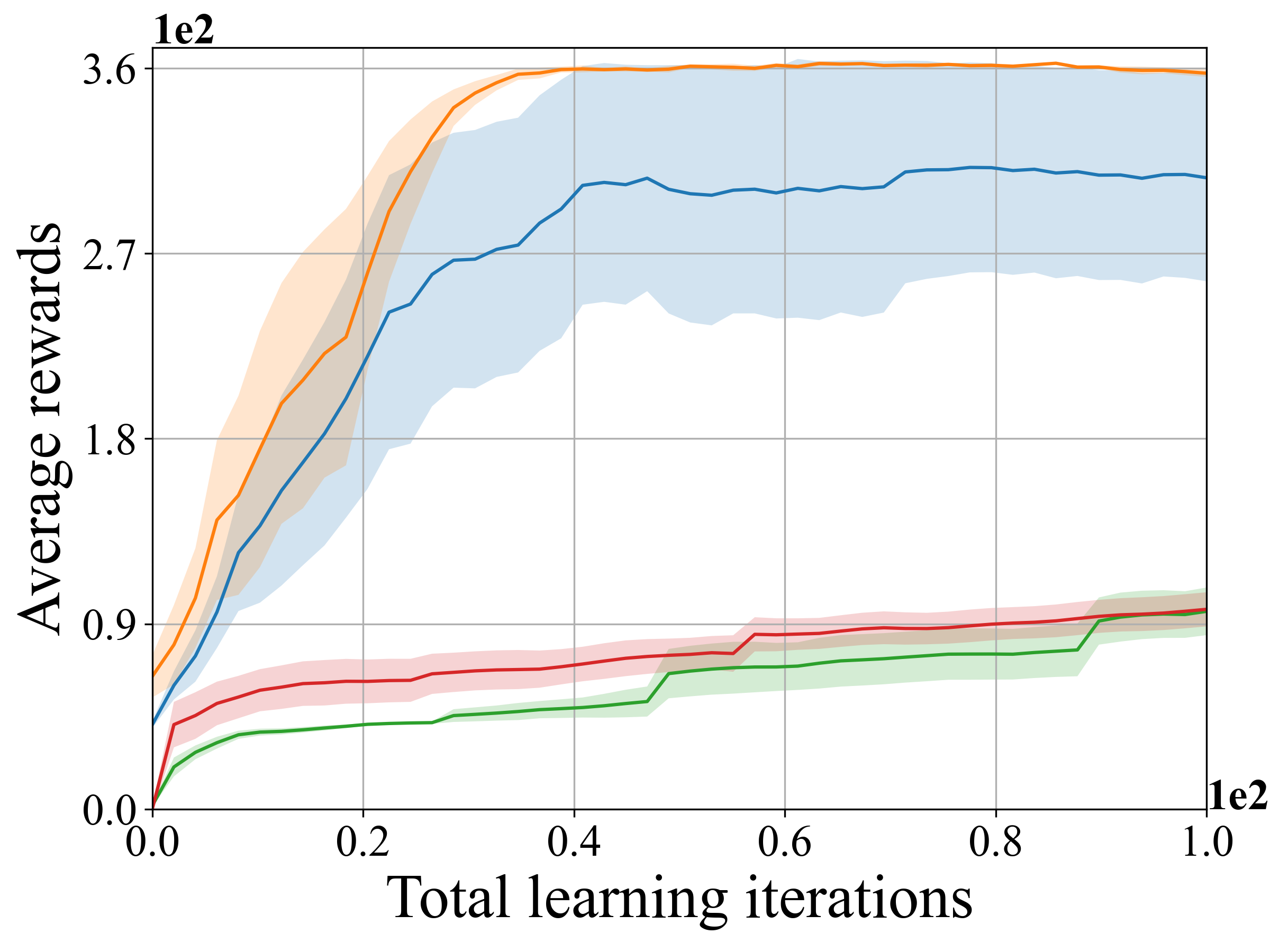}}
	\subfigure[BipedalWalker-v3]{\includegraphics[width=0.230\textwidth]{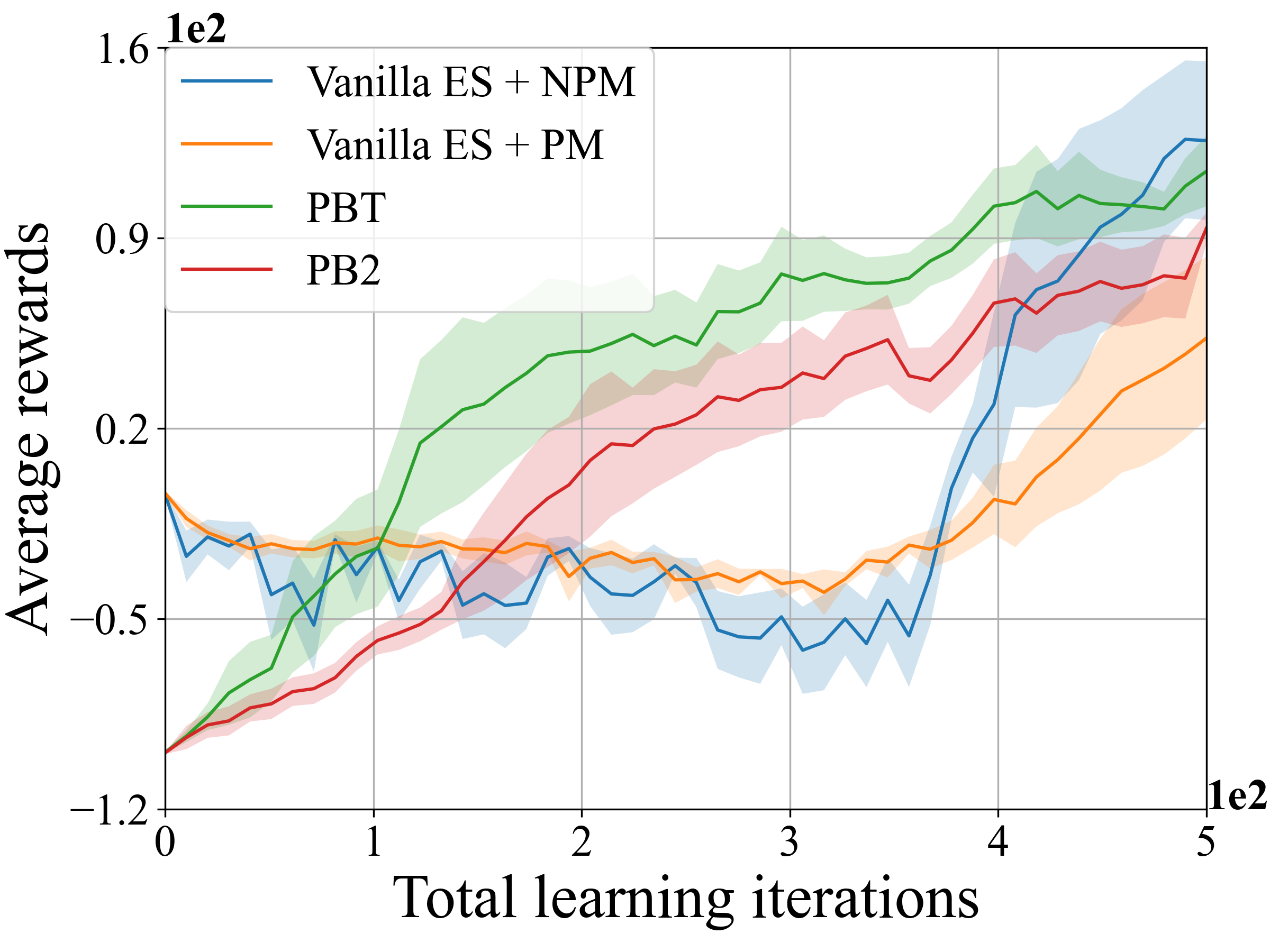}}
	
	\caption{Learning curves of extended comparative experiments with state-of-the-art ERL methods.}
	\label{fig.s10}
\end{figure*}

As illustrated in Figure~\ref{fig.s10}, BiERL outperforms both PB1 and PB2 across most tasks, demonstrating that our method can optimize multiple hyperparameters and improve learning performance, which should benefit from the fact that BiERL is a derivative-free framework. 
Its adaptive approach allows for simultaneously updating hyperparameters while training the ERL model within a single agent, further optimizing resource utilization. 
PB1 and PB2 utilize a static approach to determine optimal hyperparameters before training and keep them fixed during the learning process, while BiERL continuously tunes hyperparameters adaptively.
Our experimental results reveal the key differences between BiERL, PB1, and PB2 methods. 
The BiERL technique is a promising and efficient method for automated reinforcement learning (AutoRL), specifically in the context of ERL-based hyperparameter optimization.
It features a framework that is free from derivatives, adaptive hyperparameter updates, and a continuous tuning approach.

\begin{table*}[tb]

	\renewcommand\arraystretch{1.5}
	\centering
	\begin{tabular}{l|cccc}
		
		\hline
		\multicolumn{1}{c|}{}              & Ant-v2                        & HalfCheetah-v2                 & Walker2d-v2                   & LunarLanderContinuous-v2    \\ \hline
		Vanilla ES                         & $618.92 \pm 34.45$            & $3590.63\pm 226.63$           & $1794.47 \pm 292.98$          & $286.20 \pm 4.58$           \\
		\hdashline
		Vanilla ES + NPM & \bm{$2517.64 \pm 136.49$} & \bm{$4651.45 \pm 188.08$}           & \bm{$2637.54 \pm 380.39$} &  \bm{$290.74 \pm 1.64$}           \\
		\hdashline
		Vanilla ES + PM     & $1763.95 \pm 313.01$          & $4573.50 \pm 115.36$   & $2614.81 \pm 428.61$          & $289.51 \pm 4.88$  \\ \hline
		NSR-ES                             & $1854.40 \pm 91.07$           & $5364.80 \pm 110.13$           & $2906.00 \pm 259.53$          & $300.40 \pm 5.46$    
		\\
		\hdashline
		NSR-ES + NPM     & $2256.60 \pm 59.35$           & $5361.40 \pm 137.29$           & \bm{$3630.00 \pm 106.77$} & \bm{$305.20 \pm 6.67$}  \\
		\hdashline
		NSR-ES + PM         & \bm{$2946.40 \pm 40.19$}  & \bm{$5849.40 \pm 82.08$}   & $3163.60 \pm 75.37$           & $301.20 \pm 3.36$           \\ \hline
		ESAC                               & $-40.70 \pm 4.50$             & $1118.40 \pm 122.09$             & $1464.74 \pm 146.18$          & $158.29 \pm 46.71$          \\
		\hdashline
		ESAC + NPM       & \bm{$952.51 \pm 23.56$}   & \bm{$3126.72 \pm 306.60$}  & $1762.80 \pm 327.35$ & $ 119.41\pm 37.87$          \\
		\hdashline
		ESAC + PM           & $922.38 \pm 52.66$            & $2787.44 \pm 667.43$           & \bm{$2017.99\pm 425.81$}           & \bm{$245.57 \pm 3.55$} \\ \hline
		\multicolumn{1}{c|}{}              & Hopper-v2                     & InvertedDoublePendulum-v2      & Swimmer-v2                    & BipedalWalker-v3            \\ \hline
		Vanilla ES                         & $1498.44 \pm 407.70$          & $1182.05 \pm 296.74$           & \bm{$360.13 \pm 0.75$}    & $62.33 \pm 34.22$           \\
		\hdashline
		Vanilla ES + NPM & $1105.12 \pm 22.13$           & $8161.66 \pm 759.88$    & $306.76 \pm 50.29$            & \bm{$125.79 \pm 29.20$} \\
		\hdashline
		Vanilla ES + PM     & \bm{$2383.99 \pm 518.61$} & \bm{$8932.75 \pm 272.80$}             & $357.56 \pm 1.81$             & $53.22 \pm 29.91$           \\ \hline
		NSR-ES                             & $1306.62 \pm 87.04$           & $9356.64 \pm 1.20$             & $366.30 \pm 0.23$             & $9.00 \pm 1.47$             \\
		\hdashline
		NSR-ES + NPM     & $1422.83 \pm 175.41$          & $9359.02 \pm 0.30$             & $367.56 \pm 0.30$           & $20.20 \pm 0.82$            \\
		\hdashline
		NSR-ES + PM         & \bm{$1866.95 \pm 383.70$} & \bm{$9359.26 \pm 0.11$}    & \bm{$368.04 \pm 0.18$}    & \bm{$101.80\pm 50.93$}  \\ \hline
		ESAC                               & \bm{$1820.93 \pm 449.35$} & $4775.50 \pm 1949.00$          & $50.08 \pm 2.28$              & $-10.01\pm 5.47$             \\
		\hdashline
		ESAC + NPM       & $1416.05 \pm 355.36$          & \bm{$5287.26 \pm 1685.40$} & $45.63 \pm 1.98$              & $-4.71 \pm 7.39$             \\
		\hdashline
		ESAC + PM           & $1401.13\pm 225.16$          & $3736.99 \pm 1274.71$          & \bm{$58.72 \pm 11.73$}     & \bm{$3.74 \pm 17.54$}  \\ \hline
	\end{tabular}
	\caption{Numerical results of comparative experiments over all episodes.}
	\label{tab.4}
\end{table*}
\begin{figure}[h!]
	\centering 
	\subfigure[HalfCheetah ($\omega$)]{\includegraphics[width=0.230\textwidth]{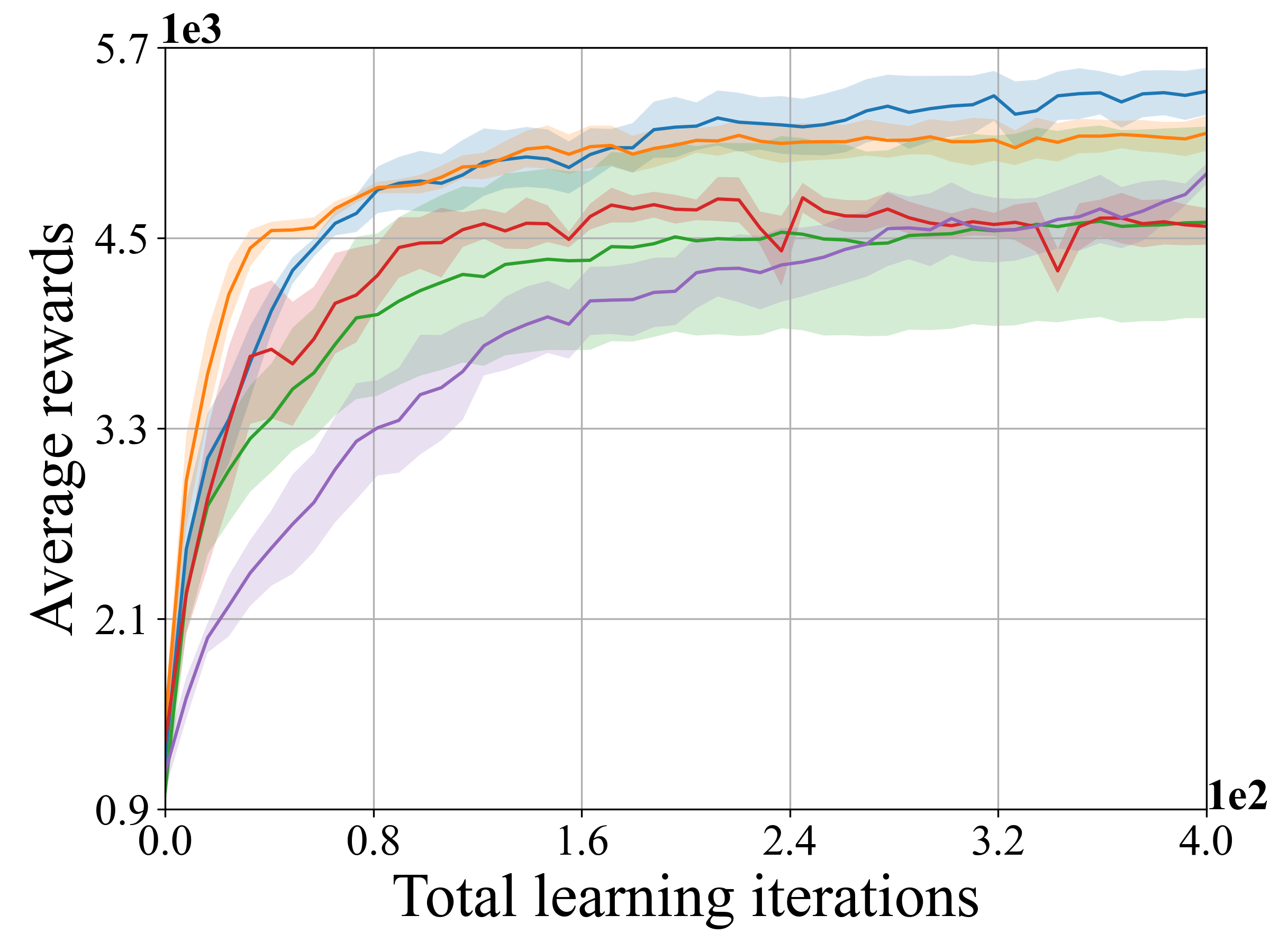}}
	\subfigure[Walker2d ($\omega$)]{\includegraphics[width=0.230\textwidth]{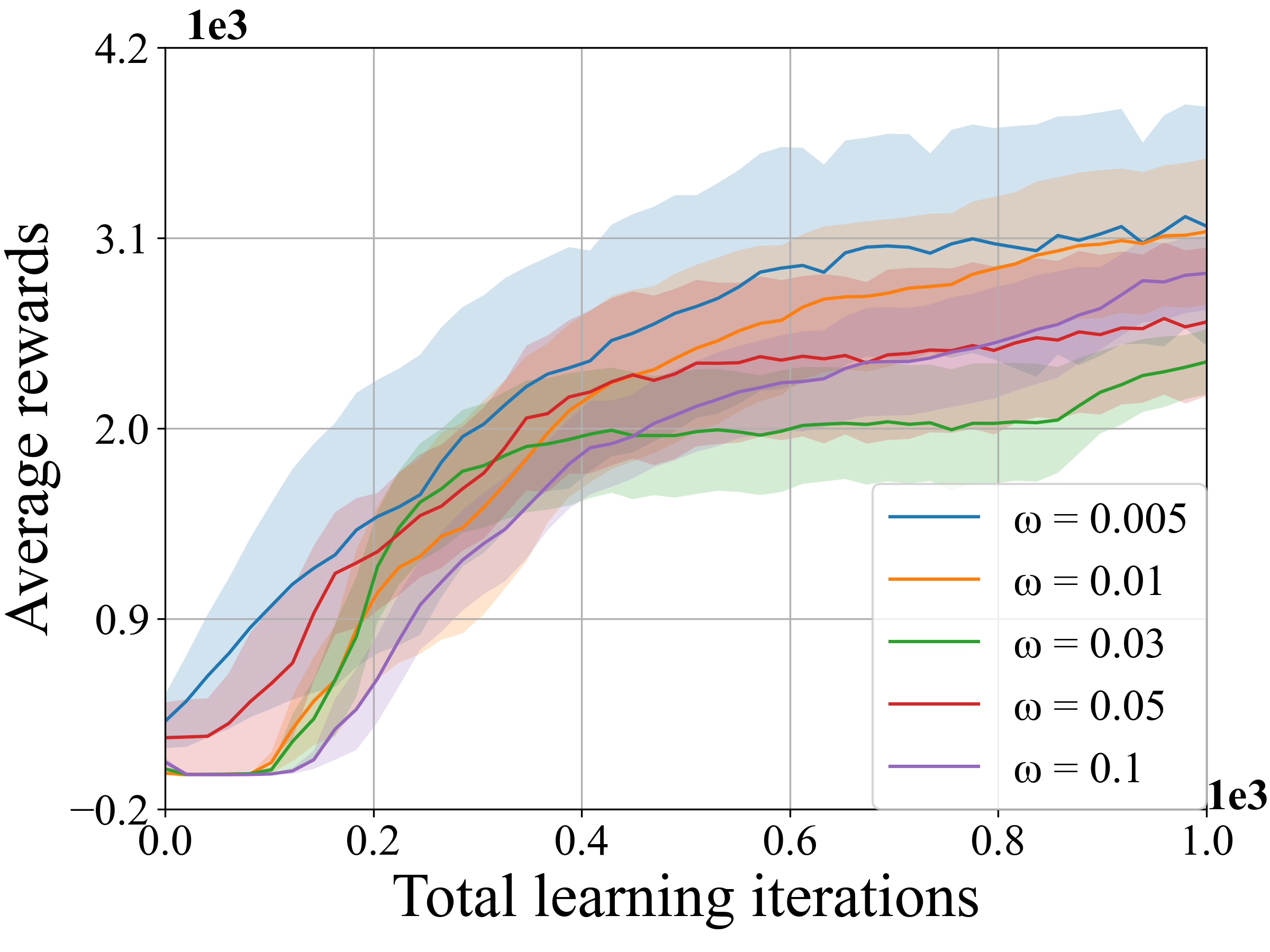}}
	\subfigure[HalfCheetah ($\beta$)]{\includegraphics[width=0.230\textwidth]{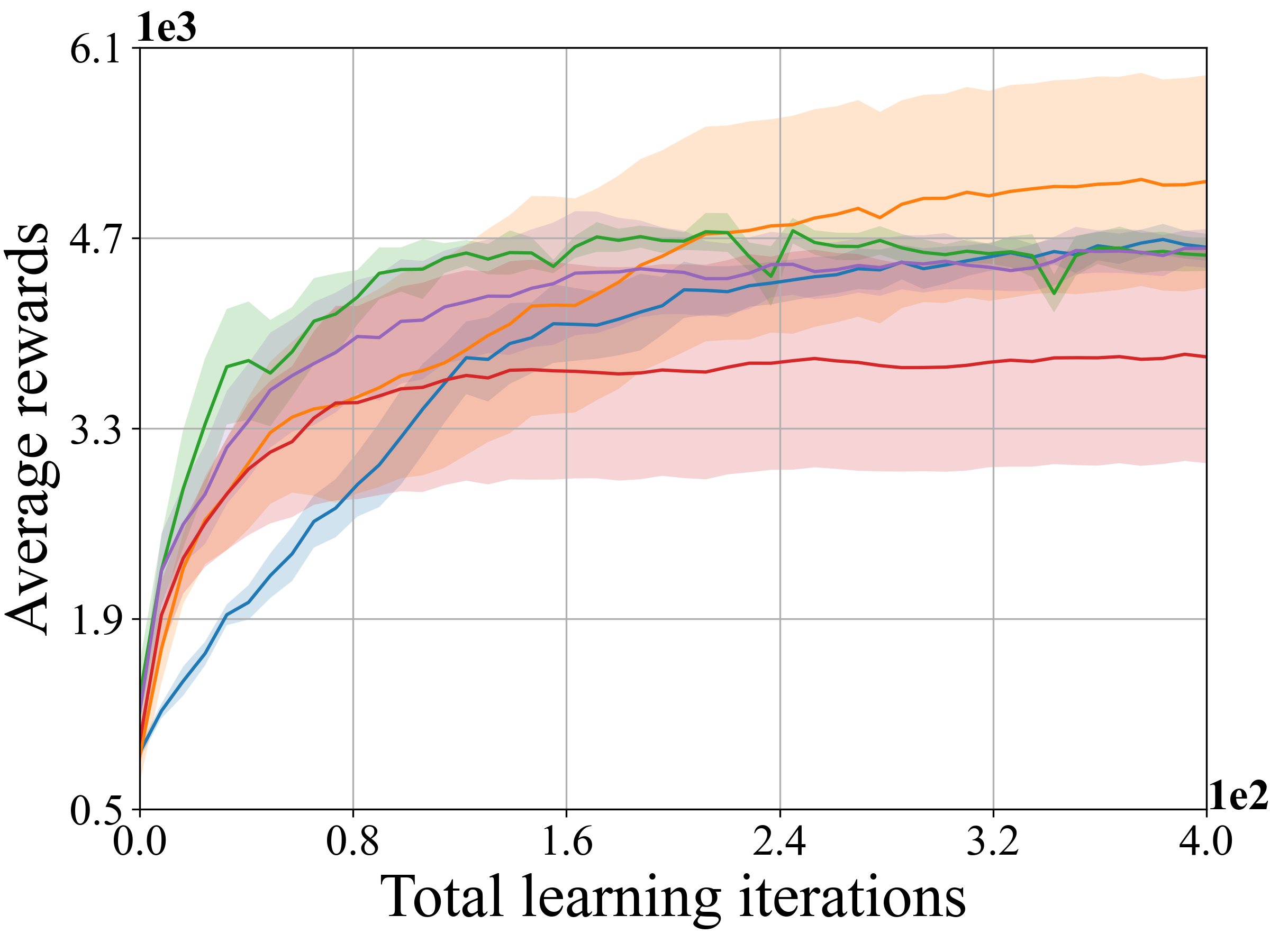}}
	\subfigure[Walker2d ($\beta$)]{\includegraphics[width=0.230\textwidth]{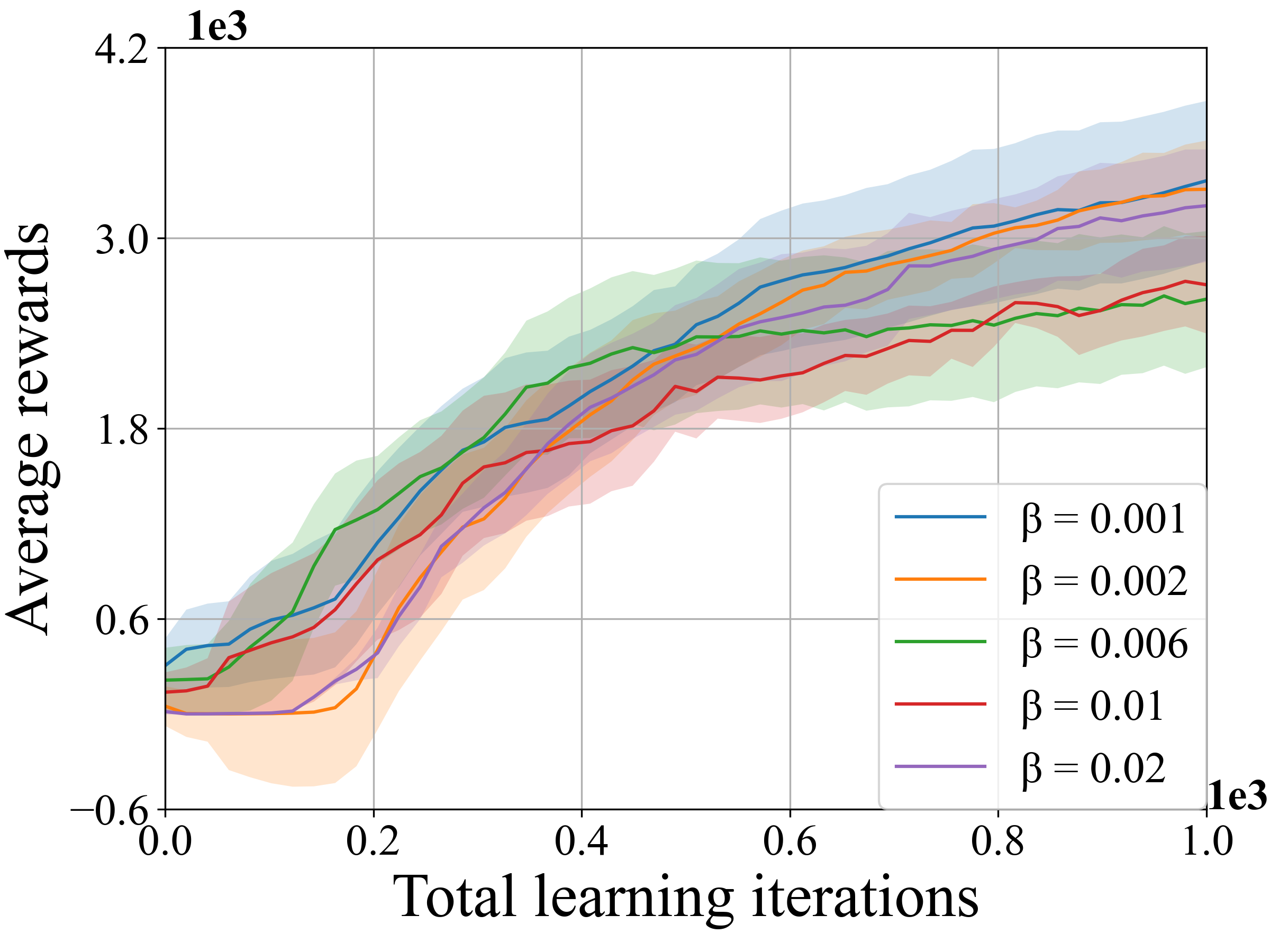}}
	\caption{Performance of BiERL with different noise covariances $\omega$ or learning rates $\beta$ of the meta-level, given the same number of iterations.}
	\label{fig.11}
\end{figure}

\section{Extended Results of Parameter Analysis}
Besides the two population sizes $n$ and $m$ analyzed in Section 4.2,
this appendix shows the influence of the other two hyperparameters, e.g., the noise covariance $\omega$ and the learning rate $\beta$ of the meta-level.
These experimental results are provided in Figure~\ref{fig.11}.  
We find that a smaller noise covariance or learning rate tends to achieve slightly higher performance. 
Generally, the performance of BiERL is much less sensitive to the meta-level's hyperparameters than those of the inner-level.


\section*{Supplementary References}

\begin{enumerate}[label={[}\arabic*{]}:]
	\item Edoardo Conti, Vashisht Madhavan, Felipe Petroski Such, et al., `Improving exploration in evolution strategies for deep reinforcement learning via a population of novelty-seeking agents', in \textit{Proceedings of Advances in Neural Information Processing Systems}, volume 31, (2018).
	
	\item Greg Brockman, Vicki Cheung, Ludwig Pettersson, et al., `OpenAI Gym', \textit{arXiv preprint arXiv:1606.01540}, (2016).
	
	\item Max Jaderberg, Valentin Dalibard, Simon Osindero, et al., `Population based training of neural networks', \textit{arXiv preprint arXiv:1711.09846}, (2017).
	
	\item Jianye Hao, Pengyi Li, Hongyao Tang, et al., `ERL-Re$^2$: Efficient Evolutionary Reinforcement Learning with Shared State Representation and Individual Policy Representation', in \textit{Proceedings of International Conference on Learning Representations}, (2023).
	
	\item Jack Parker-Holder, Vu Nguyen, and Stephen J Roberts, `Provably efficient online hyperparameter optimization with population-based bandits', in \textit{Proceedings of Advances in Neural Information Processing Systems}, volume 33, pp. 17200–17211, (2020).
	
	\item Tim Salimans, Jonathan Ho, Xi Chen, et al., `Evolution strategies as a scalable alternative to reinforcement learning', \textit{arXiv preprint arXiv:1703.03864}, (2017).
	
	\item Karush Suri, `Off-Policy Evolutionary Reinforcement Learning with Maximum Mutations', in \textit{Proceedings of International Conference on Autonomous Agents and Multiagent Systems}, pp. 1237–1245, (2022).
	
	\item Emanuel Todorov, Tom Erez, and Yuval Tassa, `MuJoCo: A physics engine for model-based control', in \textit{Proceedings of 2012 IEEE/RSJ International Conference on Intelligent Robots and Systems}, pp. 5026–5033, (2012).
\end{enumerate}

\end{document}